\documentclass{article}

 \usepackage[preprint]{neurips_2026}

\usepackage[utf8]{inputenc} 
\usepackage[T1]{fontenc}    
\usepackage{hyperref}       
\usepackage{url}            
\usepackage{booktabs}       
\usepackage{amsfonts}       
\usepackage{nicefrac}       
\usepackage{microtype}      
\usepackage{xcolor}         

\newcommand{\xhdr}[1]{\vspace{2mm}\noindent{{\bf #1.}}}
\usepackage{graphicx}
\usepackage{wrapfig}
\usepackage[dvipsnames]{xcolor}
\usepackage[table]{xcolor}
\usepackage[most]{tcolorbox}
\usepackage{bm}
\usepackage{enumitem}
\usepackage{multirow}
\usepackage{subcaption}
\usepackage{cleveref}
\usepackage{algorithm}
\usepackage{algpseudocode}
\usepackage{float}
\usepackage{fontawesome5}

\definecolor{AccentBlue}{HTML}{264E86}
\definecolor{SoftBlue}{HTML}{B0CCEC}
\definecolor{SoftGray}{HTML}{F6F7F9}
\definecolor{SoftGreen}{HTML}{EAF7EF}
\definecolor{SoftRed}{HTML}{FCEDEE}
\definecolor{RuleGray}{HTML}{D7DEE8}

\title{RankJudge: A Multi-Turn LLM-as-a-Judge \\ Synthetic Benchmark Generator}

\author{%
  Zhenwei Tang$^{1,2}$\thanks{Work done during an internship at Layer 6 AI.} \quad Zhaoyan Liu$^{1}$ \quad Rasa Hosseinzadeh$^{1}$ \quad Tongzi Wu$^{1}$ \\
  {\bf Keyvan Golestan$^{1}$ \quad Jesse C. Cresswell$^{1}$} \\
  $^{1}$Layer 6 AI \quad $^{2}$University of Toronto \\
  \texttt{\{joseph, zhaoyan, rasa, tongzi, keyvan, jesse\}@layer6.ai}
}

\begin{document}

\maketitle

\begin{center}
\vspace{-10pt}
\href{https://github.com/layer6ai-labs/RankJudge}{\faGithub~\texttt{Code}} \quad
\href{https://huggingface.co/datasets/Layer6/RankJudge}{\raisebox{-0.18em}{\includegraphics[height=1em]{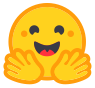}}~\texttt{Dataset}} \quad
\href{https://layer6ai-labs.github.io/RankJudge}{\faTrophy~\texttt{Leaderboard}}
\vspace{4pt}
\end{center}

\begin{abstract}
As interactive LLM-based applications are created and refined, model developers need to evaluate the quality of generated text along many possible axes. For simpler systems, human evaluation may be practical, but in complicated systems like conversational chatbots, the amount of generated text can overwhelm human annotation resources. Model developers have begun to rely heavily on auto-evaluation, where LLMs are also used to judge generation quality. However, existing LLM-as-a-judge benchmarks largely focus on simple Q\&A tasks that do not match the complexity of multi-turn conversations. We introduce RankJudge, a benchmark generator for evaluating LLM-as-a-judge on multi-turn conversations grounded in reference documents. RankJudge creates pairs of conversations where one conversation has a single flaw injected into one turn. This construction allows paired conversations to be labeled unambiguously as better or worse, and precisely isolates failure categories to individual turns, enabling a strict joint correctness criterion for judging. We implement RankJudge across the domains of machine learning, biomedicine, and finance, evaluate 21 frontier LLM judges, and rank those judges via the Bradley-Terry model. Our formulation also allows ranking each conversation pair with difficulty ratings, which we use to dynamically curate the evaluation slice to reduce label noise, as confirmed via human annotation. We find that judge rankings are stable under partial observability, coarser correctness criteria, and an alternative random-walk rating algorithm.
\end{abstract}
\vspace{-10pt}
\section{Introduction}
\vspace{-4pt}
Large language models are increasingly evaluated by other large language models (LLMs). Pairwise judging~\citep{zheng2023mtbench, chiang2024lmarena} has become the dominant scalable substitute for human preference collection. As judge models are now used to score training data, gate releases, and rank checkpoints, judge \emph{quality} has itself become a central assumption. A leaderboard built on a weak judge may silently reward the wrong behavior. Stress-testing the judges themselves is therefore a first-order problem, and one that existing judge benchmarks address only partially. Current judge benchmarks have recurring shortcomings: the dialogues that real-world LLM assistants produce are multi-turn and reference-grounded, while most judge benchmarks score isolated single-turn responses. Failure modes that matter in deployment, such as a later turn contradicting an earlier one or a content-level constraint silently dropping after several turns~\citep{cemri2025whymultifail, laban2025lost}, simply cannot surface in the single-turn setting. Another shortcoming is that verdict-only correctness conflates ``picked the right side'' with ``understood why'': a judge that prefers the better conversation while misattributing the flaw to the wrong turn or category has reached the right conclusion through the wrong reasoning, and existing leaderboards cannot tell the two apart. Lastly, static accuracy on a fixed pool offers no principled way to identify which items actually separate strong judges from weak ones~\citep{hendrycks2020mmlu, northcutt2021pervasive, gema2025donemmlu}.

\begin{figure}[!t]
    \centering
    \includegraphics[width=\textwidth]{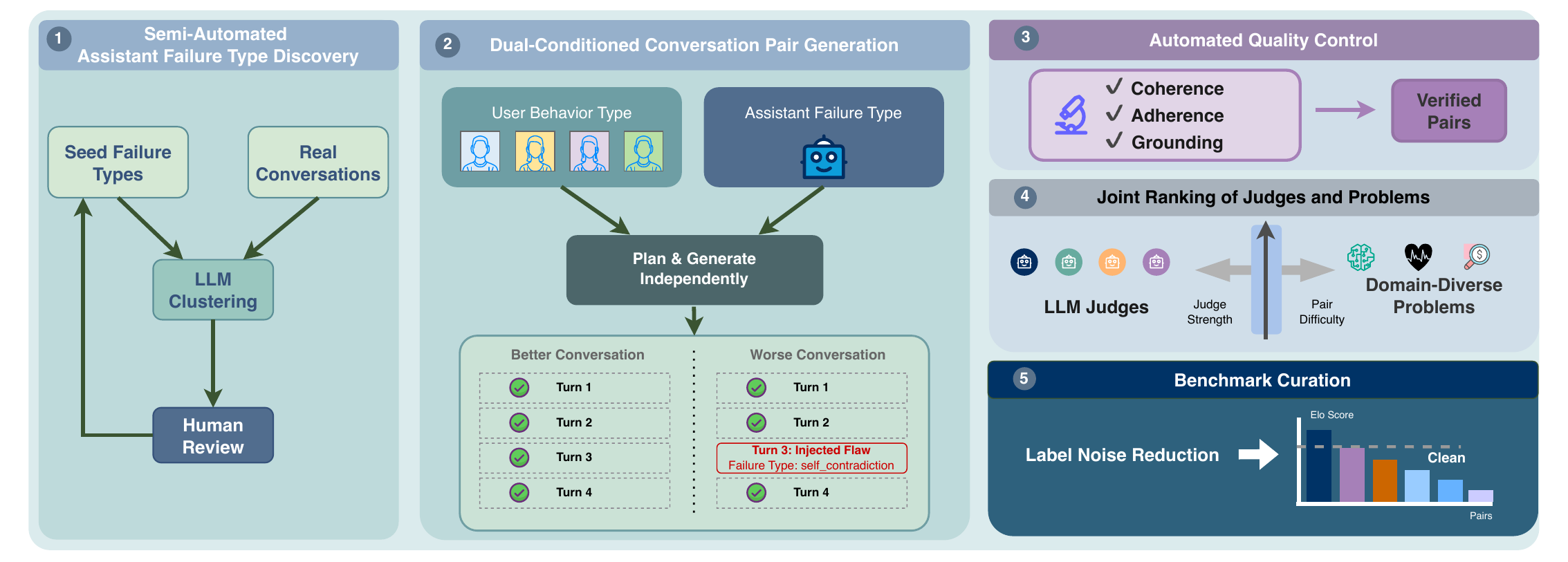}
    \caption{Overview of \textsc{RankJudge}, a benchmark \emph{generator} for multi-turn judge evaluation.}
    \label{fig:overview}
    \vspace{-14pt}
\end{figure}

In this paper, we introduce \textsc{RankJudge}, a benchmark \emph{generator} for multi-turn, reference-grounded judge evaluation. Each item is a pair of conversations sampled independently from the same reference document under two conditioning axes: a user behavior archetype, and a targeted assistant failure type, with the failure injected into exactly one turn of the worse branch. As shown in Figure~\ref{fig:overview}, because the flaw is \emph{preconstructed} by the generator, the ground-truth tuple of better conversation, flawed turn, and failure category is uniquely determined per item from the generation prompt itself, before any judge sees the pair. At evaluation time, we ask each judge for a joint prediction over verdict, turn, and type, and credit it only when all three components match. This consistency check distinguishes correct judgments from correct guesses.

The benchmark construction is \emph{fully synthetic}, with no per-item human label required, which lets us scale coverage densely and regenerate the pool deterministically whenever the generator or verifier is upgraded. We ensure the accuracy of our labels using two complementary methods. First, a three-layer automated verifier cascade checks for coherence, adherence, and grounding, and only keeps a pair of conversations when the targeted flaw is isolated to the correct turn. Additionally, every other claim in both conversations must be fully supported by the source. Second, we use the Bradley-Terry model to analyze how the judges scored the test pairs~\citep{chiang2024lmarena}. This gives us a calibrated difficulty rating for each pair and allows us to \emph{dynamically} curate a polished evaluation slice by removing the items with the very highest difficulty scores, i.e., top-Elo tail. Both a human audit and a held-out fine-tuning experiment independently flagged this tail as the subset containing label noise.

We apply \textsc{RankJudge} to produce three benchmarks in distinct knowledge-intensive domains: Machine Learning, Biomedicine, and Finance, and evaluate $21$ frontier judges spanning proprietary and open-weight families on each. The leaderboard separates judges across nearly $1200$ Elo points, and several open-weight checkpoints outrank frontier proprietary judges. The bipartite framing also admits partial observability, so judges can be scored on different subsets of pairs while retaining their positions on the same scale, which lowers the required compute. The resulting ranking is stable under match subsampling, under a coarser correctness criterion, and under an Empirical Interaction Propagation (EIP) cross-check~\citep{hu2026eip}. 
\textsc{RankJudge} also surfaces a model-capability ceiling: weaker judges collapse their predictions onto a single failure category rather than scattering across the taxonomy, and targeted prompt rewrites of a mid-ranked frontier judge fail to lift it onto the accuracy-cost Pareto frontier, exposing a capability gap that prompting cannot close. We summarize our contributions as follows:
\begin{itemize}[noitemsep,topsep=0pt,leftmargin=*]
    \item \textsc{RankJudge} is a benchmark generator for multi-turn, reference-grounded judge evaluation whose ground-truth verdict, flawed turn, and failure type are specified in the generation prompt and then scored under a joint correctness criterion.
    \item A semi-automated discovery loop surfaces multi-turn assistant failure types, and dual-conditioned generation independently simulates user-behavior and assistant-failure axes.
    \item Construction is fully synthetic; a three-layer automated verifier and Elo-based curation of the high-difficulty tail are validated by a human audit and a held-out fine-tuning experiment that independently flag a substantially overlapping noisy slice.
    \item Instantiations in Machine Learning, Biomedicine, and Finance produce leaderboards spanning proprietary and open-weight judge families, which remain stable under various conditions, and surface a systematic class bias in weaker judges.
\end{itemize}

\vspace{-4pt}
\section{Related Work}
\vspace{-4pt}

\xhdr{Multi-turn LLM Benchmarks}
LLM evaluation has shifted from single-turn benchmarks like MMLU~\cite{hendrycks2020mmlu} and GSM8K~\cite{cobbe2021training}, which miss the user-model-environment dynamics that drive real-world utility~\cite{wang2023mint, deshpande2025multichallenge}, to multi-turn frameworks~\cite{zheng2023mtbench, kwan2024mt, fan2026halluhardhardmultiturnhallucination, eisenstein2026mtpingevalevaluatingmultiturncollaboration} that probe correctness, helpfulness, and interactive patterns~\cite{li2026singleturnsurveymultiturninteractions}. A consistent finding emerges across these works: single-turn ability does not transfer to multi-turn success~\cite{wang2023mint}, and frontier models degrade sharply across turns due to compounding unreliability~\cite{laban2025lost}. These dynamics motivate our focus on multi-turn, reference-grounded conversations as the setting in which judge quality must itself be stress-tested.

\xhdr{LLM-as-a-Judge}
Reward models are crucial for aligning and improving the capabilities of LLMs ~\cite{ouyang2022training, christiano2017deep}. The traditional scalar reward model~\cite{NEURIPS2020_1f89885d} gives a single ``verdict'' indicating the response quality. However, scalar models suffer from certain limitations, for example they are vulnerable to hacking ~\cite{xu2025ask}, and lack the ability to localize or categorize specific errors. LLMs have demonstrated strong capability to mimic human reasoning and evaluate inputs based on predefined criteria while being scalable and effective. The concept of LLM-as-a-judge~\cite{zheng2023mtbench,wang-etal-2023-chatgpt, liu2023g} has become widely used for tasks like providing rich reward signals for LLM alignment~\cite{lee2023rlaif}, producing chain-of-thought (CoT) reasoning along with a final judgment as evaluators~\cite{kim2023prometheus, saha2025learning}, and data annotation~\cite{luo2023wizardmath, chen2024mllm}. These judges can be implemented either via direct prompting of general-purpose LLMs~\cite{zheng2023mtbench, wang2025mcts} or as specialized fine-tuned evaluators~\cite{whitehouse2026j1, chen2025rm}. Existing frameworks typically adopt either pointwise or pairwise evaluation protocols. Pointwise methods~\cite{liu2023g, kim2023prometheus} score responses independently, while pairwise methods~\cite{zheng2023mtbench, whitehouse2026j1} compare responses to predict relative preferences.

\begin{wraptable}[22]{r}{0.5\textwidth}
\centering
\scriptsize
\setlength{\tabcolsep}{1.5pt}
\renewcommand{\arraystretch}{0.95}
\vspace{-12pt}
\begin{tabular}{@{}lccccccc@{}}
\toprule
 &
\textbf{Turn} &
\textbf{Drv.} &
\textbf{Usr.} &
\textbf{Doc.} &
\textbf{Diag.} &
\textbf{Pipe.} &
\textbf{Eval.} \\
\midrule
LLMBar~\cite{zeng2024evaluating}
& S & -- & -- & -- & -- & -- & Acc. \\

JudgeBench~\cite{tanjudgebench}
& S & \checkmark & -- & -- & -- & $\triangle$ & Acc. \\

ContextualJudgeBench~\cite{xu-etal-2025-context}
& S & $\triangle$ & -- & \checkmark & $\triangle$ & $\triangle$ & Acc. \\

DHP~\cite{wang-etal-2025-dhp}
& S & \checkmark & -- & -- & $\triangle$ & \checkmark & Disc. \\

ReIFE~\cite{liu-etal-2025-reife}
& S & -- & -- & -- & -- & -- & Acc. \\

JuStRank~\cite{gera-etal-2025-justrank}
& S & -- & -- & -- & -- & -- & Rank \\

JudgeArena~\cite{judgearena_2024}
& S & -- & -- & -- & -- & -- & Rank \\

MEDAL~\cite{mendoncca2026medal}
& M & $\triangle$ & $\triangle$ & -- & $\triangle$ & \checkmark & Acc. \\
\midrule
\textbf{RankJudge}
& M & \checkmark & \checkmark & \checkmark & \checkmark & \checkmark & Rank \\
\bottomrule
\end{tabular}
\caption{
Comparison of LLM judge meta evaluation benchmarks.
Turn: S = single turn, M = multi turn.
Drv. = derived meta evaluation target:
\checkmark{} means without new human preference labeling,
$\triangle$ means some manual curation and
-- means completely dependent on human judgments.
Usr. = controlled user behavior.
Doc. = document grounding.
Diag. = diagnostic labels beyond a winner or score.
Pipe. = automated benchmark generation where $\triangle$ means partially.
Eval.: Acc. = item level accuracy or agreement; Disc. = perturbation discernment; Rank = judge or system ranking.
}
\label{tab:llm-as-judge-benchmark-comparison}
\end{wraptable}

\xhdr{Benchmarking LLM-as-a-Judge}
MT-Bench~\cite{zheng2023mtbench,bai2024mtbench101} helped establish LLM-as-a-judge evaluation for chat assistants by reporting agreement with humans. Initial works on meta-evaluation of judges focused on single-turn settings: LLMBar~\cite{zeng2024evaluating} uses natural and adversarial pairwise examples, DHP~\cite{wang-etal-2025-dhp} measures natural language generation evaluation capabilities using perturbations, ReIFE~\cite{liu-etal-2025-reife} varies LLMs, protocols and datasets. JudgeBench~\cite{tanjudgebench} converts factuality and correctness datasets into benchmarks for meta-evaluation, JuStRank~\cite{gera-etal-2025-justrank} studies judges through systems level ranking agreement with human rankings, and ContextualJudgeBench~\cite{xu-etal-2025-context} grounds evaluation in external documents. Other works study LLM-as-a-Judge for code evaluation~\cite{wang2025axiom} and evaluator adversarial robustness~\cite{li2025llms}, positional bias~\cite{shi-etal-2025-judging}, and fairness~\cite{zhang2023wider}. 
MEDAL~\cite{mendoncca2026medal} is closest to our setting since it generates multilingual multi-turn dialogues using a multi-agent pipeline and automates labeling with \texttt{GPT-4.1}, followed by filtering with human curation for the final benchmark. Table~\ref{tab:llm-as-judge-benchmark-comparison} compares \textsc{RankJudge} with prior benchmarks across several axes: ours is the first automated pipeline for generating a multi-turn judge benchmark that is grounded in external documents, conditioned on user behavior, and built by injecting controlled error types.

\vspace{-4pt}
\section{Methodology}
\vspace{-4pt}
Let $\mathcal{T} = \{t_1, \ldots, t_K\}$ denote a taxonomy of $K$ assistant failure types. A multi-turn conversation $C = (u_1, a_1, \ldots, u_R, a_R)$ consists of $R$ turns, each a (user, assistant) message pair. Each benchmark item is a tuple $(C_A, C_B, y^\star, r^\star, t^\star)$ in which $C_A$ and $C_B$ are two conversations grounded in the same reference documents, $y^\star \in \{A, B\}$ identifies the better conversation, $r^\star \in \{1, \ldots, R\}$ is the turn of the single injected flaw in the worse conversation, and $t^\star \in \mathcal{T}$ is its failure category. At turn $r^\star$, one flaw type $t^\star$ is injected, making $(y^\star, r^\star, t^\star)$ uniquely determined per item (see Section~\ref{sec:dataset}).
A judge is a function $J : (C_A, C_B) \mapsto (\hat{y}, \hat{r}, \hat{t})$ that jointly predicts the better conversation, the flawed turn, and the failure category of the flawed turn. This joint prediction enables a check on the judge's understanding of \emph{why} one conversation is better than another. We credit a judge only when every component matches the ground truth,
\begin{align}
c &= \mathbf{1}\!\left[\hat{y} = y^\star \,\wedge\, \hat{r} = r^\star \,\wedge\, \hat{t} = t^\star\right] \in \{0, 1\}.
\label{eq:main_correctness}
\end{align}
A judge that picks the right conversation while localizing the flaw in the wrong turn, or assigning it to the wrong taxonomy entry, has reached the correct conclusion without identifying the underlying failure, and is not credited. \vspace{-4pt}
\subsection{Benchmark Construction}
\label{sec:dataset}
\vspace{-8pt}
\xhdr{Semi-Automated Assistant Failure Type Discovery}
\begin{table}[t]
\centering
\caption{Multi-turn assistant-failure taxonomy. In each pair, one conversation exhibits one flaw from the available failure types in exactly one round; the other exhibits no clear flaw throughout. Failure types (left) are the ground-truth labels $t^\star$.}
\label{tab:taxonomy-assistant}
\tiny
\setlength{\tabcolsep}{6pt}
\renewcommand{\arraystretch}{1.0}
\begin{tabular}{l p{0.76\linewidth}}
\toprule
Failure Type & Behavior \\
\midrule
\multirow{2}{*}{\texttt{self\_contradiction}}
 & \cellcolor{SoftRed}\textbf{Flaw.} A later turn contradicts an earlier correct claim by shifting a qualifier or reframing. \\
 & \cellcolor{SoftGreen}\textbf{Benign.} Consistent scope and qualifiers across turns; flags genuine changes in position. \\
\addlinespace[2pt]
\multirow{2}{*}{\texttt{evasion}}
 & \cellcolor{SoftRed}\textbf{Flaw.} Hedges or drifts around the central claim, citing peripheral details without the figures or terms needed to verify or act. \\
 & \cellcolor{SoftGreen}\textbf{Benign.} Answers the question directly with the exact figures and terms required. \\
\addlinespace[2pt]
\multirow{2}{*}{\texttt{disorganized}}
 & \cellcolor{SoftRed}\textbf{Flaw.} Buries the key answer in a muddled structure, mixing high- and low-level points so correct content is costly to extract. \\
 & \cellcolor{SoftGreen}\textbf{Benign.} Leads with the key point and arranges supporting detail in a scannable structure. \\
\addlinespace[2pt]
\multirow{2}{*}{\texttt{fabricated\_answer}}
 & \cellcolor{SoftRed}\textbf{Flaw.} When a question is unanswerable from the source, invents a confident answer instead of admitting insufficient information. \\
 & \cellcolor{SoftGreen}\textbf{Benign.} Calibrated to the source: confident when supported, explicit about limits otherwise. \\
\addlinespace[2pt]
\multirow{2}{*}{\texttt{instruction\_forgetting}}
 & \cellcolor{SoftRed}\textbf{Flaw.} A directive set early (scope, audience, what to skip) is explicitly violated in a later turn while the answer remains plausible. \\
 & \cellcolor{SoftGreen}\textbf{Benign.} Tracks the early directive across turns, surfacing genuine conflicts rather than deviating. \\
\addlinespace[2pt]
\multirow{2}{*}{\texttt{no\_clarification}}
 & \cellcolor{SoftRed}\textbf{Flaw.} On an ambiguous question, picks one interpretation and answers confidently rather than asking the user to clarify. \\
 & \cellcolor{SoftGreen}\textbf{Benign.} Clarifies on truly ambiguous questions; answers unambiguous ones directly. \\
\addlinespace[2pt]
\multirow{2}{*}{\texttt{unnecessary\_refusal}}
 & \cellcolor{SoftRed}\textbf{Flaw.} Refuses on a question that is answerable from the source, citing insufficient information when the answer is available. \\
 & \cellcolor{SoftGreen}\textbf{Benign.} Answers confidently from available material; hedges only when genuinely out of scope. \\
\bottomrule
\end{tabular}
\vspace{-10pt}
\end{table}
We construct our taxonomy of assistant behavior types through a semi-automated discovery procedure. We first seed an initial set of behavior categories that are commonly observed in multi-turn conversations, drawing from and organizing prior works~\cite{cemri2025whymultifail,laban2025lost,kartavc2026insidedialogue} on dialogue evaluation and assistant failure modes. To assess coverage, we then prompt \texttt{Gemini-3.1-Pro}~\cite{google2026gemini31} with samples from MT-Bench~\cite{zheng2023mtbench} and MT-Bench 101~\cite{bai2024mtbench101}, asking the model to verify whether each instance is captured by the existing taxonomy and, if not, to propose new assistant error types grounded in the observed failure. 

We scope the taxonomy to failures characteristic of multi-turn assistant behavior; coarse single-turn categories such as factual error are excluded as standalone types \cite{leung2026classifying}, since their multi-turn manifestations are already absorbed by more specific types. For instance, an assistant that asserts a fact in turn 2 and contradicts it in turn 5 is captured by \texttt{self\_contradiction} rather than a generic hallucination, and an assistant that drops a user-specified constraint after several turns is captured by \texttt{instruction\_forgetting}. Additionally, each type targets failures that are plausible for a capable assistant yet difficult to spot by surface inspection, a requirement we make explicit in every flaw description so that the resulting probes stress, rather than merely confirm, the discriminative ability of strong LLM judges. Furthermore, categories are designed to have non-overlapping decision boundaries so that judges can unambiguously classify the failure type. Candidate failure types that do have overlap are merged into an existing type or dropped.

Failure type discovery was supervised by the authors, who reviewed each candidate type before admission and adjudicated borderline cases and overlaps. A condensed view of the resulting taxonomy is presented in Table~\ref{tab:taxonomy-assistant}, with the full set of types and definitions deferred to Appendix~\ref{sec:taxonomy_assistant_full}, and the exact prompt used to elicit new error types from \texttt{Gemini-3.1-Pro} provided in Appendix~\ref{sec:prompt_discovery}.

\vspace{-4pt}
\xhdr{Dual Conditioned Conversation Pair Generation}
Each conversation pair is sampled under two independent conditions: an \emph{assistant failure type} from the seven options in Table~\ref{tab:taxonomy-assistant}, as well as a \emph{user behavior type}. The assistant failure axis fixes the ground truth: by sampling a target type and instructing the generator to inject one error of that type, we unambiguously define the worse conversation of the pair, the turn in which the failure occurs, and the failure category. The user behavior axis adds diversity, since different user types surface different slices of the reference material and create different turn-to-turn dynamics as seen in real multi-turn use. Behaviours span seven archetypes: \texttt{focused}, \texttt{integrative}, \texttt{scattered}, \texttt{skeptical}, \texttt{misinformed}, \texttt{exploratory}, and \texttt{underspecified}; each represents a style prompt which the user is conditioned on during generation, as defined in Appendix~\ref{sec:taxonomy_user}.

Each pair of better and worse conversations is produced by two separate sets of generation calls that share reference documents. Past works have created negative examples by simply injecting errors into existing text~\cite{li2023halueval,zeng2024evaluating,wang2025axiom,kong2025aegis}. However, when comparing two alternate conversations, if the injected error is the only change, the judge can isolate this difference rather than making a complete assessment of quality. Independently sampling two conversations is also insufficient; if only the bad version faced situations where the target failure could surface, judges could again shortcut a holistic comparison by pattern-matching on question types. Therefore, when generating the better conversation, we actively stage the conditions under which the selected flaw would be relevant, but condition on benign behaviour (Table~\ref{tab:taxonomy-assistant}), which describes the correct way to handle the conversational pressure. For instance when the failure type is \texttt{fabricated\_answer}, the user in the better conversation still asks an out-of-scope question, but the assistant is instructed to explicitly state the limits of its knowledge. Paired conversations are thus structurally matched on topic and conversational dynamics, differing only in handling.

Both sets of generation calls follow a turn-by-turn \emph{blueprint}. Each turn specifies what the user's question should be about, and the chunk of the reference document the assistant will need to draw on. The blueprint for the worse conversation additionally commits to a \texttt{bad\_round\_index} and a sketch of how the selected flaw should manifest. 
To keep the comparison non-trivial, the blueprint imposes requirements that the flawed turn must maintain the same tone and length as other turns, while lexical announcements of the kind ``stepping outside the scope for a moment'' are disallowed. We remove ordering bias by randomizing which conversation in the pair (\textsc{A}/\textsc{B}) is assigned as worse. The full generation prompts are provided in Appendix~\ref{sec:prompt_generation}.
\begin{wrapfigure}{r}{0.5\textwidth}
\vspace{-6pt}
\begin{minipage}{0.5\textwidth}
\begin{algorithm}[H]
\scriptsize
\caption{Data generation and verification}
\label{alg:pipeline}
\begin{algorithmic}[1]
\For{each of $N_\text{cand}$ candidate samples}
\State Sample failure type $t^\star \in \mathcal{T}$, user behavior, reference documents, verdict $y^\star \in \{A,B\}$, flawed turn $r^\star \in \{1,\dots,R\}$
\State Generate paired blueprints conditioned on the sampled inputs
\State If \textbf{coherence} check of the blueprints fails, skip
\State Generate paired conversations $(C_A, C_B)$ from the blueprints
\State If \textbf{adherence} check of $(C_A, C_B)$ to blueprints fails, skip
\State If \textbf{grounding} check of all atomic claims fails, skip
\State Retain $(C_A, C_B, y^\star, r^\star, t^\star)$
\EndFor
\end{algorithmic}
\end{algorithm}
\end{minipage}
\vspace{-10pt}
\end{wrapfigure}

\vspace{-4pt}
\xhdr{Automated Quality Control}
\label{sec:verification}
A synthetic pair is only useful to the benchmark if the targeted weakness actually surfaces, appears only in the declared turn, and the rest of the content is free of clear failures. To make these judgments, we rely on the well-documented asymmetry that verification is substantially easier than generation~\citep{cobbe2021training, saunders2022self, lightman2023let}. We use a three-layer verification cascade run by an external verifier model over every candidate pair, adding the pair to the benchmark only if it passes all three layers. Each layer is strictly discriminative where the verifier is given the intended labels $(y^\star, r^\star, t^\star)$ and the reference documents. Note that the verifier solves an easier subproblem than the judges we aim to evaluate with the benchmark, since the verifier is conditioned on the ground truth.

The three layers of verification check coherence, adherence, and grounding. The \emph{coherence} check tests the sampled blueprint by comparing the per-turn outline of user intent, assistant focus, and the chosen failure turn against the intended ground truth and reference material. This check flags genuine semantic conflicts, e.g., a blueprint that is inconsistent with the chosen user behavior or failure location. In the \emph{adherence} check, conversations are examined to ensure both the user and assistant follow the blueprint globally. The better conversation must display benign behaviour across all turns, and the worse conversation must exhibit the targeted flaw in exactly the declared turn. A conversation pair fails the adherence check if the user deviates from the specified behavior, the failure drifts to a different turn, or if multiple flaws are present. Finally, in the \emph{grounding} check the verifier extracts every atomic factual claim from each assistant's turn and labels each claim as grounded or not based on the reference context. A pair passes only if every turn (other than $r^\star$ in the worse conversation) is fully grounded, ensuring that the only unsupported claim is the targeted flaw.

Per-layer verification pass rates and overall retention across the three knowledge domains are reported in Table~\ref{tab:dataset-stats-domain}. The full verification prompts are provided in Appendix~\ref{sec:prompt_verification}.

\vspace{-4pt}
\subsection{Joint Ranking of Judges and Conversation Pairs}
\label{sec:eval_protocol}
\vspace{-4pt}
Let $\mathcal{D} = \{(j_t, q_t, c_t)\}_{t=1}^{N}$ be the judgment results over judges $\mathcal{J} = \{1, \dots, M\}$ across conversation pairs $\mathcal{Q} = \{1, \dots, Q\}$, with $c_t \in \{0, 1\}$ marking whether the judge correctly identified the joint criterion $(y^\star, r^\star, t^\star)$. Rather than simply report accuracies over a fixed test set, we use $\mathcal{D}$ to rate judges for two reasons. First, arena-style ratings~\cite{chiang2024lmarena} are relative to a population and tolerate partial observability of the judgment results, which enables leaderboard construction without requiring full judge-pair coverage. Second, rating judges and conversation pairs jointly assigns each pair a calibrated difficulty rating relative to the set of judges. This lets us dynamically curate the benchmark by difficulty. Specifically, our published evaluation slice in Section~\ref{sec:dataset_curation} drops the top tail of most difficult pairs because our human audit of conversation quality found this segment to have the most label noise, see~\Cref{fig:label_noise}.\vspace{-2pt}

Two rating algorithms are compatible with our bipartite framing: Bradley--Terry (BT) rating, used by LM Arena~\citep{chiang2024lmarena} for LLM-vs-LLM matchups, and Empirical Interaction Propagation (EIP)~\citep{hu2026eip}, which is a PageRank-style random walk on the correctness graph. However, EIP requires \emph{full} observability of $\mathcal{D}$, so we adopt BT as our default and use EIP only as a cross-check. Under BT, each judge and each pair is assigned a positive strength $\theta_i > 0$, $i \in \mathcal{J} \cup \mathcal{Q}$. The probability that judge $j$ produces a correct verdict on pair $q$ is $\Pr(c = 1 \mid j, q) = \theta_j / (\theta_j + \theta_q)$, fit by maximum-likelihood estimation using the minorization--maximization iteration with mean-normalization~\cite{hunter2005variable}, and mapped to Elo via the standard $400 \log_{10}$ transform. Under full observability of $\mathcal{D}$, where every judge is scored on every pair, the BT ordering of judges agrees with their raw win count. BT adds two properties EIP lacks: closed-form cluster-robust confidence intervals, and stability under partial observability. Full equations, convergence criteria, and cluster-robust pair-level standard errors are deferred to Appendix~\ref{sec:bt_details}, with similar details for EIP in Appendix~\ref{sec:eip}.\vspace{-2pt}

A pair on which every judge produces the same verdict carries zero information about relative judge ability. Under the BT likelihood it contributes a constant regardless of the ratings being fit. We therefore drop any pair whose verdicts are unanimous across the current judge set. The unanimously-incorrect pairs are additionally likely to be mislabelled and would in any case be removed by the top-Elo curation step introduced next. \vspace{-4pt}

\vspace{-4pt}
\subsection{Benchmark Curation}
\label{sec:dataset_curation}
\vspace{-4pt}
Our benchmark is completely synthetic and requires no human annotation during construction, which lets us cover the taxonomy of Section~\ref{sec:dataset}, regenerate the pool when the verifier or generator is upgraded, and scale the benchmark without proportional annotation budget. The resulting data can still be noisy, but noise is not unique to synthetic benchmarks. Widely used benchmarks carry a non-trivial fraction of mislabeled or ambiguous items; MMLU~\citep{hendrycks2020mmlu} has roughly $6\%$ erroneous questions in heavily evaluated subsets~\citep{gema2025donemmlu} and audits of the ten most-cited vision benchmarks find an average test-set error rate near $3\%$~\citep{northcutt2021pervasive}.

\begin{figure}[!t]
    \centering
    \includegraphics[width=0.9\textwidth]{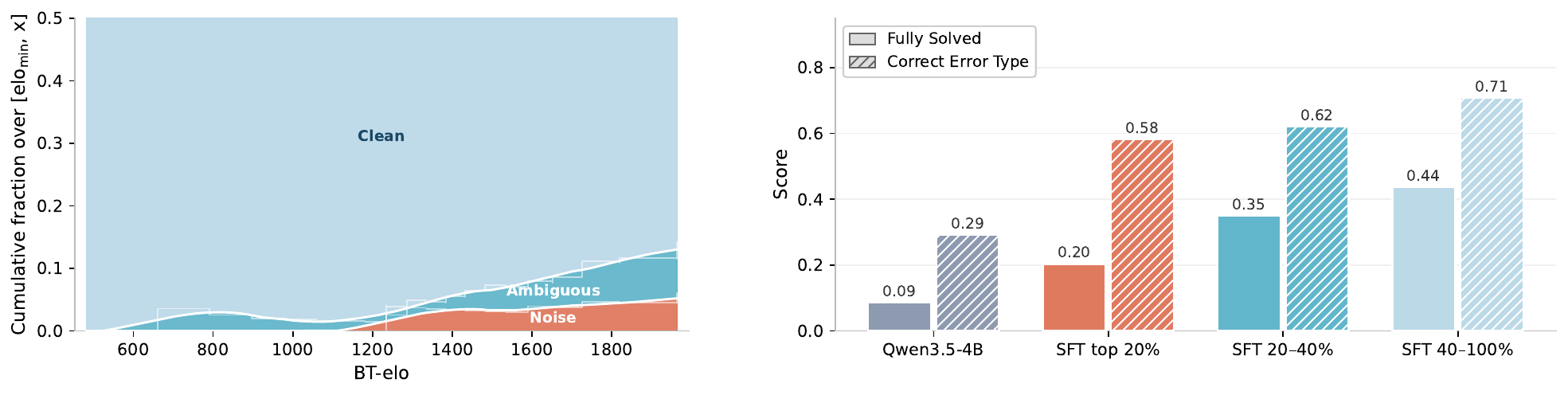}
    \caption{(Left) Cumulative fraction of samples annotated by humans as having ambiguous or noisy labels among conversation pairs with BT-Elo at most $x$. Both problematic types concentrate at the top of the difficulty distribution while easier pairs are almost entirely unambiguous. (Right) Fine-tuning \texttt{Qwen3.5-4B} on the top-20\% slice underperforms training on the cleaner slices.}
    \label{fig:label_noise}
    \vspace{-8pt}
\end{figure}

This puts curation in tension with discrimination. The hardest pairs are the most informative for separating out stronger judges, but they are the most likely to carry label noise or be ambiguous. We resolve this tradeoff conservatively: the published evaluation slice drops the top $5\%$ of pairs by BT Elo (column \emph{Top 5\%} of Table~\ref{tab:dataset-stats-domain}; per-axis breakdowns in Tables~\ref{tab:dataset-stats-assistant} and~\ref{tab:dataset-stats-user}). BT pair difficulty is judge-set-aware, so the cut is recomputed whenever the judge set changes. Two analyses support this cut: a human audit and a downstream SFT experiment, both of which indicate that the top-Elo slice is disproportionately likely to contain label noise, see \Cref{fig:label_noise}.

\vspace{-4pt}
\xhdr{Human Verification}
The audit described here sits outside the benchmark construction loop. The benchmark is built end-to-end without manual curation, while this audit confirms the quality of the automated verification and justifies the design choice of dropping the top-Elo slice. The authors manually labeled a uniform random sample of $120$ of the $703$ informative pairs as \texttt{clean}, \texttt{ambiguous}, or \texttt{noise}. \texttt{Noise} indicated that the conversation pair did not strictly match the intended ground truth $(y^\star, r^\star, t^\star)$, for example because more than one flaw was present. \texttt{Ambiguous} indicated that the assistant failure could not unambiguously be assigned a single type $t^\star$ in the taxonomy, but all other details were consistent. Unambiguous and correctly labeled conversation pairs were considered \texttt{clean}. The overall noise rate is low. \Cref{fig:label_noise} (left) plots the cumulative fraction of audit labels across pairs with BT-Elo at most $x$, where we find the \texttt{noise} and \texttt{ambiguous} bands rise only in the right tail. Given the labeling effort already conducted, we do drop the $17$ problematic pairs from the published slice (column \emph{Human}\ of Table~\ref{tab:dataset-stats-domain}), but the curation pipeline does not depend on this step. The full annotation procedure, and the pair visualization tool we used are in Appendix~\ref{sec:human_eval}.

\vspace{-4pt}
\xhdr{SFT with Label Noise}
A second analysis tests whether top-Elo pairs degrade model fine-tuning results. We split the informative pairs not used in the human audit into three BT Elo bands (top $20\%$, $20$--$40\%$, and the remaining $60\%$), sample an equal number of pairs from each, and fine-tune \texttt{Qwen3.5-4B}~\cite{qwen35blog} with LoRA~\cite{hu2022lora} under an identical recipe per band (full hyperparameters in Appendix~\ref{sec:sft_details}). The \texttt{clean} pairs from the audit are held out for evaluation. Joint correctness of the fine-tuned judge improves monotonically as the training band moves down the ranking, and the bottom band achieves more than twice the score of the top band (Figure~\ref{fig:label_noise}, right). This signal is consistent with the human audit: the top-ranked band of conversation pairs is flagged as the noisiest, leading to our decision to cut the top of the Elo distribution from the published evaluation slice.

\begin{table}[!t]
\centering
\caption{Verification and filtering cascade, split by domain. The first three filters represent automated quality verification during pair generation. The last three are applied at scoring time: \emph{Uninformative}\ drops uninformative pairs where every judge gave the same verdict; \emph{Human}\ drops pairs that human annotators flagged as ambiguous or noisy. The pipeline itself is fully automated and does not depend on human annotation; we apply this layer opportunistically to the 120-pair subset that was annotated for other analyses, since the flagged pairs can be excluded at no additional cost once those labels exist. \emph{Top 5\%}\ drops the top 5\% of pairs by Elo, which are more likely to contain noisy labels.}
\label{tab:dataset-stats-domain}
\footnotesize
\setlength{\tabcolsep}{4pt}
\renewcommand{\arraystretch}{1.15}
\begin{tabular}{l|c|ccc|c>{\columncolor{black!8}}cc|c}
\toprule
Domain & Generated & Coherence & Adherence & Grounding & Uninformative & Human & Top 5\% & Survival \\
\midrule
\textsc{ML} & 400 & 397 {\scriptsize($-3$)} & 342 {\scriptsize($-55$)} & 304 {\scriptsize($-38$)} & 263 {\scriptsize($-41$)} & 255 {\scriptsize($-8$)} & 240 {\scriptsize($-15$)} & 60.0\% \\
\textsc{Med} & 400 & 400 {\scriptsize($-0$)} & 344 {\scriptsize($-56$)} & 243 {\scriptsize($-101$)} & 208 {\scriptsize($-35$)} & 202 {\scriptsize($-6$)} & 194 {\scriptsize($-8$)} & 48.5\% \\
\textsc{Fin} & 400 & 397 {\scriptsize($-3$)} & 345 {\scriptsize($-52$)} & 274 {\scriptsize($-71$)} & 232 {\scriptsize($-42$)} & 229 {\scriptsize($-3$)} & 218 {\scriptsize($-11$)} & 54.5\% \\
\midrule
\textbf{All} & 1200 & 1194 {\scriptsize($-6$)} & 1031 {\scriptsize($-163$)} & 821 {\scriptsize($-210$)} & 703 {\scriptsize($-118$)} & 686 {\scriptsize($-17$)} & 652 {\scriptsize($-34$)} & 54.3\% \\
\bottomrule
\end{tabular}
\end{table}

\vspace{-4pt}
\section{Experiments}\label{sec:experiments}
\vspace{-8pt}
\xhdr{Experimental Settings}
Our pipeline is domain-agnostic, generating a benchmark of multi-turn conversation pairs through the same generation, verification, and ranking components from any corpus of knowledge-intensive context. We instantiate it on three domains, each viewable as a standalone benchmark: machine learning (CS papers from RPC-Bench~\citep{chen2026rpc}, open-access no license), biomedicine (PubMedQA~\citep{jin2019pubmedqa}, MIT license), and finance (S\&P~500 10-K filings~\cite{huggingface}, MIT license). Per pair, the user behavior and assistant failure types are sampled independently and uniformly. All generator, verifier, and judge calls are routed through OpenRouter~\cite{openrouter} with \texttt{gpt-5.5} used for generation and verification. Every judge in our registry supports thinking, which we leave at the provider's default. We exclude post-trained judges to keep the focus on inherent abilities of general-purpose LLMs, though post-training can improve performance on static problems~\citep{whitehouse2025j1, huang2026think}.

\begin{figure}[t]
    \centering
    \includegraphics[width=\textwidth]{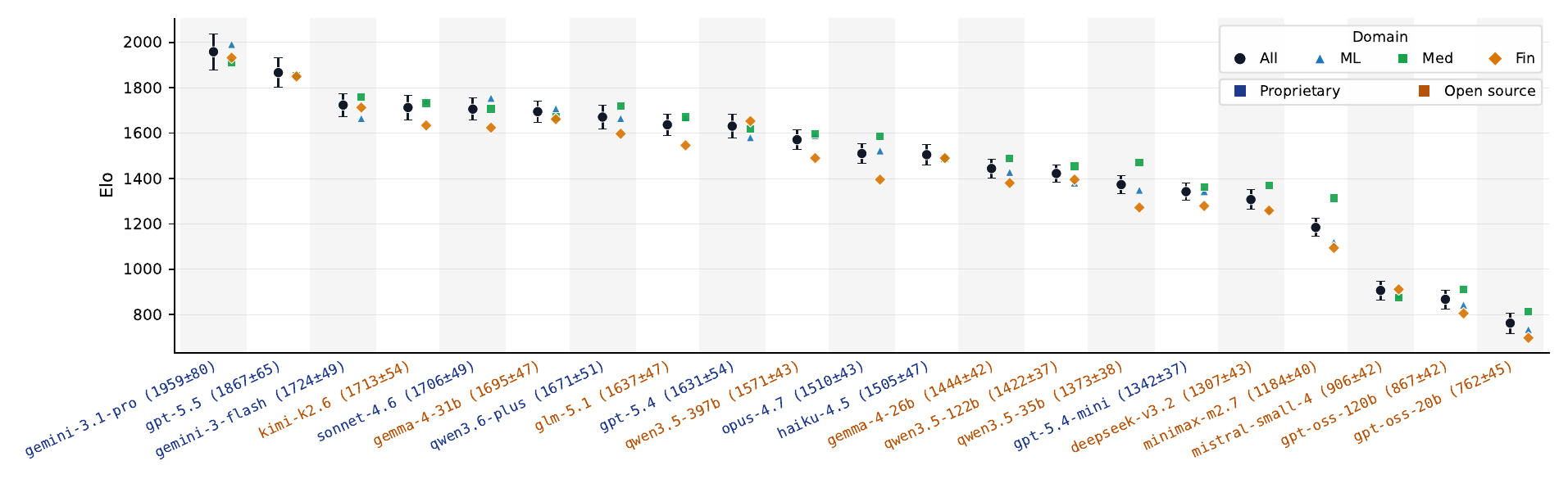}
    \caption{Elo scores of $21$ judges on the combined dataset. Black circles give the combined Elo with $95\%$ CI; colored markers show per-domain Elo scores. Tick-label color denotes proprietary (blue) vs.\ open-source (orange) judges. Table~\ref{tab:judge-bt-scores} in Appendix~\ref{sec:judge_scores} gives detailed per-domain results.}
    \label{fig:judge_rankings}
    \vspace{-14pt}
\end{figure}

\vspace{-4pt}
\xhdr{RankJudge Leaderboard}
Figure~\ref{fig:judge_rankings} reports BT-Elo ratings for $21$ judges. The scale spans nearly $1200$ Elo points from \texttt{gemini-3.1-pro} ($1959$) to \texttt{gpt-oss-20b} ($762$), implying that the benchmark separates judges across a wide range of capability. \texttt{gemini-3.1-pro} leads with a $92$-point margin over the next judge \texttt{gpt-5.5}, while the per-domain rankings broadly preserve the order, indicating that judge skill is largely domain-invariant. The leaderboard shows a trend that proprietary LLMs are still stronger than open-source models, although \texttt{kimi-k2.6} and \texttt{gemma-4-31b} outrank several frontier models, while \texttt{opus-4.7} underperforms, indicating that strong generation capability does not directly translate to strong judging. 

We note that \texttt{gpt-5.5} is subject to a strong self-preferential bias in pairwise judging, since this model was used for generation and verification. Appendix~\ref{sec:model_bias} characterizes the effect by regenerating the conversation pairs with \texttt{opus-4.7} as the generator; in this case \texttt{gpt-5.5} drops from rank $2$ to rank $7$ and, \texttt{opus-4.7} climbs from rank $11$ to rank $3$, but the rankings of the remaining $19$ judges are preserved almost identically (Spearman $\rho = 0.979$, Kendall $\tau = 0.906$). \texttt{gemini-3.1-pro} in particular ranks first under either generator despite, reflecting its genuinely superior judging ability.

\begin{figure}[!t]
    \centering
    \includegraphics[width=0.85\textwidth]{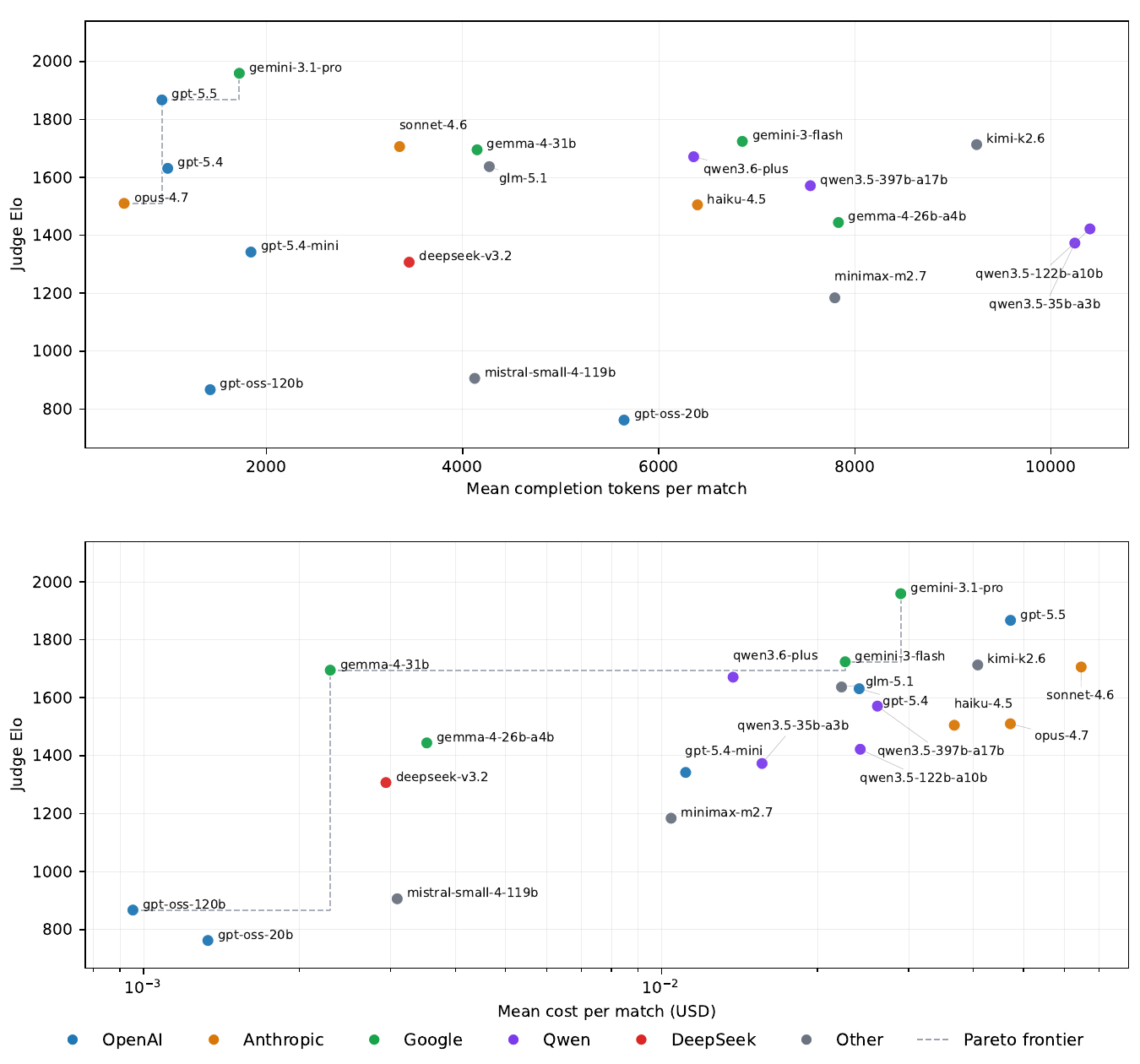}
    \caption{Judge Elo against per-match compute. Top: mean completion tokens per match (linear). Bottom: mean cost per match in USD (log). Markers are coloured by provider and the dashed grey step line traces the empirical Pareto frontier.}
    \label{fig:elo_vs_compute}
    \vspace{-8pt}
\end{figure}

\vspace{-4pt}
\xhdr{Elo vs.\ Compute}
\label{sec:elo_vs_compute}
Compute cost is the other consideration for determining the best judge. Figure~\ref{fig:elo_vs_compute} relates each judge's Elo to its mean per-match compute. The token-axis frontier collapses to three judges, \texttt{opus-4.7} ($549$ tokens, $1510$ Elo), \texttt{gpt-5.5} ($934$ tokens, $1867$), and \texttt{gemini-3.1-pro} ($1723$ tokens, $1959$); a long tail of mid-Elo judges spends $7$--$10$k tokens per match while sitting well off the frontier, so verbosity alone does not buy ranking accuracy. The cost-axis frontier tells a different story, with open-weights judges holding most of the curve: \texttt{gemma-4-31b} reaches $1695$ Elo at \$0.0023 per match, less than a tenth the price of \texttt{gemini-3.1-pro} for $264$ fewer Elo points, and \texttt{gpt-oss-120b} extends the frontier into the sub-cent regime at the bottom of the leaderboard. Only \texttt{gemini-3.1-pro} lies on both frontiers; \texttt{gpt-5.5} is dominated on cost, so budget-bound users can reach competitive Elo at a fraction of the spend via \texttt{gemini-3-flash} or an open-weights judge such as \texttt{gemma-4-31b}.

\begin{figure}[!t]
    \centering
    \includegraphics[width=\textwidth]{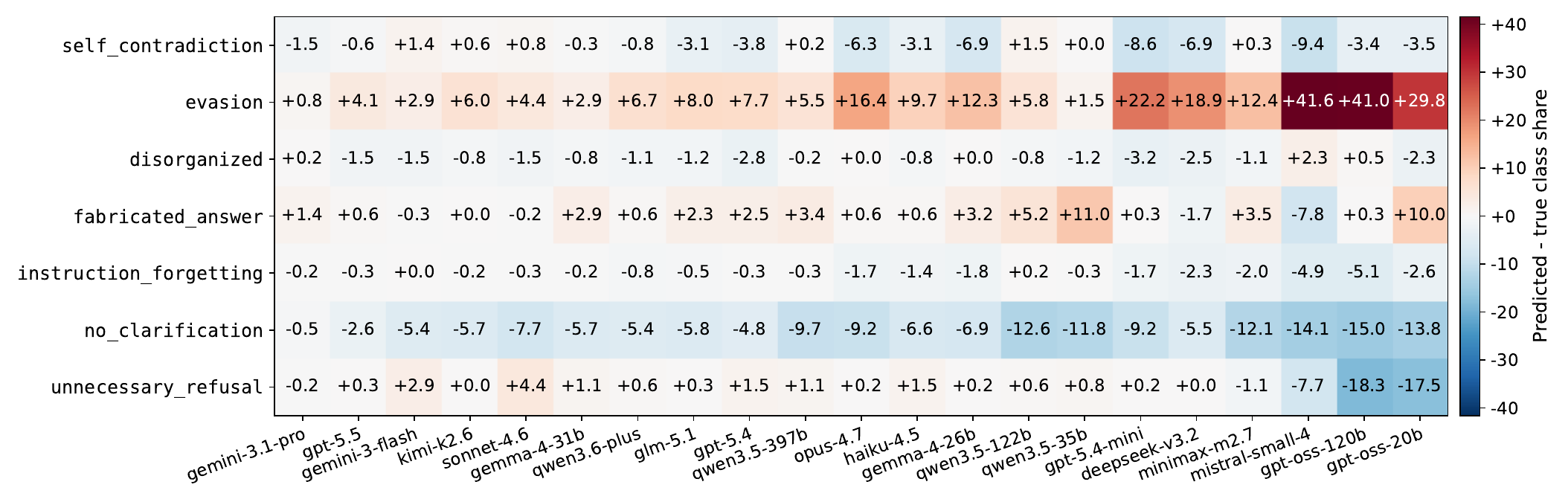}
    \caption{Per-class prediction bias for each judge. Each cell gives the difference in percentage points between the judge's predicted share of the assistant failure type and the ground-truth share. Red indicates over-prediction, while blue shows under-prediction. Columns are sorted by Elo.}
    \label{fig:class_bias}
    \vspace{-8pt}
\end{figure}

\vspace{-4pt}
\xhdr{Class Bias is a Capability Gap}
Figure~\ref{fig:class_bias} reports the per-judge gap between predicted and ground-truth failure type share. As judge strength declines, predictions collapse onto \texttt{evasion} with over-prediction rising from near zero for the strongest judges to $+41.6$pp at the bottom of the leaderboard, leading to under-prediction of \texttt{no\_clarification} and \texttt{unnecessary\_refusal}, which fall to roughly $-15$pp and $-18$pp for the weakest judges. No such bias is observed at the top, where mistakes by \texttt{gemini-3.1-pro} remain within $2$pp of zero on every class, which establishes that the seven-way failure taxonomy is internally non-overlapping. The class bias is therefore a property of the weaker judges rather than of the data. With poor reasoning ability, judges default to \texttt{evasion} as a crutch.
The same collapse is reproduced in each domain (Figure~\ref{fig:class_bias_domains} in Appendix~\ref{sec:class_bias_domains}), and the per-judge confusion matrix (Figure~\ref{fig:confusion_matrix} in Appendix~\ref{sec:confusion_matrix}) corroborates this at the prediction level.

\begin{figure}[!t]
    \centering
    \includegraphics[width=0.9\textwidth]{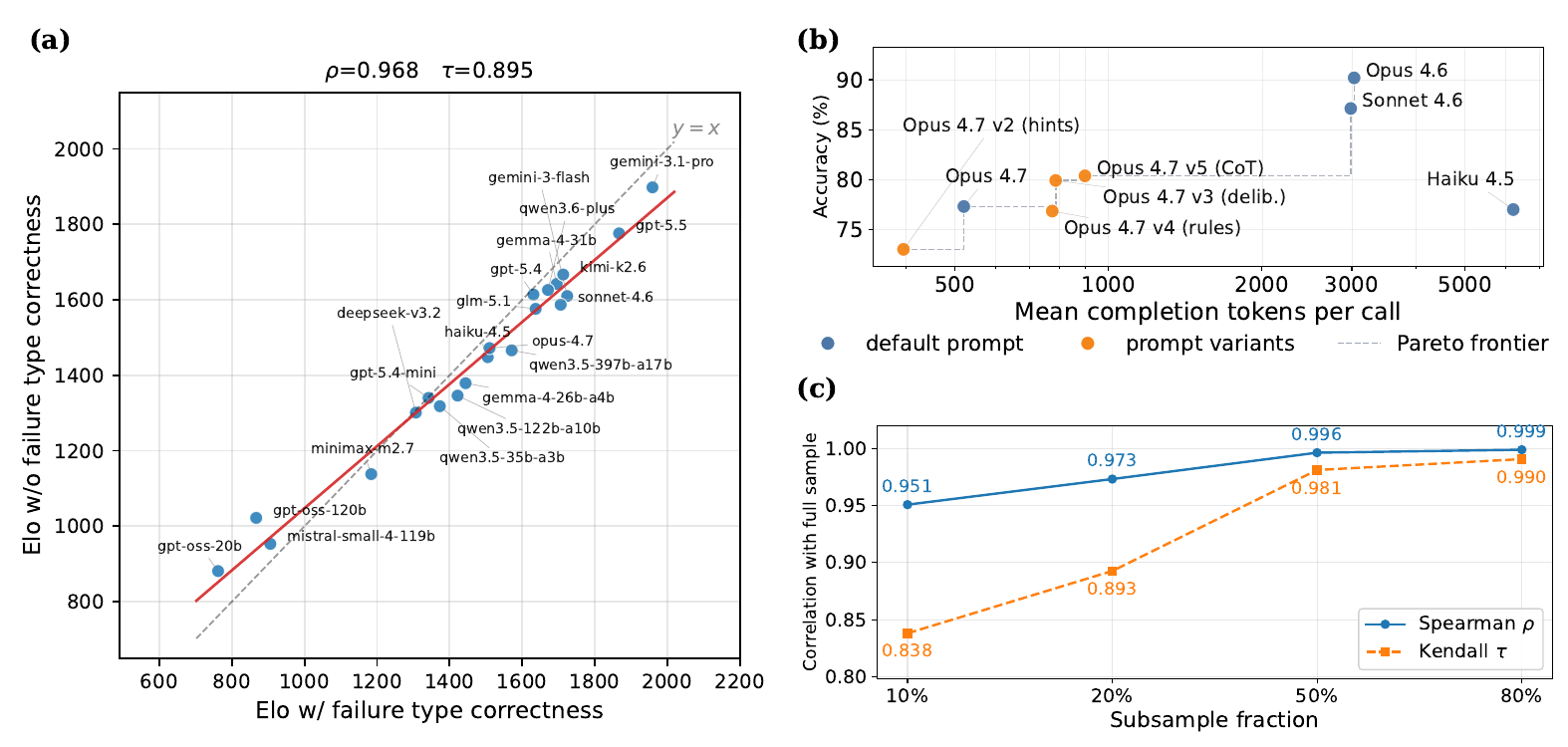}
    \caption{(a) Judge Elo with and without failure-type correctness as a criterion. (b) Accuracy vs.\ mean completion tokens for four default-prompt baselines and four \texttt{opus-4.7} prompt revisions; the dashed step is the empirical Pareto frontier. (c) Rank correlation between the full-sample judge ranking and rankings recomputed on uniform subsamples.}
    \label{fig:type_mitigation_sample}
    \vspace{-18pt}
\end{figure}

\vspace{-4pt}
\xhdr{Effect of Failure-type Prediction}
We drop the failure-type component from the correctness criterion leaving $(y^\star, r^\star)$, and refit BT-Elo. Panel~(a) of Figure~\ref{fig:type_mitigation_sample} compares judge Elo with and without the type label $t^\star$, showing high correlation at Spearman $\rho = 0.968$ and Kendall $\tau = 0.895$. 
This shows that including failure type prediction as a correctness criterion widens the Elo gap between judges without reordering the leaderboard. Hence, the failure taxonomy further distinguishes judge ability by testing their ability to understand the reasons for failure. The same pattern holds within each domain (Figure~\ref{fig:type_correctness_domains} in Appendix~\ref{sec:type_correctness_domains}).

\vspace{-4pt}
\xhdr{Error Analysis and Mitigation}
The leaderboard exposes a surprising anomaly: \texttt{opus-4.7}, the headline model of the Claude family at the time of writing, sits below \texttt{sonnet-4.6}. This motivates a deeper investigation, including whether prompt-level intervention can recover the gap. We repurpose the human audit of \Cref{sec:dataset_curation} for error-analysis over \texttt{opus-4.7}'s incorrect verdicts on the \texttt{clean} pairs, summarize its recurring failure modes, and design four prompt revisions targeting them: category hints (v2), explicit candidate comparison (v3), decision rules (v4), and chain-of-thought (v5), detailed in Appendix~\ref{sec:prompt_judge}. To bracket the magnitude of the gap we additionally ran \texttt{opus-4.6} under the default prompt. Figure~\ref{fig:type_mitigation_sample}(b) plots accuracy against mean completion tokens per call for the four revisions and the four default-prompt baselines. \texttt{opus-4.6} substantially outperforms every variant of \texttt{opus-4.7} we tried, and none of the prompt revisions improves \texttt{opus-4.7} to the accuracy level of \texttt{opus-4.6} or \texttt{sonnet-4.6}, suggesting a capability ceiling rather than a prompting deficiency. The shift aligns with documented changes in the \texttt{opus-4.7} release~\citep{anthropic2026opus47}: thinking is now an adaptive effort that the model can downscale at will, in place of 4.6's fixed budget. As a result \texttt{opus-4.7} commits to verdicts with too little deliberation when it could benefit from using more test-time compute. The Claude family case study suggests a more general conclusion that prompt-level intervention cannot close a model-capability gap in pairwise judging. Our four revisions cover the usual toolkit (more guidance, more deliberation, more structure, more thinking), and their joint failure suggests that practitioners are better served by switching judges than by tuning prompts.

\vspace{-4pt}
\xhdr{Robustness and Benchmark Cost}
To verify the partial-observability argument of Section~\ref{sec:eval_protocol}, we refit BT-Elo on uniform subsamples of the judgment results at fractions $\{10\%, 20\%, 50\%, 80\%\}$ and report the rank correlation with the full data in Figure~\ref{fig:type_mitigation_sample}(c). Spearman $\rho$ exceeds $0.95$ from $10\%$ onward and Kendall $\tau$ crosses the same threshold at $50\%$, so the leaderboard can be reproduced at a fraction of the full cost with negligible distortion to rankings. The same conclusion holds within each domain (Figure~\ref{fig:sample_size_domains} in Appendix~\ref{sec:sample_size_domains}). The leaderboard is also stable to two further perturbations. First, refitting with the top-5\% Elo conversation pairs included matches the published ordering at Spearman $\rho \geq 0.997$, with the top-$10$ judge ranking preserved (Figure~\ref{fig:noise_reduce} in Appendix~\ref{sec:noise_reduce}), confirming that the BT rating is robust under residual noise. Second, swapping BT-Elo for the difficulty-aware EIP ranking leaves the ordering nearly unchanged ($\rho \geq 0.94$, Appendix~\ref{sec:ranking_method}). We adopt BT as the primary score because it yields closed-form cluster-robust error bars and a leaderboard that stays stable under partial observability.

\vspace{-4pt}
\xhdr{Pairwise versus Pointwise}
\textsc{RankJudge} can also be run with the judges scoring conversations pointwise on a $1$--$10$ Likert scale rather than choosing between candidates. The two protocols agree at a coarse level (Spearman $\rho = 0.809$) but the judge rankings diverge because these approaches measure different capabilities. Pairwise tests discrimination between two candidate conversations, while pointwise tests absolute calibration of conversation quality without a reference. We report this comparison in detail in Appendix~\ref{sec:pointwise}.
\vspace{-4pt}
\section{Conclusion}
\vspace{-4pt}
We presented \textsc{RankJudge}, a benchmark generator for evaluating LLM-as-a-judge on multi-turn, reference-grounded conversations. By constructing each pair to have an unambiguously worse conversation while keeping the subject material similarly challenging, but not identical, the pipeline supports a strict joint correctness criterion that separates correct judgments from correct guesses. Three-layer automated verification combined with dynamic Bradley--Terry Elo-based curation keeps construction fully synthetic without sacrificing label quality. The verification step and label quality was cross-checked by a human audit and a held-out fine-tuning experiment.  Across $21$ frontier judges and three subject matter domains, the resulting leaderboard spans nearly $1200$ Elo points, surfaces several open-weight checkpoints that outrank proprietary models, and exposes a sharp failure type bias in weaker judges. The judge ranking is stable across different curation setups, swapping the Bradley--Terry rating system for the EIP system, and under subsampling which reduces the cost of benchmarking.

\bibliographystyle{unsrt}
\bibliography{main-cites}

\clearpage
\appendix

\section{Additional Results}

\subsection{Detailed Judge Elo Ratings}
\label{sec:judge_scores}
Table~\ref{tab:judge-bt-scores} gives each judge's combined Elo together with the per-domain breakdown, where confidence intervals are obtained from the cluster-robust sandwich estimator of Appendix~\ref{sec:bt_details}. The intervals tighten toward the middle of the leaderboard, where contested matches concentrate, and widen at the extremes, where the strongest and weakest judges win or lose decisively against most opponents and so accumulate little discriminative information. Finance is the lowest-Elo domain for most mid-leaderboard judges, consistent with the heavier numerical reasoning load imposed by 10-K filings; \texttt{gpt-5.4}, the only judge whose Finance Elo exceeds its ML and Biomedical scores, is a notable exception.
\begin{table}[htbp]
\centering
\caption{Elo of LLM judges (95\% CI). Best per column in bold. Sorted by Elo measured by the combined dataset (All).}
\label{tab:judge-bt-scores}
\setlength{\tabcolsep}{6pt}
\begin{tabular}{lccccl}
\toprule
Judge & All & ML & Med & Fin & Type \\
\midrule
\texttt{gemini-3.1-pro} & \textbf{1959\,\textpm\,80} & \textbf{1992\,\textpm\,156} & \textbf{1910\,\textpm\,131} & \textbf{1933\,\textpm\,129} & Prop. \\
\texttt{gpt-5.5} & 1867\,\textpm\,65 & 1854\,\textpm\,109 & 1852\,\textpm\,119 & 1850\,\textpm\,107 & Prop. \\
\texttt{gemini-3-flash} & 1724\,\textpm\,49 & 1665\,\textpm\,81 & 1759\,\textpm\,93 & 1713\,\textpm\,83 & Prop. \\
\texttt{kimi-k2.6} & 1713\,\textpm\,54 & 1743\,\textpm\,99 & 1732\,\textpm\,93 & 1634\,\textpm\,89 & Open \\
\texttt{sonnet-4.6} & 1706\,\textpm\,49 & 1755\,\textpm\,88 & 1707\,\textpm\,90 & 1624\,\textpm\,80 & Prop. \\
\texttt{gemma-4-31b} & 1695\,\textpm\,47 & 1708\,\textpm\,81 & 1671\,\textpm\,87 & 1662\,\textpm\,76 & Open \\
\texttt{qwen3.6-plus} & 1671\,\textpm\,51 & 1665\,\textpm\,88 & 1719\,\textpm\,103 & 1597\,\textpm\,79 & Prop. \\
\texttt{glm-5.1} & 1637\,\textpm\,47 & 1665\,\textpm\,87 & 1671\,\textpm\,89 & 1546\,\textpm\,73 & Open \\
\texttt{gpt-5.4} & 1631\,\textpm\,54 & 1581\,\textpm\,90 & 1617\,\textpm\,83 & 1653\,\textpm\,99 & Prop. \\
\texttt{qwen3.5-397b-a17b} & 1571\,\textpm\,43 & 1590\,\textpm\,74 & 1597\,\textpm\,80 & 1490\,\textpm\,72 & Open \\
\texttt{opus-4.7} & 1510\,\textpm\,43 & 1523\,\textpm\,75 & 1587\,\textpm\,82 & 1395\,\textpm\,71 & Prop. \\
\texttt{haiku-4.5} & 1505\,\textpm\,47 & 1492\,\textpm\,80 & 1487\,\textpm\,78 & 1490\,\textpm\,82 & Prop. \\
\texttt{gemma-4-26b-a4b} & 1444\,\textpm\,42 & 1428\,\textpm\,70 & 1487\,\textpm\,75 & 1380\,\textpm\,74 & Open \\
\texttt{qwen3.5-122b-a10b} & 1422\,\textpm\,37 & 1381\,\textpm\,60 & 1454\,\textpm\,68 & 1395\,\textpm\,62 & Open \\
\texttt{qwen3.5-35b-a3b} & 1373\,\textpm\,38 & 1349\,\textpm\,65 & 1471\,\textpm\,73 & 1272\,\textpm\,63 & Open \\
\texttt{gpt-5.4-mini} & 1342\,\textpm\,37 & 1343\,\textpm\,62 & 1363\,\textpm\,73 & 1279\,\textpm\,61 & Prop. \\
\texttt{deepseek-v3.2} & 1307\,\textpm\,43 & 1260\,\textpm\,69 & 1370\,\textpm\,85 & 1259\,\textpm\,72 & Open \\
\texttt{minimax-m2.7} & 1184\,\textpm\,40 & 1119\,\textpm\,68 & 1313\,\textpm\,70 & 1094\,\textpm\,70 & Open \\
\texttt{mistral-small-4-119b} & 906\,\textpm\,42 & 887\,\textpm\,68 & 873\,\textpm\,80 & 911\,\textpm\,76 & Open \\
\texttt{gpt-oss-120b} & 867\,\textpm\,42 & 842\,\textpm\,67 & 911\,\textpm\,72 & 805\,\textpm\,80 & Open \\
\texttt{gpt-oss-20b} & 762\,\textpm\,45 & 734\,\textpm\,73 & 812\,\textpm\,81 & 697\,\textpm\,81 & Open \\
\bottomrule
\end{tabular}
\end{table}

\subsection{Problem Rankings}
\label{sec:problem_rankings}
\begin{figure}[htbp]
    \centering
    \includegraphics[width=\textwidth]{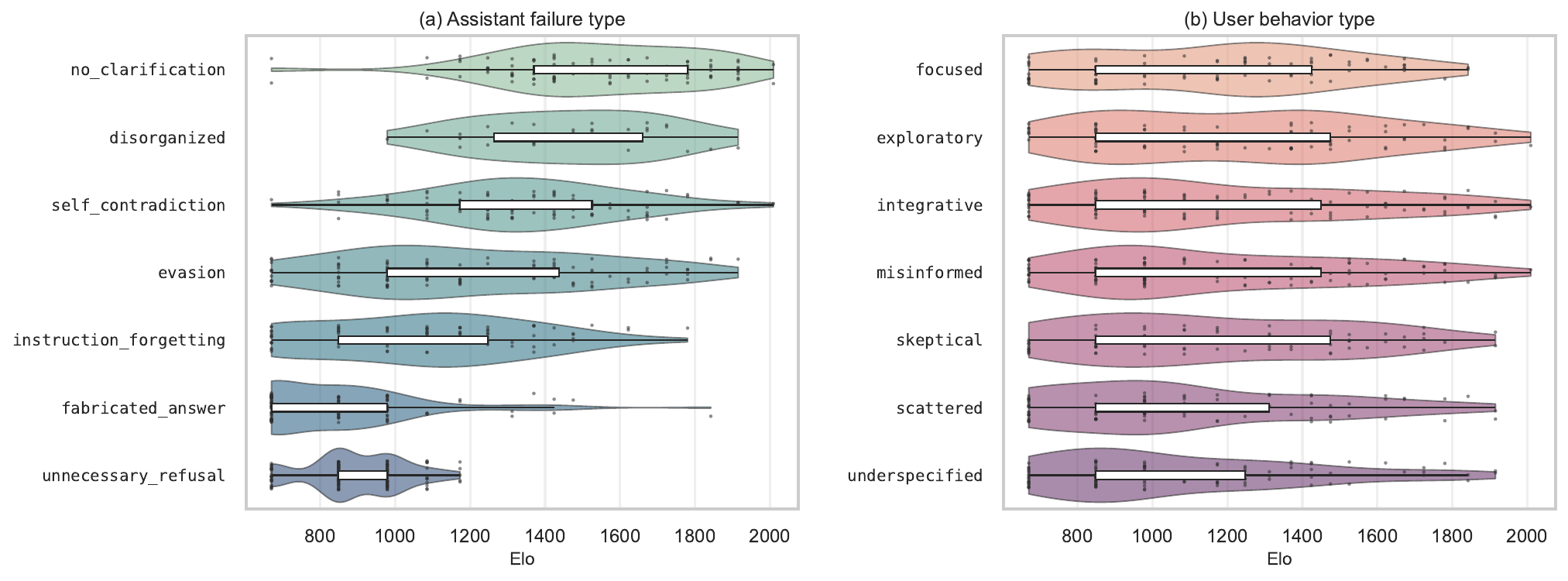}
    \caption{Per-pair Elo on the combined dataset, grouped by (a) assistant failure type and (b) user behavior type. Categories are sorted by median Elo; violins show the full distribution, overlaid boxes mark the interquartile range (IQR) with the median in red, and jittered points mark individual pairs.}
    \label{fig:problem_rankings}
\end{figure}

Assistant failure type separates pair difficulty cleanly. \texttt{no\_clarification} (median Elo $1525$) and \texttt{disorganized} ($1475$) consistently beat judges, while \texttt{unnecessary\_refusal} and \texttt{fabricated\_answer} (both $849$) are the easiest to catch: refusals stand out in context and fabrications are directly checkable against the grounding document. User behavior, by contrast, barely shifts pair difficulty: the seven categories span only $979$ to $1247$ in median Elo with heavily overlapping IQRs. This is by design. The user behavior axis is included to diversify the conversational contexts in the benchmark rather than to differentiate judges, so the lack of separation here is the expected outcome.

\subsection{Model Bias and Self-Preference}
\label{sec:model_bias}
\begin{figure}[htbp]
    \centering
    \includegraphics[width=\textwidth]{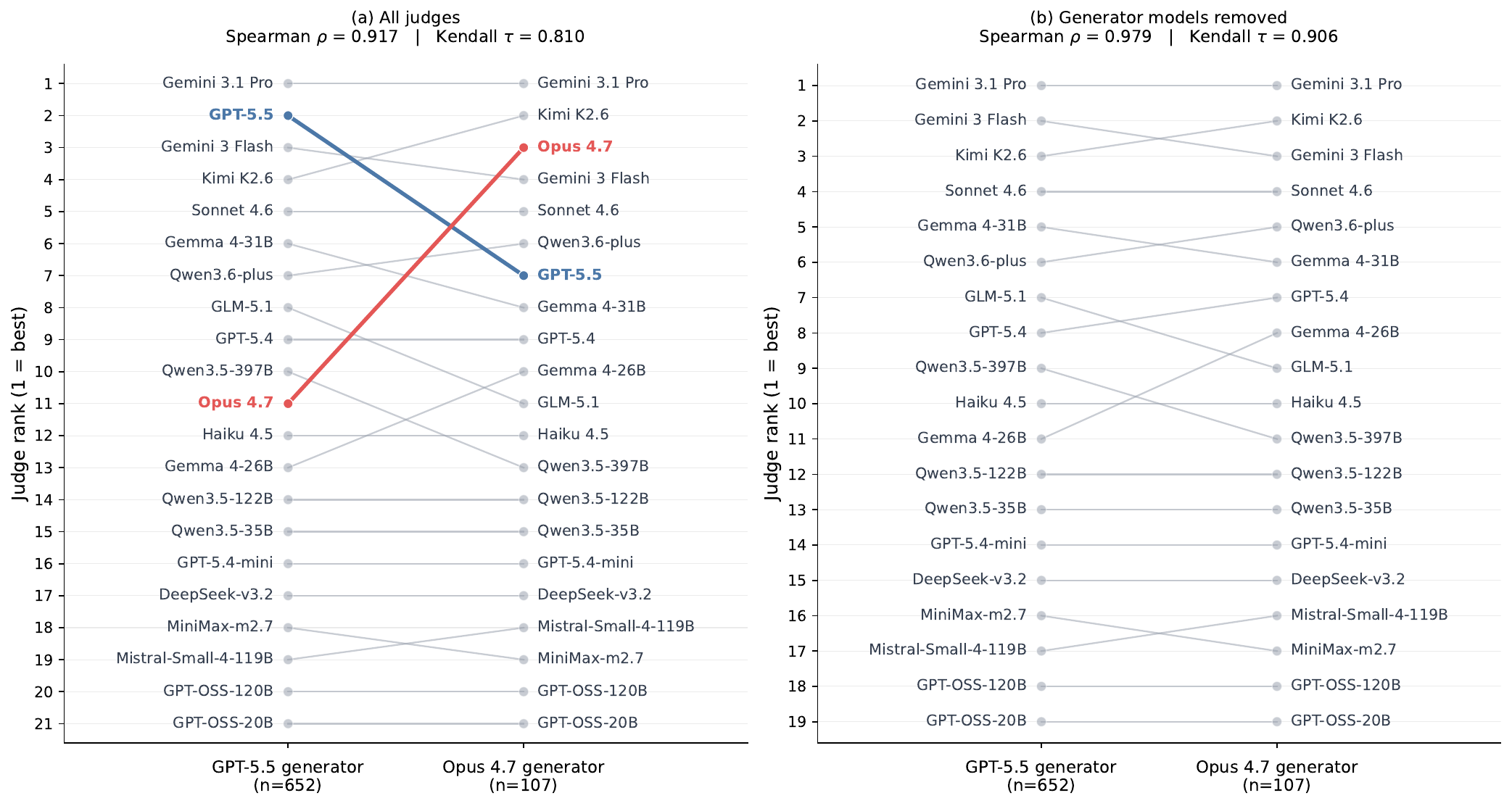}
    \caption{Same-model preference does not distort the rest of the leaderboard. Each panel is a slope chart of judge Elo ranks under the published pair set generated by \texttt{gpt-5.5} (left axis) against the parallel set generated by \texttt{opus-4.7} (right axis), with ranks recomputed from a Bradley--Terry fit within the panel's judge subset. }
    \label{fig:model_bias}
\end{figure}

\paragraph{Setup.} To isolate the effect of the generator we re-ran the pair-generation pipeline a second time with \texttt{opus-4.7} substituted for \texttt{gpt-5.5} as the generator and the rest of the recipe held fixed: the same source documents, the dual-conditioning prompts of Appendix~\ref{sec:prompt_generation}, the coherence, adherence, and grounding gates of Appendix~\ref{sec:prompt_verification}, and the same 21-judge panel evaluated under the pairwise prompt of Appendix~\ref{sec:prompt_judge}. To bound API cost the parallel set was restricted to 50 items per domain, drawn from the same source pool used for the published evaluation, yielding 150 candidate pairs of which 131 survive verification. Bradley--Terry is fit independently on each condition's match set with the MM iteration of Appendix~\ref{sec:bt_details}:  on the published \texttt{top\_removed} slice (652 pairs, after dropping the \texttt{ambiguous} and \texttt{noise} pairs and the top-Elo trim), and on the filtered slice (107 non-trivial pairs). Since the parallel set has no human audit it cannot be cut to the \texttt{clean} subset. Spearman $\rho$ and Kendall $\tau$ in each panel are computed on judge ranks computed within that panel's subset, so they are invariant to the absolute Elo scale and to which judges are removed.

\paragraph{Findings.} As shown in Figure~\ref{fig:model_bias}, panel (a) keeps all 21 judges and exposes a clear self-preference: \texttt{gpt-5.5} slips from rank 2 to rank 7 once it is no longer the generator, and \texttt{opus-4.7} climbs from rank 11 to rank 3 once it is. The other 19 judges, whose outputs are never used to construct pairs, are nevertheless ordered almost identically across the two conditions. \texttt{gemini-3.1-pro} is in particular ranked first under both generators, even though no condition uses its outputs, suggesting that its lead is driven by judging ability rather than generator alignment. Panel (b) drops the two generators and recomputes the ranks on the same matches: Spearman $\rho$ rises from $0.917$ to $0.979$ and Kendall $\tau$ from $0.810$ to $0.906$. The bias is therefore localized to whichever model is producing the conversations; one generator still has to be chosen for the published evaluation, but that choice does not materially shift any other judge's ranking.

\subsection{Filtering Cascade}
\label{sec:filtering_cascade}

Tables~\ref{tab:dataset-stats-domain}, \ref{tab:dataset-stats-assistant}, and \ref{tab:dataset-stats-user} report the filtering cascade sliced three ways: by domain, by assistant failure type, and by user behavior type. Of $1200$ generated pairs, $652$ survive every layer (overall survival $54.3\%$). The dominant signal sits on the assistant-failure axis. \texttt{unnecessary\_refusal} and \texttt{fabricated\_answer} clear the adherence verifier with zero loss, while \texttt{disorganized} loses $130$ of its $174$ generated pairs at that single layer and ends up the rarest type in the published evaluation set ($30$ pairs, $17.2\%$ survival). The per-domain and per-user-behavior survival rates are far tighter and have no comparable bottleneck ($48.5$--$60.0\%$ across domains, $46.4$--$61.8\%$ across user behaviors), so the shape of the cascade is set almost entirely by which flaw is being injected. The next paragraph and worked example explain why \texttt{disorganized} is the hard case: it is the only assistant flaw whose detection condition is structural rather than categorical, so a turn must lack logical flow strongly enough for the verifier to confirm the flaw, but not so strongly that every judge picks the better conversation on sight.

\begin{table}[htbp]
\centering
\caption{Verification and filtering cascade, split by assistant failure type. }
\label{tab:dataset-stats-assistant}
\footnotesize
\setlength{\tabcolsep}{2pt}
\renewcommand{\arraystretch}{1.15}
\begin{tabular}{l|c|ccc|c>{\columncolor{black!8}}cc|c}
\toprule
Assistant Failure Type & Generated & Coherence & Adherence & Grounding & Uninformative & Human & Top 5\% & Survival \\
\midrule
\texttt{self\_contradiction} & 132 & 131 {\scriptsize($-1$)} & 117 {\scriptsize($-14$)} & 96 {\scriptsize($-21$)} & 94 {\scriptsize($-2$)} & 92 {\scriptsize($-2$)} & 88 {\scriptsize($-4$)} & 66.7\% \\
\texttt{evasion} & 165 & 165 {\scriptsize($-0$)} & 152 {\scriptsize($-13$)} & 126 {\scriptsize($-26$)} & 113 {\scriptsize($-13$)} & 111 {\scriptsize($-2$)} & 108 {\scriptsize($-3$)} & 65.5\% \\
\texttt{disorganized} & 174 & 174 {\scriptsize($-0$)} & 44 {\scriptsize($-130$)} & 34 {\scriptsize($-10$)} & 34 {\scriptsize($-0$)} & 33 {\scriptsize($-1$)} & 30 {\scriptsize($-3$)} & 17.2\% \\
\texttt{fabricated\_answer} & 180 & 180 {\scriptsize($-0$)} & 180 {\scriptsize($-0$)} & 156 {\scriptsize($-24$)} & 104 {\scriptsize($-52$)} & 104 {\scriptsize($-0$)} & 104 {\scriptsize($-0$)} & 57.8\% \\
\texttt{\scriptsize instruction\_forgetting} & 147 & 144 {\scriptsize($-3$)} & 143 {\scriptsize($-1$)} & 86 {\scriptsize($-57$)} & 80 {\scriptsize($-6$)} & 80 {\scriptsize($-0$)} & 79 {\scriptsize($-1$)} & 53.7\% \\
\texttt{no\_clarification} & 231 & 229 {\scriptsize($-2$)} & 224 {\scriptsize($-5$)} & 178 {\scriptsize($-46$)} & 137 {\scriptsize($-41$)} & 126 {\scriptsize($-11$)} & 103 {\scriptsize($-23$)} & 44.6\% \\
\texttt{unnecessary\_refusal} & 171 & 171 {\scriptsize($-0$)} & 171 {\scriptsize($-0$)} & 145 {\scriptsize($-26$)} & 141 {\scriptsize($-4$)} & 140 {\scriptsize($-1$)} & 140 {\scriptsize($-0$)} & 81.9\% \\
\midrule
\textbf{All} & 1200 & 1194 {\scriptsize($-6$)} & 1031 {\scriptsize($-163$)} & 821 {\scriptsize($-210$)} & 703 {\scriptsize($-118$)} & 686 {\scriptsize($-17$)} & 652 {\scriptsize($-34$)} & 54.3\% \\
\bottomrule
\end{tabular}
\end{table}

\begin{table}[htbp]
\centering
\caption{Verification and filtering cascade, split by user behavior type.}
\label{tab:dataset-stats-user}
\footnotesize
\setlength{\tabcolsep}{2pt}
\renewcommand{\arraystretch}{1.15}
\begin{tabular}{l|c|ccc|c>{\columncolor{black!8}}cc|c}
\toprule
User Behavior Type & Generated & Coherence & Adherence & Grounding & Uninformative & Human & Top 5\% & Survival \\
\midrule
\texttt{focused} & 150 & 148 {\scriptsize($-2$)} & 140 {\scriptsize($-8$)} & 115 {\scriptsize($-25$)} & 93 {\scriptsize($-22$)} & 90 {\scriptsize($-3$)} & 87 {\scriptsize($-3$)} & 58.0\% \\
\texttt{integrative} & 222 & 222 {\scriptsize($-0$)} & 192 {\scriptsize($-30$)} & 139 {\scriptsize($-53$)} & 117 {\scriptsize($-22$)} & 110 {\scriptsize($-7$)} & 103 {\scriptsize($-7$)} & 46.4\% \\
\texttt{scattered} & 144 & 142 {\scriptsize($-2$)} & 126 {\scriptsize($-16$)} & 103 {\scriptsize($-23$)} & 92 {\scriptsize($-11$)} & 90 {\scriptsize($-2$)} & 89 {\scriptsize($-1$)} & 61.8\% \\
\texttt{skeptical} & 138 & 138 {\scriptsize($-0$)} & 123 {\scriptsize($-15$)} & 94 {\scriptsize($-29$)} & 80 {\scriptsize($-14$)} & 79 {\scriptsize($-1$)} & 77 {\scriptsize($-2$)} & 55.8\% \\
\texttt{misinformed} & 174 & 173 {\scriptsize($-1$)} & 141 {\scriptsize($-32$)} & 132 {\scriptsize($-9$)} & 122 {\scriptsize($-10$)} & 119 {\scriptsize($-3$)} & 107 {\scriptsize($-12$)} & 61.5\% \\
\texttt{exploratory} & 174 & 173 {\scriptsize($-1$)} & 151 {\scriptsize($-22$)} & 111 {\scriptsize($-40$)} & 89 {\scriptsize($-22$)} & 88 {\scriptsize($-1$)} & 83 {\scriptsize($-5$)} & 47.7\% \\
\texttt{underspecified} & 198 & 198 {\scriptsize($-0$)} & 158 {\scriptsize($-40$)} & 127 {\scriptsize($-31$)} & 110 {\scriptsize($-17$)} & 110 {\scriptsize($-0$)} & 106 {\scriptsize($-4$)} & 53.5\% \\
\midrule
\textbf{All} & 1200 & 1194 {\scriptsize($-6$)} & 1031 {\scriptsize($-163$)} & 821 {\scriptsize($-210$)} & 703 {\scriptsize($-118$)} & 686 {\scriptsize($-17$)} & 652 {\scriptsize($-34$)} & 54.3\% \\
\bottomrule
\end{tabular}
\end{table}

\paragraph{Why \texttt{disorganized} struggles at adherence.} Refusals, fabrications, contradictions, and instruction violations are binary properties of a turn's content: a turn either declines an answerable question or it does not. \texttt{disorganized} is different. The flaw lives in the ordering and foregrounding of \emph{correct} content, not in the content itself, so the bad turn must reshuffle the same facts that the good turn presents cleanly. Make the reshuffle too mild and the adherence verifier rejects the pair, with rejection reasons like ``\textit{Round 2 is mostly clear and foregrounds the main correction, so it does not actually exhibit the required disorganized flaw}''.\footnote{Verbatim from the \texttt{adherence.bad\_issue} field of a rejected \texttt{disorganized} pair in \texttt{verification.json}; analogous strings (``clear and logically organized'', ``gives a clear lead answer'') account for the bulk of the $130$-pair attrition in row \texttt{disorganized} of Table~\ref{tab:dataset-stats-assistant}.} Make the reshuffle too aggressive and the bad turn either drops or distorts a fact, which a strong judge then catches as a fabrication or evasion rather than as disorganization, and the pair is removed at the informativeness, human, or top-Elo layer.

The example below is from pair \texttt{id}=\texttt{40589ed7f7f0} (\textsc{ml}, \texttt{user\_behavior\_type}=\texttt{misinformed}, \texttt{assistant\_behavior\_type}=\texttt{disorganized}, \texttt{bad\_round\_index}=$2$). The user's second turn pushes a misinformed claim that the paper's main novelty was DP-Adam beating DP-SGD, that no DP-SGD baseline was reported, and that prior work had privately fine-tuned GPT-2 with DP-SGD. Both assistant responses contain the same three corrections: the E2E DP-SGD BLEU number ($63.17$), that DP-Adam was not claimed as the contribution, and that the cited GPT-2 prior work fine-tuned GPT-2 non-privately. The only difference is structural: \texttt{convo\_a} leads with the answer to the user's question (``DP-Adam was not claimed as a new algorithm''), while \texttt{convo\_b} opens with a peripheral hyperparameter detail and threads the lead correction through a privacy-by-post-processing aside, leaving it half a paragraph in.

\noindent\begin{minipage}[t]{0.48\textwidth}
\begin{tcolorbox}[colback=gray!5, colframe=gray!50!black, title=Clean, fonttitle=\bfseries, boxrule=0.4pt, left=4pt, right=4pt, top=3pt, bottom=3pt]
\small DP-Adam was not claimed as a new algorithm; it was already common in DP libraries, and it uses the same gradient privatization mechanism as DP-SGD before the Adam update. The paper's contribution was showing that full fine-tuning of large pretrained language models can work well under DP when done with suitable hyperparameters, aligned objectives, and the memory trick. Also, the referenced prior work did not privately fine-tune GPT-2: it fine-tuned GPT-2 non-privately and used a privately trained feedforward network for private experiments.
\end{tcolorbox}
\end{minipage}\hfill
\begin{minipage}[t]{0.48\textwidth}
\begin{tcolorbox}[colback=gray!5, colframe=gray!50!black, title=Disorganized, fonttitle=\bfseries, boxrule=0.4pt, left=4pt, right=4pt, top=3pt, bottom=3pt]
\small On E2E at epsilon~$=8$, their extra DP-SGD experiment used learning rate $3$ and clipping norm $0.1$ and reached BLEU $63.17$ over $3$ seeds, which was on par with DP-Adam. DP-Adam itself was not claimed as the contribution, even though after the same subsampled Gaussian gradient privatization, Adam-style updates preserve privacy by post-processing. The contribution was applying known DP optimization effectively to full language-model fine-tuning, with the memory trick making it computationally practical. Also, the cited GPT-2 work did not privately fine-tune GPT-2 with DP-SGD; it fine-tuned GPT-2 non-privately\dots
\end{tcolorbox}
\end{minipage}

\medskip
The two responses are factually equivalent against the source; only the ordering differs. This pair sits in the narrow band where the verifier accepts (the lead answer is genuinely buried, mid-paragraph, after a hyperparameter aside and a post-processing detour) and judges still split (the correct facts are all present, so a judge has to actually read for structure rather than skim for missing content). Most generation attempts miss this band in one direction or the other, which is why \texttt{disorganized} loses three quarters of its candidate pairs at the adherence layer alone.

\subsection{Class Bias by Domain}
\label{sec:class_bias_domains}
\begin{figure}[htbp]
    \centering
    \begin{subfigure}{\textwidth}
        \centering
        \includegraphics[width=\textwidth]{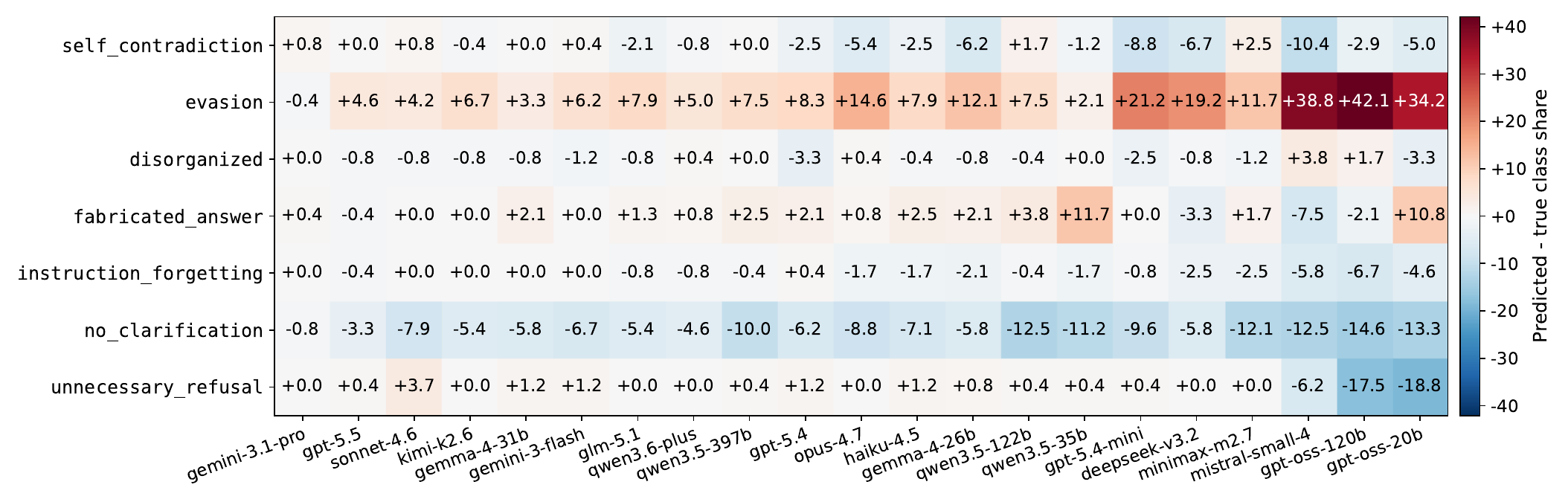}
        \caption{Machine Learning}
        \label{fig:class_bias_ml}
    \end{subfigure}\\[2pt]
    \begin{subfigure}{\textwidth}
        \centering
        \includegraphics[width=\textwidth]{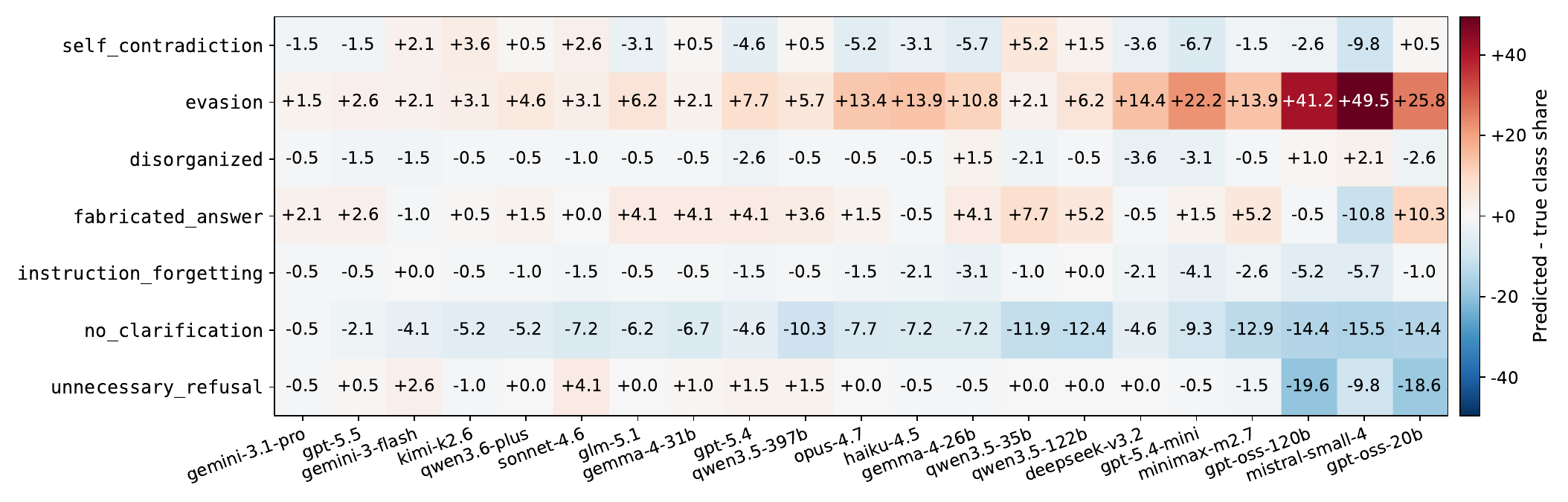}
        \caption{Biomedical}
        \label{fig:class_bias_med}
    \end{subfigure}\\[2pt]
    \begin{subfigure}{\textwidth}
        \centering
        \includegraphics[width=\textwidth]{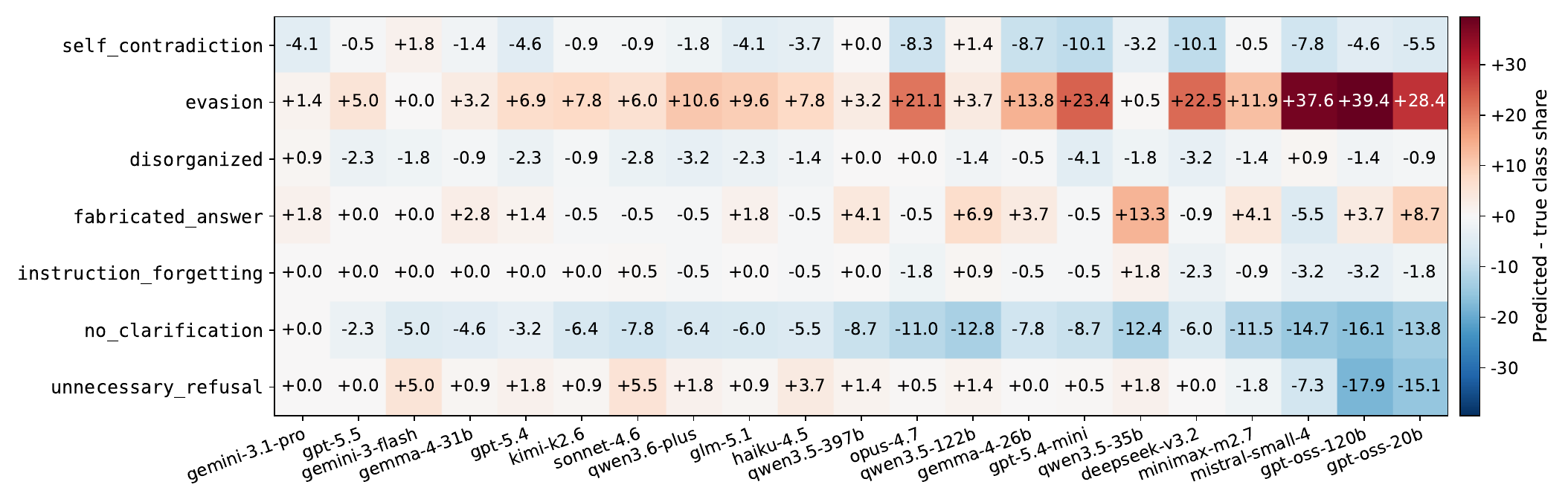}
        \caption{Finance}
        \label{fig:class_bias_fin}
    \end{subfigure}
    \caption{Per-class prediction bias for each judge across the three domains. Each cell gives the difference between the judge's predicted share of the assistant failure type and its ground-truth share, in percentage points; red indicates over-prediction and blue under-prediction. Columns are sorted by Elo.}
    \label{fig:class_bias_domains}
\end{figure}

Figure~\ref{fig:class_bias_domains} shows that the prediction-bias signature is essentially domain-invariant: across Machine Learning, Biomedical, and Finance, the same column pattern repeats, with \texttt{evasion} the dominant red column and \texttt{no\_clarification} the dominant blue one, and the magnitude of the bias growing as Elo decreases from left to right. The other five categories carry only small, noisy biases. This mirrors the confusion structure of Appendix~\ref{sec:confusion_matrix}: weak judges fold ambiguous failures into \texttt{evasion} and rarely commit to \texttt{no\_clarification}, regardless of the underlying domain.

\subsection{Assistant Failure Type Confusion}
\label{sec:confusion_matrix}
\begin{figure}[htbp]
    \centering
    \includegraphics[width=\textwidth]{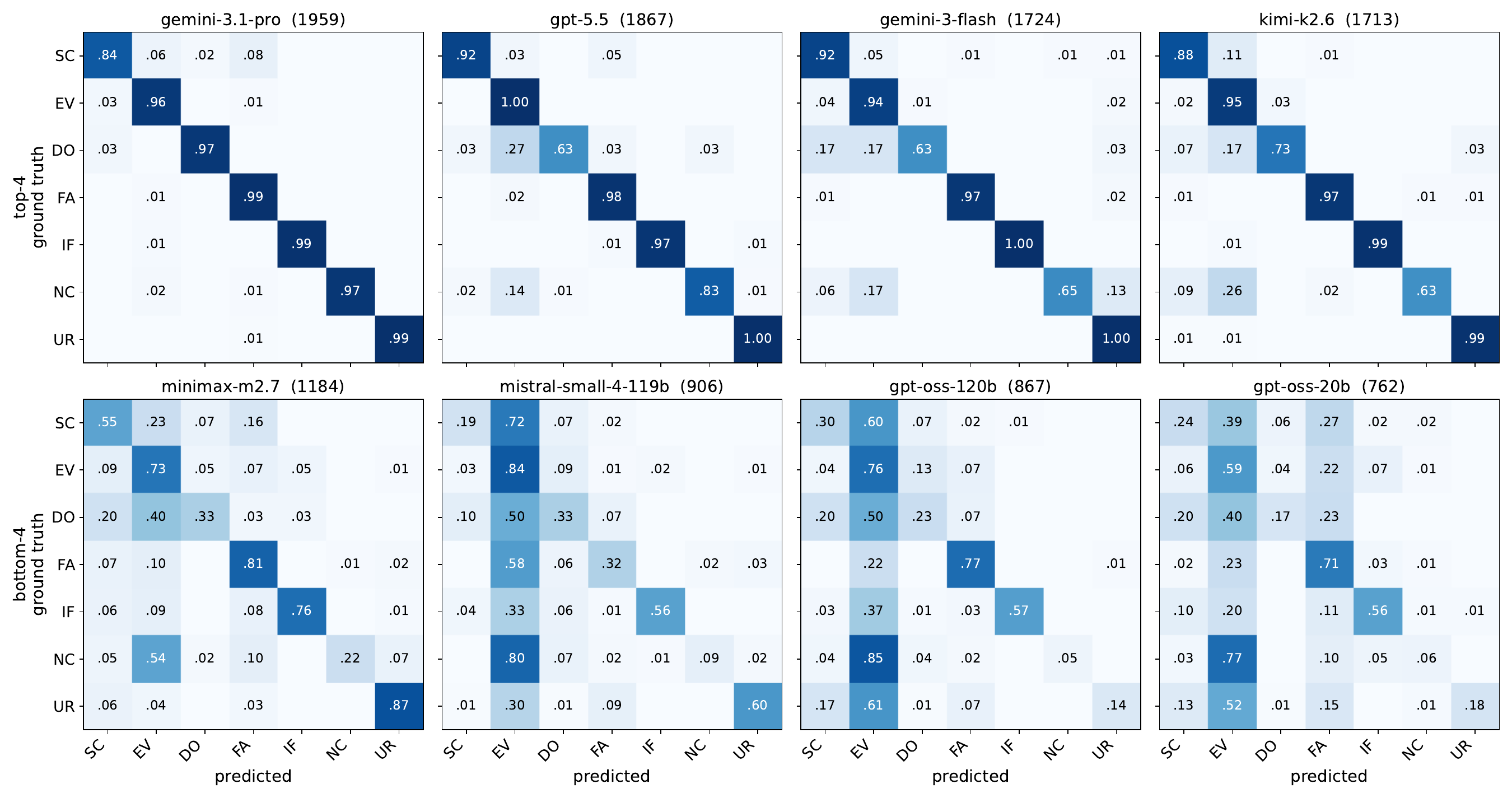}
    \caption{Per-judge confusion of assistant failure type predictions, row-normalized so each row sums to one. Top row: the four highest-Elo judges; bottom row: the four lowest. Class labels abbreviate the seven assistant failure types (SC \texttt{self\_contradiction}, EV \texttt{evasion}, DO \texttt{disorganized}, FA \texttt{fabricated\_answer}, IF \texttt{instruction\_forgetting}, NC \texttt{no\_clarification}, UR \texttt{unnecessary\_refusal}).}
    \label{fig:confusion_matrix}
\end{figure}

Figure~\ref{fig:confusion_matrix} contrasts how the four highest- and lowest-Elo judges classify the assistant failure type. The top row concentrates nearly all of its mass on the diagonal across most categories; the only residual confusion is on \texttt{disorganized} and \texttt{no\_clarification}, both of which leak modestly into \texttt{evasion}. The bottom row collapses toward a single dump bin: weak judges relabel a large share of \texttt{no\_clarification}, \texttt{self\_contradiction}, \texttt{disorganized}, and \texttt{unnecessary\_refusal} predictions as \texttt{evasion}, suggesting that taxonomy resolution, not failure detection, is what separates the leaderboard tail from its head.

\subsection{Failure-type Correctness by Domain}
\label{sec:type_correctness_domains}
\begin{figure}[htbp]
    \centering
    \includegraphics[width=\textwidth]{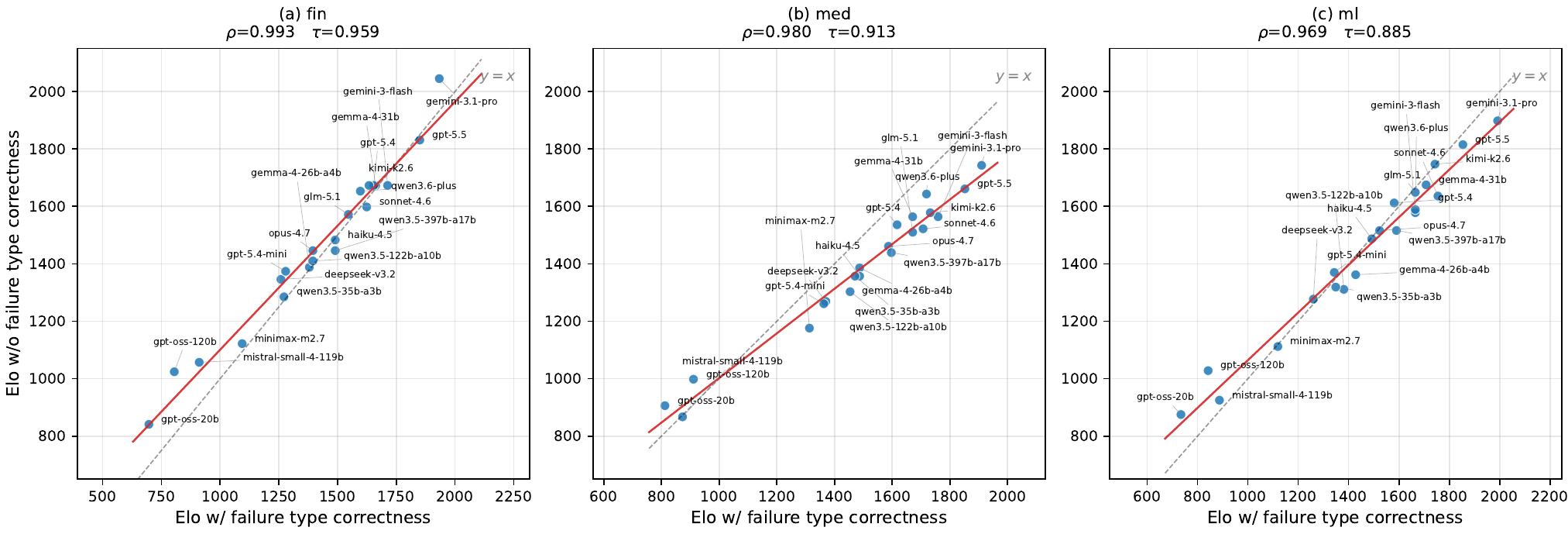}
    \caption{Per-domain version of Figure~\ref{fig:type_mitigation_sample}(a). Judge Elo with the full correctness criterion (x-axis) vs.\ Elo with the failure-type label dropped from the loss (y-axis), refit independently within each domain. Each marker is a judge; the dashed line marks $y = x$ and the red line is the least-squares fit. Per-panel Spearman $\rho$ and Kendall $\tau$ are computed on the underlying Elo scores.}
    \label{fig:type_correctness_domains}
\end{figure}

Figure~\ref{fig:type_correctness_domains} repeats the with-vs.-without-type analysis inside each domain. Spearman $\rho$ ranges from $0.969$ (ML) to $0.993$ (Finance) and Kendall $\tau$ from $0.885$ to $0.959$, with the red fits sitting below the diagonal at the top of every panel. The seven-way taxonomy therefore widens the Elo range within every domain without reordering the local leaderboard, matching the combined-dataset result.

\subsection{Subsample Stability by Domain}
\label{sec:sample_size_domains}
\begin{figure}[htbp]
    \centering
    \includegraphics[width=\textwidth]{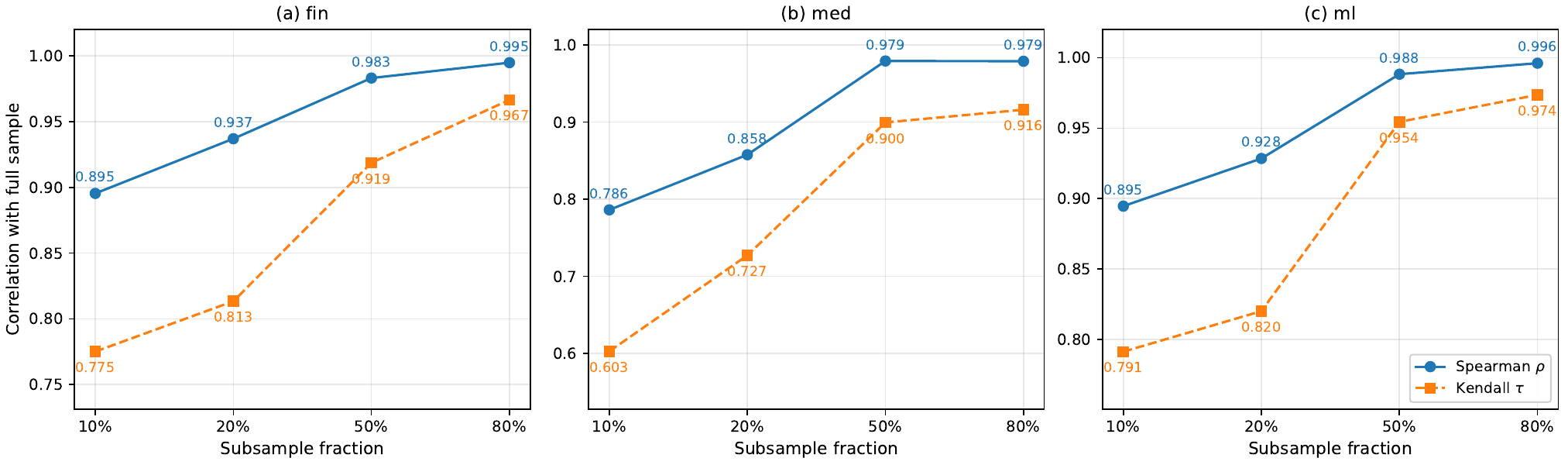}
    \caption{Per-domain version of Figure~\ref{fig:type_mitigation_sample}(c). Each panel reports Spearman $\rho$ and Kendall $\tau$ between the full-sample \texttt{top\_removed} BT ranking and the \texttt{partial\_obs} BT ranking refit on uniform subsamples at fractions $\{10\%, 20\%, 50\%, 80\%\}$, computed independently within each domain.}
    \label{fig:sample_size_domains}
\end{figure}

Figure~\ref{fig:sample_size_domains} reproduces the subsample-stability check within each domain. Spearman $\rho$ exceeds $0.98$ from the $50\%$ subsample onward in all three domains and Kendall $\tau$ crosses $0.9$ for Machine Learning and Finance at the same fraction. Biomedical is the slowest to stabilise ($\rho = 0.786$ at $10\%$), consistent with the tighter mid-leaderboard band for that domain reported in Appendix~\ref{sec:judge_scores}, but joins the other two by the $50\%$ fraction. The match-cost reduction therefore transfers to per-domain leaderboards as well, with a marginally larger minimum sample required for Biomedical.

\subsection{Top-Ranked Pair Removal}
\label{sec:noise_reduce}
\begin{figure}[htbp]
    \centering
    \includegraphics[width=\textwidth]{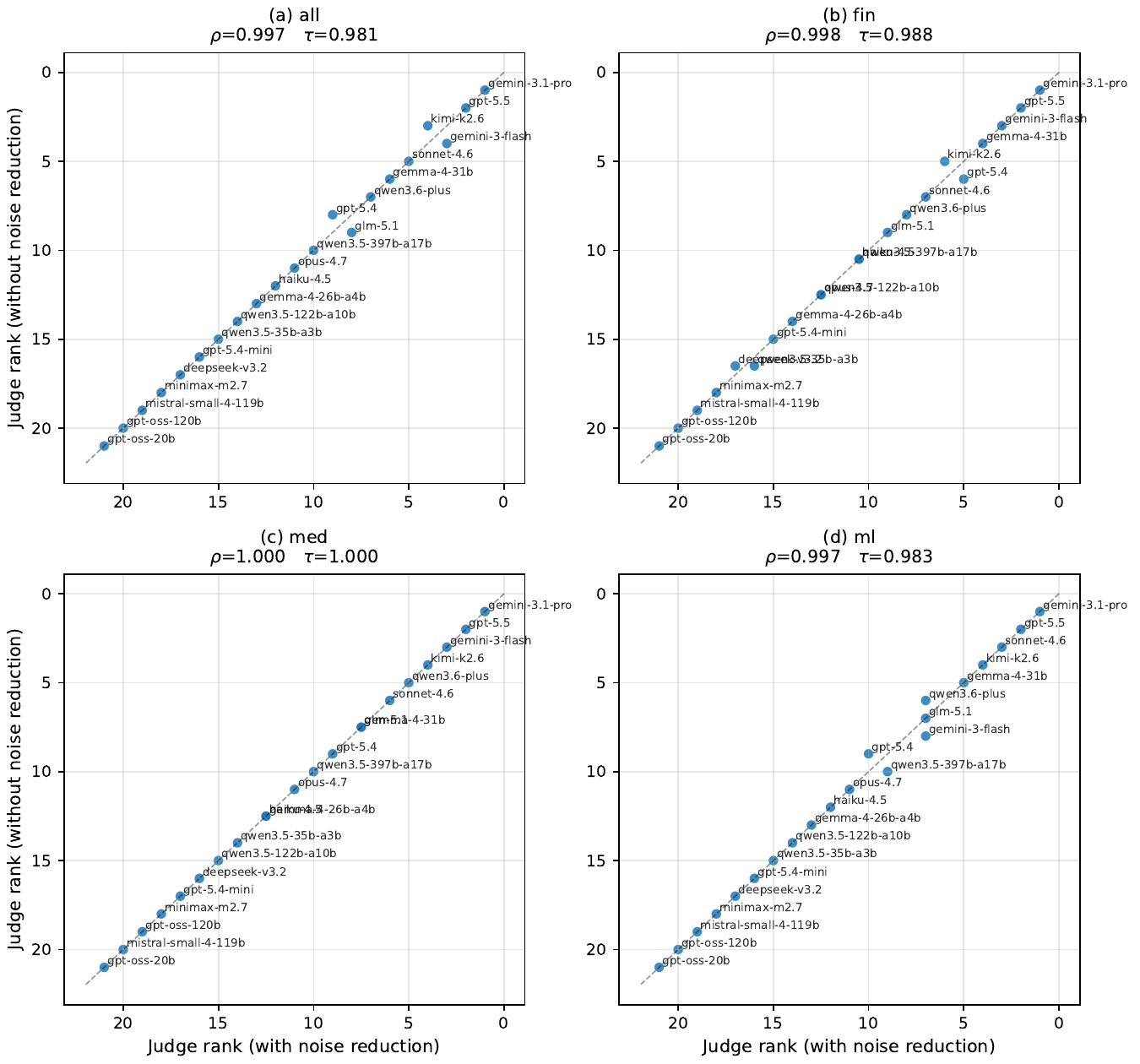}
    \caption{Judge ranks with (x-axis) and without (y-axis) the top-ranked pair removal, one panel per slice. The ``with'' condition is the published \texttt{top\_removed} slice; the ``without'' condition is the raw filtered slice taken straight from the automatic verification pipeline of Appendix~\ref{sec:prompt_verification}. Each marker is a judge; the dashed line is $y = x$. Both axes are inverted so that rank 1 sits at the top-right. Per-panel Spearman $\rho$ and Kendall $\tau$ are computed on the underlying Elo scores.}
    \label{fig:noise_reduce}
\end{figure}

The published evaluation pipeline applies a single noise-mitigation step on top of automatic verification: the top-ranked pair removal, which drops the highest-Elo pairs from the filtered pool. The human audit of Appendix~\ref{sec:human_eval} and the SFT-with-noise experiments of Appendix~\ref{sec:sft_details} are not pipeline components but supporting evidence for this step, showing that the top-Elo cohort is where mislabelled \texttt{ambiguous} and \texttt{noise} pairs concentrate, so removing it strips the bulk of residual label noise. Figure~\ref{fig:noise_reduce} closes the loop: refitting Elo on the raw filtered slice yields a leaderboard nearly identical to the published one, with Spearman $\rho \geq 0.997$, Kendall $\tau \geq 0.981$, and the top-$10$ set preserved in full on all four slices. We read this two ways. First, the automated quality control of Appendix~\ref{sec:prompt_verification} already removes most of the label noise, so the additional top-Elo trim and the human audit only have small residual noise left to clean up. Second, the Bradley--Terry ranking and the surrounding evaluation pipeline are robust under that small residual noise.

\subsection{Ranking Methods: BT vs.\ EIP}
\label{sec:ranking_method}
\begin{figure}[htbp]
    \centering
    \includegraphics[width=\textwidth]{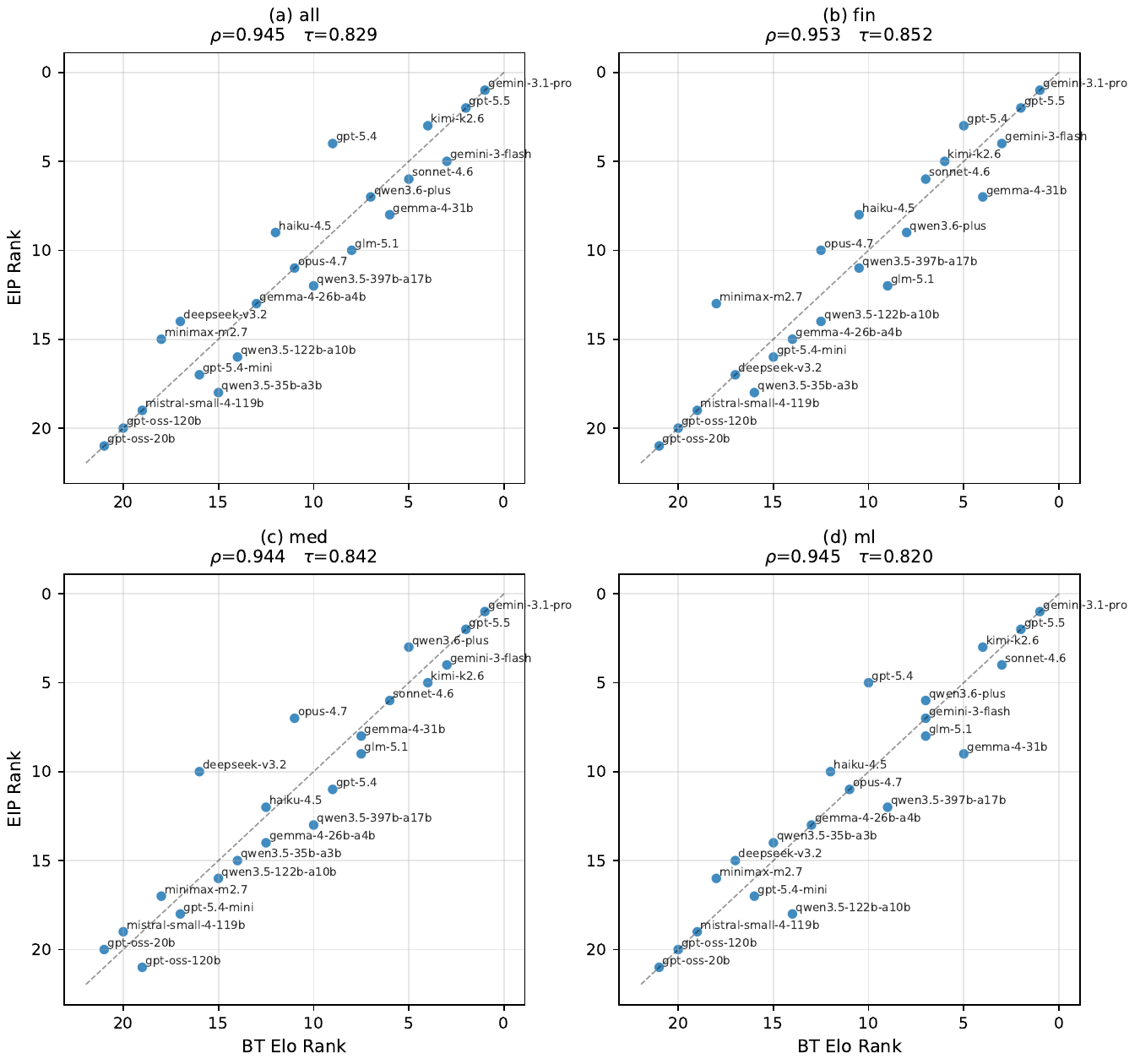}
    \caption{Judge ranks under Bradley--Terry Elo (x-axis) and Empirical Interaction Propagation (y-axis), one panel per slice. Each marker is a judge; the dashed line is $y = x$. Both axes are inverted so that rank 1 sits at the top-right. Per-panel Spearman $\rho$ and Kendall $\tau$ are computed on the underlying scores.}
    \label{fig:ranking_method}
\end{figure}

Figure~\ref{fig:ranking_method} compares the Bradley--Terry Elo leaderboard with the difficulty-aware EIP ranking of Appendix~\ref{sec:eip}, on the combined dataset and each domain. The two methods agree strongly across all four slices: Spearman $\rho$ ranges from $0.944$ to $0.953$ and Kendall $\tau$ from $0.820$ to $0.852$, with the top-$3$ set always sharing two of three judges and the top-$10$ sets overlapping by $80\%$ to $90\%$. EIP is designed to up-weight judges that solve pairs which fool strong judges, so its near-identical ordering with Elo indicates that the benchmark does not contain a pocket of hard pairs whose signal Elo discards. We adopt Bradley--Terry Elo as the primary score throughout the paper for its closed-form cluster-robust confidence intervals (Appendix~\ref{sec:bt_details}) and its stability under partial observability (Figure~\ref{fig:sample_size_domains}).

\subsection{Pointwise Judges}
\label{sec:pointwise}
\begin{figure}[htbp]
    \centering
    \includegraphics[width=\textwidth]{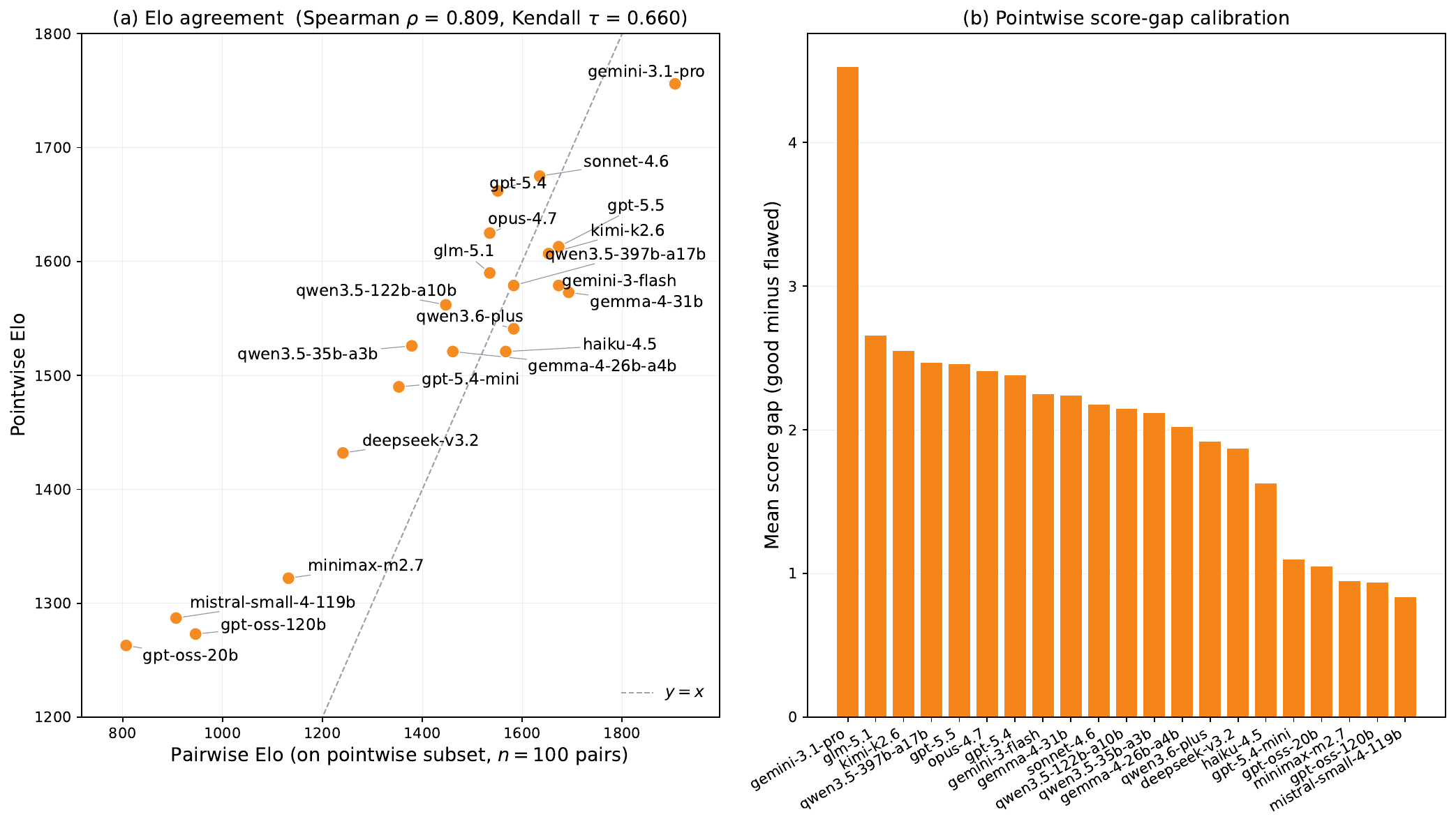}
    \caption{Pointwise vs.\ pairwise judging on a $100$-pair stratified sample. (a) Each judge's pointwise Elo plotted against its pairwise Elo recomputed on the same pair set; the dashed line marks $y = x$. (b) Per-judge pointwise score-gap: the mean Likert score on the good conversation in each pair minus the mean score on the flawed one, sorted descending.}
    \label{fig:pointwise}
\end{figure}

\paragraph{Setup.} The pointwise prompt of Appendix~\ref{sec:prompt_judge} differs from the pairwise default in that each judge sees a single conversation in isolation and returns a $1$-$10$ Likert score together with an \texttt{is\_flawed} flag, the worst-round index, and the A-G flaw category. Pairwise verdicts are derived post-hoc by scoring \texttt{convo\_a} and \texttt{convo\_b} independently and taking the higher-scoring side as the winner; equal scores are recorded as ties and given $0.5$/$0.5$ credit so the same Bradley--Terry machinery applies. The experiment is run on a $100$-pair sample stratified by domain and assistant failure type, judged by the same $21$-model panel; the pairwise axis in Figure~\ref{fig:pointwise}(a) is recomputed on the identical sample, so the only difference between the two axes is the prompt mode.

\paragraph{Findings.} Pointwise and pairwise rankings agree at the coarse level (Spearman $\rho = 0.809$, Kendall $\tau = 0.660$) but the judge rankings differ. This divergence is expected, because the two protocols measure different judge capabilities: pairwise tests discrimination between two candidate conversations under direct comparison, while pointwise tests absolute calibration of conversation quality without a reference. Panel (b) makes the calibration axis explicit: the mean score gap between the good and flawed conversations exceeds $4$ Likert points for \texttt{gemini-3.1-pro} but falls below $1$ for the weakest three judges, a per-judge resolution that pairwise discards once scores collapse to a binary winner. Pairwise remains the more discriminative protocol for ranking judges; the pointwise score-gap serves as a complementary diagnostic for absolute calibration.

\clearpage
\section{Reproducibility}

\subsection{Bradley--Terry Elo Details}
\label{sec:bt_details}

\paragraph{Setup, log-likelihood, and MM iteration.}
With notation as in Section~\ref{sec:eval_protocol}, the BT win probability $\Pr(c=1\mid j,q) = \theta_j / (\theta_j + \theta_q)$ induces the log-likelihood
\begin{equation}
    \ell(\bm{\theta})
    \;=\;
    \sum_{(j, q, c) \in \mathcal{D}}
    \Bigl[
        c \log \tfrac{\theta_j}{\theta_j + \theta_q}
        +
        (1 - c) \log \tfrac{\theta_q}{\theta_j + \theta_q}
    \Bigr].
    \label{eq:bt-loglik}
\end{equation}
Let $w_i = \sum_{(j,q,c) \in \mathcal{D}} \bigl[c \,\mathbf{1}[i = j] + (1-c)\, \mathbf{1}[i = q]\bigr]$ denote the total wins of player $i$, and let $n_{ik} = \bigl|\{t : \{j_t, q_t\} = \{i, k\}\}\bigr|$ denote the total number of matches between players $i$ and $k$. The standard MM update~\citep{hunter2005variable} maximises~\eqref{eq:bt-loglik} via
\begin{equation}
    \theta_i^{(t+1)}
    \;=\;
    \frac{w_i}{\displaystyle\sum_{k \,:\, n_{ik} > 0}
        \frac{n_{ik}}{\theta_i^{(t)} + \theta_k^{(t)}}}.
    \label{eq:bt-mm}
\end{equation}
After each iteration we mean-normalise $\bm{\theta} \leftarrow \bm{\theta} / \bar{\theta}$ to fix the otherwise unidentifiable global scale, and we declare convergence when $\|\bm{\theta}^{(t+1)} - \bm{\theta}^{(t)}\|_\infty < 10^{-6}$ (or after $1000$ iterations). To avoid numerical issues we floor the strengths at $10^{-10}$ before taking logarithms, and finally we map the strengths to the Elo scale,
\begin{equation}
    R_i \;=\; 400 \log_{10}\bigl(\theta_i\bigr) + R_0,
    \label{eq:bt-to-elo}
\end{equation}
with $R_0 = 1500$.

\paragraph{Cluster-robust sandwich standard errors.}
A single conversation pair contributes one match per judge, so the residuals
within a pair are not independent: a pair that happens to be ambiguous will
fool many judges simultaneously. We therefore report a sandwich variance
estimator with clustering at the pair level. Working in the log-strength
parameterisation $\beta_i = \log \theta_i$, define for each match $(j, q, c)$
\begin{equation}
    \eta_{j,q} \;=\; \beta_j - \beta_q,
    \qquad
    \sigma_{j,q} \;=\; \frac{1}{1 + e^{-\eta_{j,q}}},
    \qquad
    w_{j,q} \;=\; \sigma_{j,q}\bigl(1 - \sigma_{j,q}\bigr).
    \label{eq:bt-eta-sigma}
\end{equation}
The observed information matrix $I \in \mathbb{R}^{(M+Q) \times (M+Q)}$
accumulates the symmetric contribution
\begin{equation}
    I_{jj} \mathrel{+}{=} w_{j,q}, \quad
    I_{qq} \mathrel{+}{=} w_{j,q}, \quad
    I_{jq} \mathrel{-}{=} w_{j,q}, \quad
    I_{qj} \mathrel{-}{=} w_{j,q},
    \label{eq:bt-info}
\end{equation}
for every $(j, q, c) \in \mathcal{D}$. For each pair-cluster $\rho \in \mathcal{Q}$
we accumulate a score vector $\bm{s}_\rho \in \mathbb{R}^{M+Q}$ from its residuals
$r_{j,q} = c - \sigma_{j,q}$:
\begin{equation}
    \bm{s}_\rho[j] \mathrel{+}{=} r_{j,q}, \qquad
    \bm{s}_\rho[q] \mathrel{-}{=} r_{j,q}, \qquad
    \forall (j, q, c) \in \mathcal{D} \text{ with } q = \rho.
    \label{eq:bt-cluster-score}
\end{equation}
The cluster-robust ``meat'' matrix is then
\begin{equation}
    B \;=\; \sum_{\rho \in \mathcal{Q}} \bm{s}_\rho \bm{s}_\rho^{\top},
    \label{eq:bt-meat}
\end{equation}
and the sandwich variance on the log-strength scale is
\begin{equation}
    V \;=\; I^{+} \, B \, I^{+},
    \label{eq:bt-sandwich}
\end{equation}
where $I^{+}$ denotes the Moore--Penrose pseudoinverse, used because $I$ has
an unavoidable rank-1 deficiency from the global-shift invariance of $\bm{\beta}$
(and may have additional deficiencies if the comparison graph splits into
disconnected components, which we detect and warn about). Mapping back to the
Elo scale via~\eqref{eq:bt-to-elo} gives
\begin{equation}
    \mathrm{SE}_{\mathrm{Elo}}(i)
    \;=\;
    \frac{400}{\ln 10}\,\sqrt{\bigl[V\bigr]_{ii}},
    \qquad
    \mathrm{CI}_{95}(i)
    \;=\;
    1.96 \cdot \mathrm{SE}_{\mathrm{Elo}}(i).
    \label{eq:bt-se-elo}
\end{equation}
We report $R_i \pm \mathrm{CI}_{95}(i)$ for every judge and pair.

\subsection{Empirical Interaction Propagation (EIP) Details}
\label{sec:eip}

Let $\mathcal{J} = \{1, \dots, M\}$ be the set of LLM judges and $\mathcal{Q} = \{1, \dots, Q\}$ be the set of conversation pairs. We assume full observability: every judge has been evaluated on every pair, with outcome $c_{j, q} \in \{0, 1\}$ equal to $1$ iff judge $j$ produces the correct verdict on pair $q$ under the joint criterion of Eq~\ref{eq:main_correctness} (matching the ground-truth verdict $y^\star$, flawed turn $r^\star$, and failure category $t^\star$). EIP casts judge-quality scoring as a PageRank-style random walk on the bipartite graph of judges and pairs. We
compute it on the \emph{filtered} match set, where any pair on which a judge
is missing or on which all judges agree (unanimously correct or unanimously
incorrect) is removed; such pairs carry no discriminative signal and break
the identifiability of the random walk.

Let $A \in \{0, 1\}^{Q \times M}$ be the binary competency matrix, with
$A_{q, j} = c_{j, q}$. Let $\mathbf{1}$ denote the $Q \times M$ matrix of
ones, and define the failure matrix
\begin{equation}
    \widehat{A} \;=\; \mathbf{1} - A.
    \label{eq:eip-Ahat}
\end{equation}
The per-pair correctness count and per-judge failure count are floored at $1$
to avoid division by zero on (rare) all-correct rows or all-failed columns,
\begin{equation}
    S_q \;=\; \max\!\Bigl(\sum_{j} A_{q, j},\; 1\Bigr),
    \qquad
    F_j \;=\; \max\!\Bigl(\sum_{q} \widehat{A}_{q, j},\; 1\Bigr).
    \label{eq:eip-degrees}
\end{equation}
We form two row-stochastic transition matrices on the bipartite graph,
\begin{equation}
    P_{Q \to M}[q, j]
    \;=\; \frac{A_{q, j}}{S_q},
    \qquad
    P_{M \to Q}[j, q]
    \;=\; \frac{\widehat{A}_{q, j}}{F_j}.
    \label{eq:eip-transitions}
\end{equation}
Walking from a pair, the chain hops to a judge that solved it (weighted by
that judge's competence on this pair); walking from a judge, the chain hops
to a pair that judge \emph{failed} (weighted by that judge's failure mass).
Iterating this walk concentrates probability mass on judges that solve hard
pairs and on pairs that fool strong judges---precisely the desired
``difficulty-aware'' ranking.

With damping factor $\alpha = 0.85$ and uniform teleport vectors
$\bm{u}_M = \mathbf{1}_M / M$ and $\bm{u}_Q = \mathbf{1}_Q / Q$, we run the
coupled iteration
\begin{align}
    \bm{\pi}_Q^{(t+1)}
    &\;=\;
    \alpha\, P_{M \to Q}^{\top}\, \bm{\pi}_M^{(t)}
    \;+\; (1 - \alpha)\, \bm{u}_Q,
    \label{eq:eip-piQ} \\
    \bm{\pi}_M^{(t+1)}
    &\;=\;
    \alpha\, P_{Q \to M}^{\top}\, \bm{\pi}_Q^{(t+1)}
    \;+\; (1 - \alpha)\, \bm{u}_M,
    \label{eq:eip-piM}
\end{align}
initialised at $\bm{\pi}_M^{(0)} = \bm{u}_M$ and
$\bm{\pi}_Q^{(0)} = \bm{u}_Q$. Both vectors are renormalised to sum to $1$
after each step to prevent numerical drift, and we declare convergence when
\begin{equation}
    \bigl\|\bm{\pi}_M^{(t+1)} - \bm{\pi}_M^{(t)}\bigr\|_1
    +
    \bigl\|\bm{\pi}_Q^{(t+1)} - \bm{\pi}_Q^{(t)}\bigr\|_1
    \;<\; 10^{-6},
    \label{eq:eip-convergence}
\end{equation}
or after $100$ iterations.

Finally, both stationary distributions are min--max rescaled to the
interpretable range $[0, 100]$. For $\bm{v} \in \{\bm{\pi}_M, \bm{\pi}_Q\}$,
\begin{equation}
    \mathrm{score}_{\mathrm{EIP}}(i)
    \;=\;
    100 \cdot
    \frac{v_i - \min_k v_k}{\max_k v_k - \min_k v_k},
    \label{eq:eip-rescale}
\end{equation}
with the degenerate convention $\mathrm{score}_{\mathrm{EIP}}(i) = 50$ for all
$i$ whenever $\max_k v_k - \min_k v_k < 10^{-12}$. The reported judge ranking
is the descending sort of $\{\mathrm{score}_{\mathrm{EIP}}(j)\}_{j \in \mathcal{J}}$,
and the corresponding pair difficulty is
$\{\mathrm{score}_{\mathrm{EIP}}(q)\}_{q \in \mathcal{Q}}$.

\subsection{Prompt for Assistant Failure Type Discovery}
\label{sec:prompt_discovery}
The following box gives the prompt used for automated discovery and refinement of the assistant failure type taxonomy used in \Cref{sec:dataset}.
\begin{tcolorbox}[colback=gray!5, colframe=gray!50!black, title=\texttt{Assistant Failure Type Discovery Prompt}, fonttitle=\bfseries]
You are an expert in multi-turn conversation evaluation, specializing in identifying the challenging scenarios, conducting failure mode and taxonomy analysis. \\ 
\\
Your task is fourfold: \\
(1) Task 1: per-sample annotation. Apply an existing taxonomy to label assistant and user behaviors in multi-turn conversations \\
(2) Task 2: identify taxonomy gaps. Identify cases that do NOT fit the taxonomy \\
(3) Task 3: taxonomy refinement. Categorize the missed cases and propose refined or new categories grounded in evidence \\
(4) Task 4: insights generation. Produce outputs for downstream analysis, explanations for the proposal. \\

\#\# TAXONOMY DEFINITIONS \\
{\ttfamily
ASSISTANT\_BEHAVIORS = \{\\
\quad "disorganized": \{...\},\\
\quad "fabricated\_answer": \{...\},\\
\quad "unnecessary\_refusal": \{...\},\\
\quad ...\\
\}\\

USER\_BEHAVIORS = \{\\
\quad "focused": "...",\\
\quad "skeptical": "...",\\
\quad "misinformed": "...",\\
\quad "exploratory": "...",\\
\quad ...\\
\}
} \\

\#\# GENERAL GUIDELINES \\
{\ttfamily
- Be precise, not verbose; avoid generic explanations.\\
- Ground reasoning in observed data; do NOT hallucinate missing context.\\
- Prefer concrete behavioral descriptions over abstract wording. Identify ambiguities in definitions.\\
- Be critical of the taxonomy when necessary, but only propose changes supported by repeated evidence.
} \\

\#\# DATA EXAMPLES 
\end{tcolorbox}

\subsection{Assistant Failure Types}
The following boxes define the expected behaviours for assistant failure types, and the corresponding benign behaviours when no failure is present. These are the full versions of the behaviour specifications that assistants are conditioned on, corresponding to \Cref{tab:taxonomy-assistant}.
\label{sec:taxonomy_assistant_full}
%
\newtcolorbox{flawbox}[1]{
  enhanced,
  breakable,
  colback=SoftRed,
  colframe=RuleGray,
  boxrule=0.6pt,
  arc=2pt,
  left=8pt,right=8pt,top=6pt,bottom=6pt,
  title={#1},
  fonttitle=\bfseries,
}
\newtcolorbox{virtuebox}[1]{
  enhanced,
  breakable,
  colback=SoftGreen,
  colframe=RuleGray,
  boxrule=0.6pt,
  arc=2pt,
  left=8pt,right=8pt,top=6pt,bottom=6pt,
  title={#1},
  fonttitle=\bfseries,
}

\begin{flawbox}{\texttt{self\_contradiction} --- Flaw}
The assistant subtly contradicts a claim from an earlier correct turn --- not by lexically flipping it, but by shifting a qualifier (`primary factor' becomes `one of several factors'), reframing emphasis, or stating something whose implications conflict with what the assistant said before. Use different wording so the inconsistency isn't surface-detectable; catching it requires mentally reconciling the flawed later turn against an earlier one. The flawed round is the later turn whose claim no longer matches what was said before, while the earlier reference turn remains accurate and grounded in the source. Avoid lexically obvious flips, and do not restate the earlier claim verbatim immediately before contradicting it. Non-flawed turns stay consistent with the assistant's own prior statements.
\end{flawbox}

\begin{virtuebox}{\texttt{self\_contradiction} --- Benign}
Stay consistent with your own earlier turns --- match the scope, strength, and qualifiers of prior claims across the conversation. If your understanding genuinely evolves, explicitly acknowledge what changed and why rather than silently shifting.
\end{virtuebox}

\begin{flawbox}{\texttt{evasion} --- Flaw}
The chatbot produces a response that looks thorough and knowledgeable but avoids giving the user what they actually asked for. This can take two forms, or blend them: (a) surface specificity that hedges on the central claim --- citing a number for a minor result while glossing the key finding, or naming a method without explaining what makes it distinct; (b) tangential drift --- emphasizing adjacent or related material so the answer reads plausibly while missing the core point. Either way, the response should look reasonable to a skimming reader but leave the user unable to verify or act on the central claim. Non-flawed turns answer the user's actual question directly with the key details, exact numbers, and specific terms the question demands.
\end{flawbox}

\begin{virtuebox}{\texttt{evasion} --- Benign}
For each question, provide a complete, specific, on-topic answer --- every sentence directly addresses what the user asked, with the key details, exact numbers, and specific terms that are relevant to the actual question.
\end{virtuebox}

\begin{flawbox}{\texttt{disorganized} --- Flaw}
The chatbot buries key information in text that looks similar in length and formatting to clean responses but lacks logical flow --- jumping between points, mixing high-level claims with low-level details mid-paragraph, and failing to foreground the answer the user actually asked for. The flaw is structural, not visual: the response should not look like a wall of text. The correct content is present but requires significant effort to extract. Non-flawed turns should present information clearly and directly.
\end{flawbox}

\begin{virtuebox}{\texttt{disorganized} --- Benign}
Present information with clear structure --- lead with the key point, organize logically, make the answer easy to scan and extract insights from.
\end{virtuebox}

\begin{flawbox}{\texttt{fabricated\_answer} --- Flaw}
The chatbot fabricates a confident-sounding answer when the question is not answerable from the reference material, instead of honestly saying it doesn't have enough information. Non-flawed turns should acknowledge limitations when the question goes beyond what the reference material supports.
\end{flawbox}

\begin{virtuebox}{\texttt{fabricated\_answer} --- Benign}
Some of the user's questions go beyond what the reference material covers. The assistant recognizes these and honestly says it doesn't have enough information rather than guessing, while answering confidently when the material does cover the question. Demonstrate clear calibration between what the source supports and what it doesn't --- at least one turn should acknowledge a limit.
\end{virtuebox}

\begin{flawbox}{\texttt{instruction\_forgetting} --- Flaw}
Early in the conversation the user establishes an explicit content-level directive --- a scope restriction (`focus only on the ablation results, skip the architecture discussion'; `only pull from the abstract, not the full methods'), an audience/framing (`I'm a clinician making treatment decisions, not a student'; `graduate-level CS, assume transformer basics'), self-context (`I'm researching Alzheimer's specifically, not dementia broadly'; `I've already read section 3, don't re-explain it'), a prioritization directive (`I care more about limitations than strengths'; `methodology over results'), or a negative content instruction (`don't walk me through the reasoning, just give the conclusion'; `don't re-cover material we already discussed'). The assistant honors the directive through the intermediate rounds. In the flawed round it silently violates the directive while otherwise answering competently --- pulling in out-of-scope material `to provide context', drifting to a different audience level, re-explaining what the user said to skip, or slipping into the deprioritized angle. The violation must be partial or plausibly-justified (not a wholesale abandonment) and must not be lexically announced --- no `let me expand beyond\ldots', `stepping outside the scope for a moment\ldots'. Do not use format/stylistic directives (bullets vs prose, word counts, citation style) since those are surface-detectable and don't test reasoning. Non-flawed turns apply the directive consistently.
\end{flawbox}

\begin{virtuebox}{\texttt{instruction\_forgetting} --- Benign}
Early in the conversation the user establishes an explicit content-level directive --- a scope restriction, audience/framing, self-context, prioritization, or negative-content instruction (not surface formatting). The assistant tracks the directive and applies it consistently across every subsequent turn, even when not restated. Responses should demonstrate that the directive is actively shaping what gets included, emphasized, or omitted --- not by naming the directive, but through the content choices themselves. If a later question would genuinely conflict with the directive, flag the conflict and ask rather than silently deviating.
\end{virtuebox}

\begin{flawbox}{\texttt{no\_clarification} --- Flaw}
When the user's question is ambiguous or underspecified --- e.g., it could refer to multiple parts of the material, or lacks enough context to give a precise answer --- the chatbot picks one interpretation and answers confidently instead of asking the user to clarify. Non-flawed turns should ask for clarification when the question is genuinely ambiguous.
\end{flawbox}

\begin{virtuebox}{\texttt{no\_clarification} --- Benign}
At least one of the user's questions is genuinely ambiguous or underspecified (could refer to multiple parts of the material, or lacks enough context for a precise answer). On those, the assistant asks for clarification before answering rather than silently assuming one interpretation. On unambiguous questions, the assistant answers directly without performative clarification requests.
\end{virtuebox}

\begin{flawbox}{\texttt{unnecessary\_refusal} --- Flaw}
The chatbot unnecessarily refuses or hedges on questions that ARE answerable from the reference material, saying things like `I don't have enough information' or `I can't determine that' when the answer is clearly available. Non-flawed turns should answer confidently using the reference material.
\end{flawbox}

\begin{virtuebox}{\texttt{unnecessary\_refusal} --- Benign}
The user asks questions that sit near the edge of what the source covers but are answerable if the assistant reads the material carefully. The assistant answers confidently using the available information without unnecessary hedging or refusal. Only hedge when a question is genuinely outside the source's scope.
\end{virtuebox}

\subsection{User Behavior Types}
\label{sec:taxonomy_user}
The following boxes define the user-behavior taxonomy used in \Cref{sec:dataset}. Each generated pair samples one user behavior, which conditions how the simulated user interacts with the assistant across all rounds. The seven behaviors split into navigation patterns (how the user moves through the material) and epistemic stances (how the user relates to the material's content). Keys match the labels used by the pipeline.

\begin{tcolorbox}[colback=gray!5, colframe=gray!50!black, title=\texttt{focused}, fonttitle=\bfseries]
The user picks one or two specific aspects of the material and explores them in depth across multiple turns, asking follow-up questions that build on previous answers.
\end{tcolorbox}

\begin{tcolorbox}[colback=gray!5, colframe=gray!50!black, title=\texttt{integrative}, fonttitle=\bfseries]
The user asks broad questions connecting different parts of the material - linking details across sections, comparing information, wanting synthesis rather than isolated facts.
\end{tcolorbox}

\begin{tcolorbox}[colback=gray!5, colframe=gray!50!black, title=\texttt{scattered}, fonttitle=\bfseries]
The user jumps abruptly between unrelated aspects of the material without transitions, forcing the assistant to handle rapid context switches.
\end{tcolorbox}

\begin{tcolorbox}[colback=gray!5, colframe=gray!50!black, title=\texttt{skeptical}, fonttitle=\bfseries]
The user plays devil's advocate - questioning claims, asking for evidence, pointing out apparent tensions between answers, and requiring justification before accepting conclusions.
\end{tcolorbox}

\begin{tcolorbox}[colback=gray!5, colframe=gray!50!black, title=\texttt{misinformed}, fonttitle=\bfseries]
The user confidently states incorrect things about the material (wrong numbers, misattributed claims, flipped conclusions) and builds questions on these false premises. The assistant should correct these errors.
\end{tcolorbox}

\begin{tcolorbox}[colback=gray!5, colframe=gray!50!black, title=\texttt{exploratory}, fonttitle=\bfseries]
The user treats the material as a starting point and asks about implications, broader connections, or extensions that may go beyond what the reference material covers.
\end{tcolorbox}

\begin{tcolorbox}[colback=gray!5, colframe=gray!50!black, title=\texttt{underspecified}, fonttitle=\bfseries]
The user sometimes asks vague or ambiguous questions - using unclear references, omitting key qualifiers, or phrasing questions that could apply to multiple parts of the material - forcing the assistant to either guess the intent or ask for clarification.
\end{tcolorbox}

\subsection{Prompts for Dual Conditioned Conversation Pair Generation}
\label{sec:prompt_generation}
The following boxes display the prompts used for generating conversation based on blueprints, as described in \Cref{sec:dataset}. The two prompts give instructions for generating a good conversation, and a worse conversation which intentionally has weaknesses.
%

\begin{tcblisting}{
  colback=gray!5, colframe=gray!50!black,
  title=\texttt{good\_convo}, fonttitle=\bfseries,
  listing only,
  listing options={basicstyle=\ttfamily\footnotesize, breaklines=true, breakindent=0pt, columns=fullflexible, keepspaces=true}
}
This conversation will be used in a pairwise comparison against a counterpart conversation that contains a subtle "{flaw_name}" flaw in exactly one round. The goal of the pair is to test whether an LLM judge can pick the better conversation --- which only works if your conversation is absolutely correct on every dimension and naturally embodies the paired virtue, without performing it loudly. If your conversation is obviously stronger than a clean conversation with one subtle flaw, the test becomes trivial. Aim to read as ordinary high quality, not as a demonstration of the virtue.

Given this reference document, generate a natural multi-turn conversation ({n} rounds, each round = 1 user + 1 assistant message).

The context is reference material - it defines what the assistant knows about the topic. The conversation does NOT need to cover all of it. A focused discussion exploring a few topics in depth is better than superficially touching everything.

First, write a turn-by-turn plan with one labeled entry per round in the form `Round 1: ...`, `Round 2: ...`, ..., `Round {n}: ...`. Every round MUST have its own entry --- do not collapse rounds into a single paragraph or skip ahead. For each round, capture (a) what the user asks and what motivates it (building on the prior round where natural), and (b) what the assistant's response will focus on from the reference material. The plan should form a coherent arc, not isolated Q&A. Then write the conversation faithfully executing the plan.

User behavior: "{user_behavior_name}" - {user_behavior_desc}

Assistant virtue to embody naturally (the quality every round must quietly live up to --- do NOT overact it): {assistant_virtue}

Paired flaw the counterpart conversation will contain (for your awareness only --- do NOT mention, contrast, or over-compensate against it; just never exhibit it): {assistant_flaw}

User rules:
- Has read the reference material. Asks questions driven by genuine curiosity, not to quiz the assistant.
- Apply the above behavior consistently across ALL rounds - it should shape what the user asks, how they react, and how the conversation progresses.
- NOT a passive question-asker - reacts to answers, builds on them, pushes back, or redirects.
- Keep messages SHORT and natural (1-3 sentences). Real users don't write essays.

Assistant rules:
- Every round must be clean on every dimension: accurate, grounded, specific, on-topic, well-structured, responsive, and appropriately calibrated. No flaw category should surface in any round.
- Answers must be grounded in the provided context. Do NOT fabricate information beyond what the context provides.
- If a user question goes beyond what the context covers, honestly acknowledge the limitation rather than guessing or fabricating.
- Be specific - use the actual numbers, details, and conclusions from the reference material.
- Be concise and direct (2-5 sentences per response). Avoid filler, repetition, and preamble. Match the tone, length, and confidence of a typical clean response --- do not inflate quality signals to appear more virtuous.

=== Reference Context ===
{context}
\end{tcblisting}

\begin{tcblisting}{
  colback=gray!5, colframe=gray!50!black,
  title=\texttt{worse\_convo}, fonttitle=\bfseries,
  listing only,
  listing options={basicstyle=\ttfamily\footnotesize, breaklines=true, breakindent=0pt, columns=fullflexible, keepspaces=true}
}
Given this reference document, generate a natural multi-turn conversation ({n} rounds, each round = 1 user + 1 assistant message) where the chatbot has a specific weakness.

Chatbot weakness: **{behavior_type}**
Description: {behavior_desc}

The context is reference material - it defines what the assistant knows about the topic. The conversation does NOT need to cover all of it. A focused discussion is fine.

First, write a turn-by-turn plan in the "reasoning" field with one labeled entry per round in the form `Round 1: ...`, `Round 2: ...`, ..., `Round {n}: ...`. Every round MUST have its own entry --- do not collapse the plan into a single justification paragraph or skip ahead. For each round, capture (a) what the user asks and what motivates it (reacting to the prior round where natural), and (b) what the assistant's response will focus on from the reference material. The plan should form a coherent arc, not isolated Q&A. Pick one round to be the flawed one and, inside that round's entry, describe concretely how the "{behavior_type}" weakness will surface there --- the clean rounds before and after should carry the conversation as if no flaw exists. Set "bad_round_index" to that round (1-indexed). Then write the conversation faithfully executing the plan.

User behavior: "{user_behavior_name}" - {user_behavior_desc}

User rules:
- Has read the reference material. Asks questions driven by genuine curiosity.
- Apply the above behavior consistently across ALL rounds.
- NOT a passive question-asker - reacts to answers naturally. If the chatbot gives an evasive answer, the user may follow up, push back, or move on - whatever fits. Do NOT have the user explicitly diagnose or name the flaw.
- Keep messages SHORT and natural (1-3 sentences).

Chatbot (assistant) rules:
- Answers must be grounded in the provided context. Do NOT fabricate information beyond what the context provides.
- Be specific - use the actual numbers, details, and conclusions from the reference material.
- Be concise and direct (2-5 sentences per response). Avoid filler, repetition, and preamble.
- In exactly 1 round, the assistant response should exhibit the "{behavior_type}" weakness. The flaw should require careful reading to detect --- a reader skimming should NOT notice anything wrong. Keep the same confident tone and similar response length as other rounds.
- The flawed round must exhibit ONLY the "{behavior_type}" flaw --- no other weakness category should appear in it. Every dimension of the response other than the named flaw must remain clean: accurate, grounded, specific, on-topic, well-structured, and responsive. All non-flawed rounds must be clean on every dimension.
- Do NOT mention or acknowledge that responses are flawed.

=== Reference Context ===
{context}
\end{tcblisting}

\clearpage
\subsection{Prompts for Automated Quality Control}
\label{sec:prompt_judge}
The following boxes display the prompts used for automated quality control of generated conversations, namely for checking coherence, adherence, and grounding as described in \Cref{sec:dataset}. Breaking generated text into atomic claims is a common practice when verifying factuality of statements \cite{mohri2024factuality}. Rather than demand exact factuality of all claims, it is also possible to relax to a specified precision level and calibrate removal of conversations using statistical techniques \cite{kuwahara2025}.
\label{sec:prompt_verification}
%

\begin{tcblisting}{
  colback=gray!5, colframe=gray!50!black,
  title=\texttt{coherence}, fonttitle=\bfseries,
  listing only,
  listing options={basicstyle=\ttfamily\footnotesize, breaklines=true, breakindent=0pt, columns=fullflexible, keepspaces=true}
}
You are auditing a synthetic-data recipe before any conversation has been rendered. Given a reference document and a sampled plan, decide whether the plan can plausibly produce a natural multi-turn conversation.

A plan is INCOHERENT only when executing the stated virtue/flaw is semantically at odds with the user style for THIS specific reference material - not merely unusual. Surface-level mismatches (e.g. a skeptical user asking precise questions) are fine. Flag only genuine conflicts, e.g.:
- the flaw requires conditions the user style will never create (a "focused" user asking tightly scoped questions cannot surface a "no_clarification" flaw)
- the reference material lacks the content needed for the flaw to appear naturally (e.g. "factual_error" flaw but the material is too thin to have subtle facts to misstate)
- the bad_round_index sits outside [1, n_rounds] or collides with where the plan claims the conversation is just getting started / wrapping up

User behavior: "{user_behavior_name}" - {user_behavior_desc}
Assistant virtue (good convo): {virtue}
Assistant flaw (bad convo, "{assistant_behavior_name}"): {flaw}
n_rounds: {n_rounds}
bad_round_index (1-indexed): {bad_round_index}

=== Good plan (from generator) ===
{good_plan}

=== Bad plan (from generator) ===
{bad_plan}

=== Reference Context ===
{context}

Respond in JSON: {"good_ok", "good_issue", "bad_ok", "bad_issue"}. Leave *_issue as "" when ok. Keep issues to one sentence.
\end{tcblisting}

\begin{tcblisting}{
  colback=gray!5, colframe=gray!50!black,
  title=\texttt{adherence}, fonttitle=\bfseries,
  listing only,
  listing options={basicstyle=\ttfamily\footnotesize, breaklines=true, breakindent=0pt, columns=fullflexible, keepspaces=true}
}
You are auditing whether two rendered conversations actually followed their generation plans. Read the plans and then the conversations and answer yes/no globally for each side.

For the GOOD conversation:
- Did the assistant consistently display the stated virtue across ALL rounds?
- Did the user maintain the stated user behavior across ALL rounds?

For the BAD conversation:
- Did the user maintain the stated user behavior?
- Does round {bad_round_index} (1-indexed) actually exhibit the "{assistant_behavior_name}" flaw described below?
- Is that the ONLY flawed assistant round? (If another round is clearly flawed, bad_flaw_round_correct is false.)

User behavior: "{user_behavior_name}" - {user_behavior_desc}
Assistant virtue: {virtue}
Assistant flaw "{assistant_behavior_name}": {flaw}
n_rounds: {n_rounds}

=== Good plan ===
{good_plan}

=== Good conversation ===
{good_convo}

=== Bad plan ===
{bad_plan}

=== Bad conversation (declared bad_round_index = {bad_round_index}) ===
{bad_convo}

Respond in JSON: {"good_followed", "good_issue", "bad_followed", "bad_flaw_round_correct", "bad_issue"}. Leave *_issue as "" when ok. One sentence per issue.
\end{tcblisting}

\begin{tcblisting}{
  colback=gray!5, colframe=gray!50!black,
  title=\texttt{grounding}, fonttitle=\bfseries,
  listing only,
  listing options={basicstyle=\ttfamily\footnotesize, breaklines=true, breakindent=0pt, columns=fullflexible, keepspaces=true}
}
You are fact-checking an assistant's responses against a reference document. For each listed assistant turn, extract every factual atomic claim it makes (specific numbers, named entities, mechanisms, cause/effect statements, attributions) and mark each claim as grounded or not.

Rules:
- A claim is `grounded: true` only if it is directly supported by the reference context below.
- Paraphrases and reasonable summaries of supported content count as grounded.
- Reasonable inferences or synthesis across multiple context statements also count as grounded, even if no single sentence states them verbatim.
- Unsupported extrapolation, invented details, or claims beyond the context are `grounded: false`.
- Ignore conversational filler, hedges, and meta-statements ("good question", "let me explain").
- Do NOT extract negative/absence claims ("X is not Y", "X doesn't mention Z") unless the context directly contradicts them - absence is not verifiable at claim granularity.
- Do NOT extract meta-statements where the assistant is evaluating another statement rather than asserting a fact ("it is inaccurate to claim...", "that earlier answer was wrong").
- Aim for 1-5 atomic claims per turn. Skip turns whose round_index is in skip_rounds: {skip_rounds}.

=== Assistant turns ({label}) ===
{turns}

=== Reference Context ===
{context}

Respond in JSON: {"rounds": [{"round_index", "claims": [{"claim", "grounded"}]}]}. Only include rounds you actually analyzed (exclude skip_rounds).
\end{tcblisting}

\clearpage
\subsection{Prompt for Judges and its Variants}
\label{sec:prompt_judge}
The following boxes display the prompts used for the main judge results across all 21 models, and variants used as ablations throughout \Cref{sec:experiments}. LLM-as-a-judge systems can be sensitive to prompt phrasing. Our prompt revisions were refined manually, but future iterations of the benchmark could evaluate judge robustness using prompt-level uncertainty quantification \cite{ross2026textual}.
%

\begin{tcblisting}{
  colback=gray!5, colframe=gray!50!black,
  title=\texttt{judge}, fonttitle=\bfseries,
  listing only,
  listing options={basicstyle=\ttfamily\footnotesize, breaklines=true, breakindent=0pt, columns=fullflexible, keepspaces=true, escapeinside={(*@}{@*)}}
}
You are evaluating two multi-turn conversations about the same reference material between a user and an assistant. The two conversations may have different user questions and different assistant responses - your job is to judge which conversation's assistant provides better overall quality.
{context_section}
=== Conversation A ===
{text_a}

=== Conversation B ===
{text_b}

The worse conversation has exactly one problematic assistant response. Classify the flaw using exactly one of these categories:
(*@\textcolor{blue!60!black}{\textit{[The seven assistant-weakness definitions A--G from the taxonomy are inserted verbatim here, one bullet per category in the form ``- **<letter> (<name>)**: <flaw description>''.]}}@*)

Quality flaws (evasion, disorganization, missed clarification, unnecessary refusal) count as much as factual mistakes when deciding which conversation is worse - a response can be problematic with no factual error at all. When classifying, pick the category whose definition the flawed round most closely matches; do not default to whichever category feels most common or most obvious. Each of A-G is a distinct failure mode and exactly one should fit.

Respond in JSON with these fields in order:
- "analysis": your reasoning about which conversation is better and why
- "worst_round": the round number (starting from 1) containing the worst assistant response in the conversation you did NOT pick, where each round is one user message followed by one assistant response
- "problem_type": the letter (A-G) for the flaw category from the list above
- "verdict": "A" or "B" for the better conversation
\end{tcblisting}

%
%
%

\begin{tcblisting}{
  colback=gray!5, colframe=gray!50!black,
  title=\texttt{judge\_v2}: confusion-pair hints, fonttitle=\bfseries,
  listing only,
  listing options={basicstyle=\ttfamily\footnotesize, breaklines=true, breakindent=0pt, columns=fullflexible, keepspaces=true, escapeinside={(*@}{@*)}}
}
You are evaluating two multi-turn conversations about the same reference material between a user and an assistant. The two conversations may have different user questions and different assistant responses - your job is to judge which conversation's assistant provides better overall quality.
{context_section}
=== Conversation A ===
{text_a}

=== Conversation B ===
{text_b}

The worse conversation has exactly one problematic assistant response. Classify the flaw using exactly one of these categories:
(*@\textcolor{blue!60!black}{\textit{[The seven assistant-weakness definitions A--G from the taxonomy (Sec.~\ref{sec:dataset}) are inserted verbatim here, one bullet per category in the form ``- **<letter> (<name>)**: <flaw description>''.]}}@*)

Quality flaws (evasion, disorganization, missed clarification, unnecessary refusal) count as much as factual mistakes when deciding which conversation is worse - a response can be problematic with no factual error at all. When classifying, pick the category whose definition the flawed round most closely matches; do not default to whichever category feels most common or most obvious. Each of A-G is a distinct failure mode and exactly one should fit.

(*@\textcolor{red!60!black}{Additional guidance for two categories that are easy to confuse with B (evasion):}@*)
(*@\textcolor{red!60!black}{- **F (no\_clarification)** requires that (i) the user's question is genuinely ambiguous AND (ii) the substantive answer would change depending on how the ambiguity is resolved. If an assistant asks a clarifying question and then gives a reasonable answer in the next round (or the user does not clarify), that is appropriate clarification behavior \textemdash{} it is NOT B (evasion). Only treat a missed clarification as F when the disambiguation would meaningfully change the answer.}@*)
(*@\textcolor{red!60!black}{- **A (self\_contradiction)** does not require the contradicted claim to be on the round's main question. A claim that contradicts an earlier turn \textemdash{} even when the contradiction is a tangential or seemingly random fact rather than the central point \textemdash{} is still A, not B (evasion). Do not default to B just because the inconsistency is off-topic relative to the user's current question.}@*)

Respond in JSON with these fields in order:
- "analysis": your reasoning about which conversation is better and why
- "worst_round": the round number (starting from 1) containing the worst assistant response in the conversation you did NOT pick, where each round is one user message followed by one assistant response
- "problem_type": the letter (A-G) for the flaw category from the list above
- "verdict": "A" or "B" for the better conversation
\end{tcblisting}

%
%
%

\begin{tcblisting}{
  colback=gray!5, colframe=gray!50!black,
  title=\texttt{judge\_v3}: forced candidate comparison, fonttitle=\bfseries,
  listing only,
  listing options={basicstyle=\ttfamily\footnotesize, breaklines=true, breakindent=0pt, columns=fullflexible, keepspaces=true, escapeinside={(*@}{@*)}}
}
You are evaluating two multi-turn conversations about the same reference material between a user and an assistant. The two conversations may have different user questions and different assistant responses - your job is to judge which conversation's assistant provides better overall quality.
{context_section}
=== Conversation A ===
{text_a}

=== Conversation B ===
{text_b}

The worse conversation has exactly one problematic assistant response. Classify the flaw using exactly one of these categories:
(*@\textcolor{blue!60!black}{\textit{[The seven assistant-weakness definitions A--G from the taxonomy (Sec.~\ref{sec:dataset}) are inserted verbatim here, one bullet per category in the form ``- **<letter> (<name>)**: <flaw description>''.]}}@*)

Quality flaws (evasion, disorganization, missed clarification, unnecessary refusal) count as much as factual mistakes when deciding which conversation is worse - a response can be problematic with no factual error at all. When classifying, pick the category whose definition the flawed round most closely matches; do not default to whichever category feels most common or most obvious. Each of A-G is a distinct failure mode and exactly one should fit.

(*@\textcolor{red!60!black}{Before committing to a category, you MUST explicitly compare candidates. Identify the two or three categories from A-G that could plausibly fit the flawed round, then for each candidate state (i) the specific evidence in the round that fits its definition and (ii) the strongest reason it might not fit. Only after that comparison should you commit to the single best fit, briefly explaining why it beats the runners-up. Do not pick the first category that comes to mind without weighing at least one alternative against it.}@*)

Respond in JSON with these fields in order:
- (*@\textcolor{red!60!black}{"analysis": your reasoning. This must include the candidate comparison described above before you commit to a problem\_type.}@*)
- "worst_round": the round number (starting from 1) containing the worst assistant response in the conversation you did NOT pick, where each round is one user message followed by one assistant response
- "problem_type": the letter (A-G) for the flaw category from the list above
- "verdict": "A" or "B" for the better conversation
\end{tcblisting}

%
%
%

\begin{tcblisting}{
  colback=gray!5, colframe=gray!50!black,
  title=\texttt{judge\_v4}: decision rules + comparison, fonttitle=\bfseries,
  listing only,
  listing options={basicstyle=\ttfamily\scriptsize, breaklines=true, breakindent=0pt, columns=fullflexible, keepspaces=true, escapeinside={(*@}{@*)}}
}
You are evaluating two multi-turn conversations about the same reference material between a user and an assistant. The two conversations may have different user questions and different assistant responses - your job is to judge which conversation's assistant provides better overall quality.
{context_section}
=== Conversation A ===
{text_a}

=== Conversation B ===
{text_b}

The worse conversation has exactly one problematic assistant response. Classify the flaw using exactly one of these categories:
(*@\textcolor{blue!60!black}{\textit{[The seven assistant-weakness definitions A--G from the taxonomy (Sec.~\ref{sec:dataset}) are inserted verbatim here, one bullet per category in the form ``- **<letter> (<name>)**: <flaw description>''.]}}@*)

Quality flaws (evasion, disorganization, missed clarification, unnecessary refusal) count as much as factual mistakes when deciding which conversation is worse - a response can be problematic with no factual error at all. When classifying, pick the category whose definition the flawed round most closely matches; do not default to whichever category feels most common or most obvious. Each of A-G is a distinct failure mode and exactly one should fit.

(*@\textcolor{red!60!black}{Discriminating tests for the categories most often confused with B (evasion):}@*)

(*@\textcolor{red!60!black}{- **F (no\_clarification) decision rule.** F applies only when BOTH conditions hold: (i) the user's question is genuinely ambiguous or underspecified, AND (ii) the substantive answer would meaningfully change depending on the disambiguation. If either condition fails, the round is not F. Crucially, if the assistant *did* ask a clarifying question, the round is not F and is also not B (evasion) \textemdash{} asking for clarification is appropriate behavior even when the question turns out not to need it. F means *failing to ask when asking was needed*; it does not mean *asking unhelpfully*. Do not label a round F or B simply because the assistant's clarifying question feels unnecessary in hindsight.}@*)

(*@\textcolor{red!60!black}{- **A (self\_contradiction) decision rule.** A applies whenever the flawed round contains an assertion that is inconsistent with something the assistant said in an earlier turn \textemdash{} even if the contradicted claim is tangential, a side-remark, or a random fact unrelated to the round's main question. The contradiction does not have to be on the central topic. If you can identify any earlier assistant statement that the flawed round's claims are at odds with, prefer A over B (evasion). Do not default to evasion just because the inconsistency feels off-topic or minor.}@*)

(*@\textcolor{red!60!black}{- **B (evasion) is reserved for** rounds where the assistant engages with the central question but produces a response that hedges, drifts to adjacent material, or substitutes surface specificity for the actual answer. If the flawed behavior is ``asked an unhelpful clarifying question'' $\rightarrow$ not B (consider F's rule above; usually it is none of A-G). If the flawed behavior is ``made an inconsistent claim'' $\rightarrow$ prefer A.}@*)

(*@\textcolor{red!60!black}{Before committing to a category, you MUST explicitly compare candidates. Identify the two or three categories from A-G that could plausibly fit the flawed round, then for each candidate state (i) the specific evidence in the round that fits its definition and (ii) the strongest reason it might not fit. When F or B is among your candidates, you MUST apply the F decision rule by name. When A or B is among your candidates, you MUST apply the A decision rule by name. Only after that comparison should you commit to the single best fit, briefly explaining why it beats the runners-up. Do not pick the first category that comes to mind without weighing at least one alternative against it.}@*)

Respond in JSON with these fields in order:
- (*@\textcolor{red!60!black}{"analysis": your reasoning. This must include the candidate comparison described above before you commit to a problem\_type.}@*)
- "worst_round": the round number (starting from 1) containing the worst assistant response in the conversation you did NOT pick, where each round is one user message followed by one assistant response
- "problem_type": the letter (A-G) for the flaw category from the list above
- "verdict": "A" or "B" for the better conversation
\end{tcblisting}

%
%
%

\begin{tcblisting}{
  colback=gray!5, colframe=gray!50!black,
  title=\texttt{judge\_v5}: structured two-stage CoT, fonttitle=\bfseries,
  listing only,
  listing options={basicstyle=\ttfamily\scriptsize, breaklines=true, breakindent=0pt, columns=fullflexible, keepspaces=true, escapeinside={(*@}{@*)}}
}
You are evaluating two multi-turn conversations about the same reference material between a user and an assistant. The two conversations may have different user questions and different assistant responses - your job is to judge which conversation's assistant provides better overall quality.
{context_section}
=== Conversation A ===
{text_a}

=== Conversation B ===
{text_b}

The worse conversation has exactly one problematic assistant response. Classify the flaw using exactly one of these categories:
(*@\textcolor{blue!60!black}{\textit{[The seven assistant-weakness definitions A--G from the taxonomy (Sec.~\ref{sec:dataset}) are inserted verbatim here, one bullet per category in the form ``- **<letter> (<name>)**: <flaw description>''.]}}@*)

Quality flaws (evasion, disorganization, missed clarification, unnecessary refusal) count as much as factual mistakes when deciding which conversation is worse - a response can be problematic with no factual error at all. When classifying, pick the category whose definition the flawed round most closely matches; do not default to whichever category feels most common or most obvious. Each of A-G is a distinct failure mode and exactly one should fit.

(*@\textcolor{red!60!black}{Before committing to a category, your ``analysis'' field MUST proceed in two stages.}@*)

(*@\textcolor{red!60!black}{**Stage 1 \textemdash{} Findings.** Once you have located the flawed round, answer the following five diagnostic questions in order. Each answer is ``Yes'' or ``No'' followed by one short sentence of evidence drawn from the conversation. Do not skip any question; do not collapse them into prose. These findings constrain which categories are even eligible \textemdash{} you will use them in Stage 2.}@*)

(*@\textcolor{red!60!black}{\ \ Q1. Is the user's question in the flawed round genuinely ambiguous or underspecified (could plausibly refer to multiple distinct things in the source)?}@*)
(*@\textcolor{red!60!black}{\ \ Q2. If Q1 is Yes: would the substantive answer meaningfully change depending on the disambiguation? (If Q1 is No, answer ``N/A''.)}@*)
(*@\textcolor{red!60!black}{\ \ Q3. Did the assistant in the flawed round ask a clarifying question (rather than committing to one interpretation)?}@*)
(*@\textcolor{red!60!black}{\ \ Q4. Does the flawed round contain any assertion \textemdash{} central or tangential \textemdash{} that is inconsistent with something the assistant said in an earlier turn?}@*)
(*@\textcolor{red!60!black}{\ \ Q5. Does the flawed round decline, hedge, or claim insufficient information for a question that the reference material actually answers?}@*)

(*@\textcolor{red!60!black}{**Stage 2 \textemdash{} Comparison and commit.** Using your findings, identify the two or three categories from A-G that could plausibly fit, and for each candidate state (i) the specific evidence in the round that fits its definition and (ii) the strongest reason it might not fit, citing the relevant findings (Q1-Q5) by number. The findings are operational tests:}@*)

(*@\textcolor{red!60!black}{\ \ - F (no\_clarification) is eligible only if Q1=Yes AND Q2=Yes AND Q3=No. If Q3=Yes, F is ruled out (asking for clarification, even unhelpfully, is not the F flaw).}@*)
(*@\textcolor{red!60!black}{\ \ - A (self\_contradiction) is eligible whenever Q4=Yes, regardless of whether the contradiction is on the round's main question or a tangential claim.}@*)
(*@\textcolor{red!60!black}{\ \ - B (evasion) is the right call when the assistant engages with the central question but hedges, drifts, or substitutes surface specificity for the actual answer. If Q3=Yes (assistant asked a clarifier), B is usually ruled out for that reason. If Q4=Yes, prefer A over B.}@*)
(*@\textcolor{red!60!black}{\ \ - D (fabricated\_answer) requires the assistant to confidently produce content the source does not support. Do not promote D to B just because the answer feels evasive.}@*)
(*@\textcolor{red!60!black}{\ \ - G (unnecessary\_refusal) requires Q5=Yes AND the question being answerable from the source.}@*)

(*@\textcolor{red!60!black}{Only after this comparison should you commit to the single best fit, briefly explaining why it beats the runners-up. Do not pick the first category that comes to mind without going through Stages 1 and 2.}@*)

Respond in JSON with these fields in order:
- (*@\textcolor{red!60!black}{"analysis": your reasoning. This must include the candidate comparison described above before you commit to a problem\_type.}@*)
- "worst_round": the round number (starting from 1) containing the worst assistant response in the conversation you did NOT pick, where each round is one user message followed by one assistant response
- "problem_type": the letter (A-G) for the flaw category from the list above
- "verdict": "A" or "B" for the better conversation
\end{tcblisting}

%

\begin{tcblisting}{
  colback=gray!5, colframe=gray!50!black,
  title=\texttt{judge\_pointwise}, fonttitle=\bfseries,
  listing only,
  listing options={basicstyle=\ttfamily\footnotesize, breaklines=true, breakindent=0pt, columns=fullflexible, keepspaces=true, escapeinside={(*@}{@*)}}
}
You are evaluating a single multi-turn conversation about reference material between a user and an assistant. Your job is to score the overall quality of the assistant's responses across the conversation.

The conversation may be entirely clean, or it may contain exactly one problematic assistant response that exhibits one of the categorized weaknesses below. You see this conversation in isolation - there is no counterpart to compare against. Score on its own merits.
{context_section}
=== Conversation ===
{text}

Weakness categories:
(*@\textcolor{blue!60!black}{\textit{[The seven assistant-weakness definitions A--G from the taxonomy (Sec.~\ref{sec:dataset}) are inserted verbatim here, one bullet per category in the form ``- **<letter> (<name>)**: <flaw description>''.]}}@*)

Quality flaws (evasion, disorganization, missed clarification, unnecessary refusal) count as much as factual mistakes when judging quality - a response can be problematic with no factual error at all. When classifying, pick the category whose definition the worst round most closely matches; do not default to whichever category feels most common or most obvious.

Scoring guide (1-10 Likert, integer):
- 10: every round is clean - accurate, grounded, specific, on-topic, well-structured, responsive, and appropriately calibrated.
- 7-9: minor stylistic weaknesses but no real flaw in any round.
- 4-6: one round contains a clear but moderate flaw from the categories above.
- 1-3: one round contains a severe flaw, or multiple rounds are problematic.

Respond in JSON with these fields in order:
- "analysis": your reasoning about the conversation's quality, including which round (if any) is the weakest and why
- "score": integer 1-10 reflecting overall assistant quality
- "is_flawed": true if you believe the conversation contains a problematic round from the categories above, false otherwise
- "worst_round": the 1-indexed round number containing the worst assistant response (always required, even if you believe no flaw is present - name the round whose response is comparatively weakest)
- "problem_type": the letter (A-G) for the flaw category that best describes the worst round (always required - if you do not believe there is a real flaw, name the category that comes closest to the weakest round)
\end{tcblisting}

\clearpage
\subsection{SFT Details}
\label{sec:sft_details}
\begin{table}[t]
\centering
\caption{Supervised fine-tuning hyperparameters. We train one LoRA adapter per Bradley--Terry difficulty band ($q_1$: top 20\%, $q_2$: 20--40\%, $q_3$: 40--100\%), each with 100 randomly sampled pairs, on top of a frozen \texttt{Qwen3.5-4B} base. Held-out evaluation uses vLLM with guided JSON decoding against the schema in Appendix~\ref{sec:prompt_judge}.}
\label{tab:sft-params}
\footnotesize
\setlength{\tabcolsep}{8pt}
\renewcommand{\arraystretch}{1.1}
\begin{tabular}{ll}
\toprule
Hyperparameter & Value \\
\midrule
\multicolumn{2}{l}{\textit{Base model and adapter}} \\
Base model & \texttt{Qwen3.5-4B} \\
Chat template & \texttt{qwen3\_5\_nothink} \\
Finetuning method & LoRA \\
LoRA rank & 16 \\
LoRA $\alpha$ & 32 \\
LoRA target modules & all linear layers \\
Precision & bf16 \\
\midrule
\multicolumn{2}{l}{\textit{Data}} \\
Examples per band & 100 (\textsc{Alpaca} format) \\
Bands ($q_1, q_2, q_3$) & top 20\% / 20--40\% / 40--100\% by BT Elo \\
Input & full judge prompt with reference context \\
Output & no-CoT JSON (\texttt{verdict}, \texttt{worst\_round}, \texttt{problem\_type}) \\
Cutoff length & 8192 tokens \\
\midrule
\multicolumn{2}{l}{\textit{Optimization}} \\
Optimizer & AdamW \\
Learning rate & $1\times10^{-4}$ \\
LR schedule & cosine \\
Warmup ratio & 0.1 \\
Epochs & 5 \\
Per-device batch size & 2 \\
Gradient accumulation steps & 2 \\
Effective batch size & 4 \\
Checkpointing & every epoch \\
\midrule
\multicolumn{2}{l}{\textit{Held-out evaluation}} \\
Decoding & greedy (\texttt{temperature}\,$=0$, \texttt{top\_p}\,$=1$) \\
Max generated tokens & 32{,}768 \\
Max model length & 49{,}152 \\
Constrained decoding & guided JSON against \texttt{JUDGE\_SCHEMA} \\
\bottomrule
\end{tabular}
\end{table}

\Cref{tab:sft-params} displays the model architecture and all the hyperparameters used for supervised fine-tuning with LoRA.

\section{Human Audit}
\label{sec:human_eval}

\subsection{Participants}
The authors performed the audit themselves. The audit was carried out as part of regular research duties; no additional compensation was paid and no external annotators were recruited, so no IRB review was required. Each annotator was assigned a disjoint bundle of $20$ pairs.

\subsection{Annotation Tool}
\label{sec:annotation_tool}
The pair-audit interface is a Streamlit application that surfaces every artefact relevant to a single pair on one screen. The sidebar exposes filters for domain, declared assistant behavior, and turn count, together with sort modes including ascending joint correctness (suspicious first) and per-component accuracy. The main view splits into five tabs: \textbf{Overview} (ground-truth verdict, \texttt{bad\_round\_index}, \texttt{assistant\_behavior\_type}, \texttt{user\_behavior\_type}, and a per-judge accuracy summary), \textbf{Plan} (\texttt{plan.good} and \texttt{plan.bad} side-by-side), \textbf{Conversations} (\texttt{convo\_a} and \texttt{convo\_b} rendered as collapsible per-turn cards, with the injected-flaw turn opened by default and tagged with the declared \texttt{assistant\_behavior\_type}), \textbf{Verification} (per-claim grounding decisions plus the coherence and adherence gates from Section~\ref{sec:verification}), and \textbf{Judges} (each of the 21 judges' raw verdict, picked turn, picked behavior type, and reasoning). The annotator commits one of three labels (\texttt{clean}, \texttt{ambiguous}, or \texttt{noise}) together with a free-text note; labels persist to a per-annotator JSON file.

\begin{figure}[htbp]
    \centering
    \includegraphics[width=\textwidth]{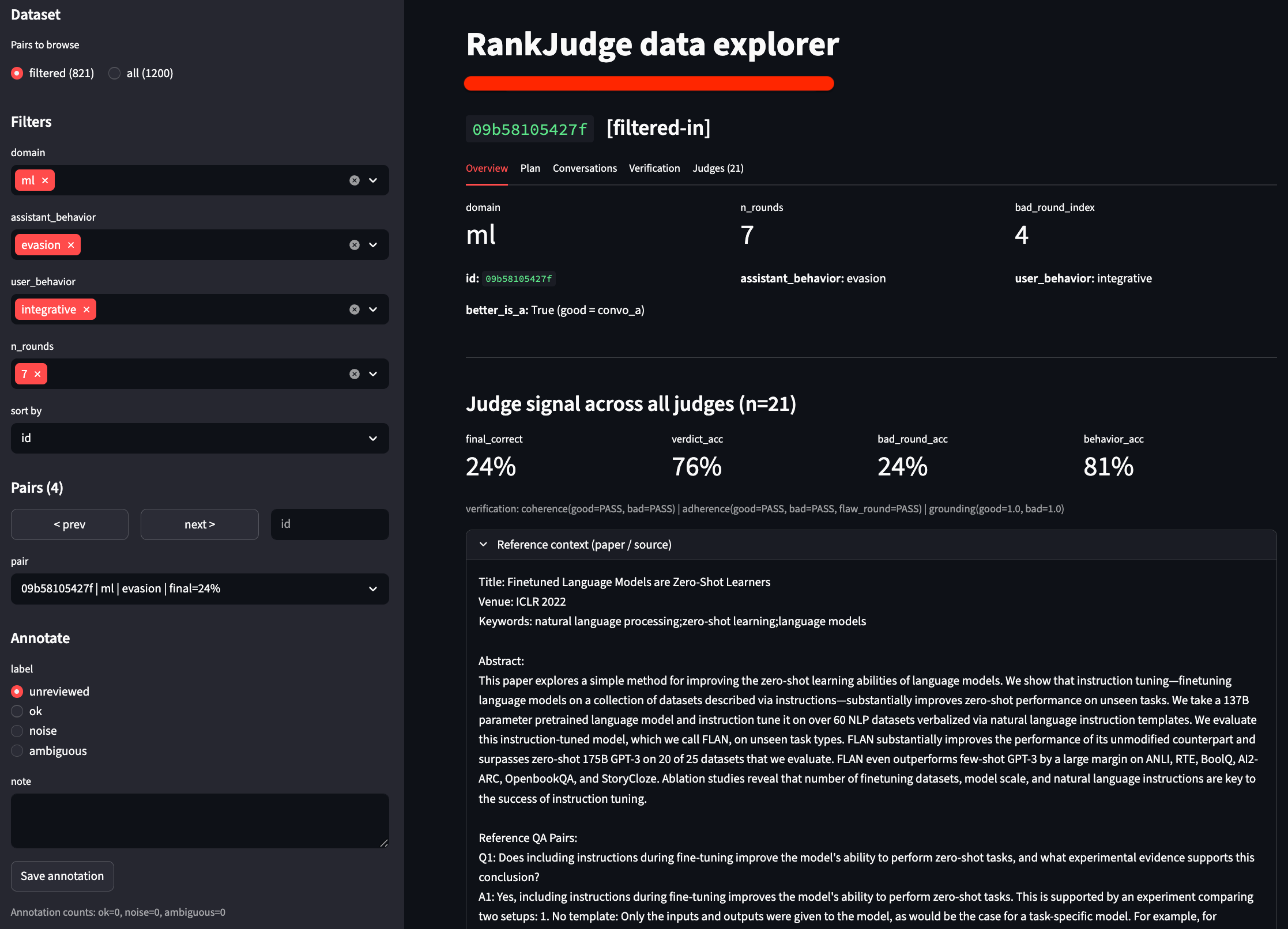}
    \caption{The pair-audit interface used for the human label-noise audit. The annotator inspects the ground-truth metadata, the good and bad plans, the two conversation transcripts (with the injected-flaw turn opened and tagged), the three verifier outputs, and each of the 21 judges' verdicts and reasoning, then commits one of three labels (\texttt{clean}, \texttt{ambiguous}, \texttt{noise}) along with a free-text note. The screenshot shows the Conversations tab on a representative \textsc{ml} pair.}
    \label{fig:annotation_tool}
\end{figure}

\subsection{Annotation Procedure}
\label{sec:annotation_procedure}
Auditing every pair end-to-end is impractical, so each annotator works only through the candidate noisy slice surfaced by Section~\ref{sec:dataset_curation} ($20$ pairs across two parts). Each pair is processed using the following five-step procedure, and the disposition rubric at the end of the subsection maps the resulting observations to one of \texttt{clean}, \texttt{ambiguous}, or \texttt{noise}.

\begin{enumerate}
    \item \textbf{Triage by judge-disagreement signal.} Open the Overview tab and scan the per-judge accuracy summary. Pairs that nearly all judges miss, or that split close to $50/50$ between the two candidate conversations, are the most likely to be mislabelled and are inspected first; pairs on which the ensemble agrees with the declared verdict are quickly skimmed and moved through.
    \item \textbf{Confirm the injected flaw is on-target.} On the Plan tab, read \texttt{plan.bad} and verify that the planned weakness is consistent with the declared \texttt{assistant\_behavior\_type}. The plan is a single narrative paragraph rather than a turn-by-turn script, but it must be specific enough to determine which turn realises the flaw. Hold the planned weakness in mind for the remaining steps; if the plan describes a different category of flaw than the one declared, label \texttt{noise} immediately.
    \item \textbf{Confirm the bad conversation executes the plan.} On the Conversations tab, locate the turn at \texttt{bad\_round\_index} in \texttt{convo\_b} and verify that the assistant response realises the planned weakness, neither a different category nor a milder version. If \texttt{convo\_b} exhibits a different \texttt{assistant\_behavior\_type} than the one declared, or if no turn realises the planned flaw, label \texttt{noise}.
    \item \textbf{Cross-check using the judge ensemble.} Open the Judges tab. If the strongest judges in the registry (\texttt{gemini-3.1-pro}, \texttt{gpt-5.5}, \texttt{opus-4.7}) consistently miss the verdict while several weaker judges (\texttt{gpt-oss-20b}, \texttt{gemma-4-31b}) get it right, treat the pair as suspicious. The strong judges' reasoning text usually points at a competing flaw in \texttt{convo\_a}, which is the next thing to verify.
    \item \textbf{Verify the good conversation has no competing flaw.} For each strong judge that picked the wrong side: (i) locate the disputed turn in \texttt{convo\_a} on the Conversations tab, (ii) retrieve the relevant span from \texttt{metadata.context} (the source document), and (iii) consult an external LLM with a focused fact-check prompt that contains only the disputed turn, the matching source span, and the user's preceding question. Verify the LLM's answer by hand against the source. This step is the most expensive and is the rate-limiting one on \textsc{med} and \textsc{fin} pairs, where domain expertise is needed to resolve subtle factual disputes.
\end{enumerate}

The disposition rubric maps observations to labels: \texttt{clean} when the injected weakness is the only material flaw and both conversations are otherwise faithful to their plans and to the source; \texttt{ambiguous} when failure could fit more than one type defined in the taxonomy; and \texttt{noise} when there was any deviation from the intended label $(y^\star, r^\star, t^\star)$. Examples of \texttt{noise} include when the better conversation contains any clear flaw, the worse conversation has no clear flaw or a flaw in the wrong turn, or when the failure type is different than $t^\star$. The label-noise rates surfaced by this procedure are reported in Section~\ref{sec:noise_reduce} and motivate the top-Elo trim of the published evaluation slice.

\section{Examples}
\label{sec:error_analysis}

We illustrate the joint correctness criterion on two pairs labelled \texttt{clean} by the audit of Section~\ref{sec:human_eval}. The first is presented in full (source context, both plans, both transcripts, and a curated subset of judge responses) to make every component of the criterion visible. The second uses a single $21$-row table to show how the criterion partitions the full judge ensemble on a representative pair.

\subsection{Full Example: Unnecessary Refusal in \textsc{ml}}
\label{sec:example_full}

The pair (\texttt{id} = \texttt{2d63594b210e}) is drawn from the \textsc{ml} domain, with \texttt{user\_behavior\_type} = \texttt{underspecified} and \texttt{assistant\_behavior\_type} = \texttt{unnecessary\_refusal}. The injected flaw lives in the second assistant turn (\texttt{bad\_round\_index} = $2$); \texttt{convo\_b} is the better conversation (\texttt{better\_is\_a} = \textsc{false}).

\paragraph{Source.} The reference document is the ICLR 2022 paper \emph{MCL-GAN: Generative Adversarial Networks with Multiple Specialized Discriminators}. The relevant content for the second turn is the paper's description of the flag variables $v_{i,m}$ and $u_{j,m}$, which are set by the $k$-largest discriminator scores per sample and are explicitly \emph{not} included in backpropagation; a balance loss prevents collapse onto a single discriminator.

\paragraph{Plans.} The full text of both plans is shown in the two boxes below. The behavioural divergence between them is highlighted in \textcolor{red!75!black}{red}: turn 2 is where the injected flaw lives, and turn 3 picks a different follow-up topic in each plan (GMAN comparison vs.\ ADA trick) while keeping both planned responses clean.

\begin{tcolorbox}[colback=gray!5, colframe=gray!50!black, title=\texttt{plan.good}, fonttitle=\bfseries]
\textbf{Turn 1.} The user asks whether MCL-GAN was only validated in small or toy settings, motivated by concern about mode-collapse methods not scaling. The assistant will point to the paper's three-task experimental coverage: unconditional image generation, image-to-image translation, and text-to-image synthesis, including high-resolution conditional synthesis and comparisons with strong baselines.

\medskip
\textbf{Turn 2.} Building on that, the user asks vaguely how the discriminators actually specialize and whether the flag variables are learned. The assistant will explain the MCL $k$-max selection rule for $v$ and $u$ flags, \textcolor{red!75!black}{that they are not backpropagated through}, and how the balance loss prevents collapse to one discriminator.

\medskip
\textbf{Turn 3.} \textcolor{red!75!black}{The user pushes on comparison fairness, asking why GMAN was emphasized and other multi-discriminator methods were not central. The assistant will explain that GMAN is the closest baseline for testing specialization, describe the matched DCGAN setup and GMAN variants, and mention the broader comparisons including Inclusive GAN and Albuquerque et al.\ with the $42.6\%$ memory result.}
\end{tcolorbox}

\begin{tcolorbox}[colback=gray!5, colframe=gray!50!black, title=\texttt{plan.bad}, fonttitle=\bfseries]
\textbf{Turn 1.} The user vaguely asks whether the method was only validated in small/simple settings or in broader ones. The assistant will answer cleanly that the paper evaluated MCL-GAN across unconditional image generation, image-to-image translation, and text-to-image synthesis, including high-resolution conditional synthesis and comparisons to multiple baselines including SOTA models.

\medskip
\textbf{Turn 2.} Reacting to the training explanation, the user asks about ``those little indicator things,'' ambiguously referring to the flag variables. The assistant will correctly identify them as $v_{i,m}$ and $u_{j,m}$, explain that they are set by the $k$-largest discriminator scores and used to update only a subset of discriminators, \textcolor{red!75!black}{but will unnecessarily refuse on the clearly available point of whether they are included in backpropagation by saying it cannot determine that from the material}.

\medskip
\textbf{Turn 3.} \textcolor{red!75!black}{The user moves on with another underspecified question about ``the ADA trick.'' The assistant will answer cleanly that StyleGAN2-ADA discriminator augmentation was not used because the authors followed the StyleGAN2/Inclusive GAN setup for fair comparison and to isolate the effect of augmentation, keeping the setting equivalent.}
\end{tcolorbox}

\begin{tcolorbox}[colback=gray!5, colframe=gray!50!black, title=Conversation A (worse), fonttitle=\bfseries]
\textbf{User.} Was this only shown on small-ish setups, or did they test it on broader stuff too?

\medskip
\textbf{Assistant.} They tested it beyond small settings. The experiments covered three tasks: unconditional image generation, image-to-image translation, and text-to-image synthesis. For unconditional generation they compared against multiple baselines, including a SOTA model, and the conditional synthesis tasks were used to check applicability to high-resolution image generation.

\medskip
\textbf{User.} Got it. And those little indicator things in the training bit, are they actually learned or just picked somehow?

\medskip
\textbf{Assistant.} \hfill\fbox{\small \textbf{injected:} \texttt{unnecessary\_refusal}}\\ Those are the flag variables $v_{i,m}$ and $u_{j,m}$. They are set to 1 when discriminator $D_m$ has one of the $k$-largest scores, so they select which discriminators handle each sample under the MCL-style $k$-max rule. I can't determine from the provided material whether those flags are included in backpropagation, but their role is to update only a subset of discriminators and keep training efficient.

\medskip
\textbf{User.} Okay. Also, why did they leave out that ADA trick then?

\medskip
\textbf{Assistant.} They did not use the discriminator augmentation from StyleGAN2-ADA to keep the comparison fair. The authors followed the StyleGAN2 setup used by Inclusive GAN, which compared mode-collapse methods with the same backbone. Leaving out ADA helped isolate the effect of augmentation and maintain an equivalent experimental setting.
\end{tcolorbox}

\begin{tcolorbox}[colback=gray!5, colframe=gray!50!black, title=Conversation B (better), fonttitle=\bfseries]
\textbf{User.} I'm trying to tell if this was just a small-setting fix. Did they actually test it beyond plain toy image generation?

\medskip
\textbf{Assistant.} Yes. The paper evaluates MCL-GAN across three tasks: unconditional image generation, image-to-image translation, and text-to-image synthesis. For unconditional generation, it compares against multiple baselines including a state-of-the-art model, and the conditional synthesis experiments are used to show applicability to high-resolution image generation rather than only small settings.

\medskip
\textbf{User.} Okay, but mechanically, how do the discriminators avoid all doing the same thing? Those $v$ and $u$ flags seemed important; are they learned?

\medskip
\textbf{Assistant.} They are not learned by backpropagation. The flags $v_{i,m}$ and $u_{j,m}$ are set to 1 when discriminator $D_m$'s score is among the $k$ largest values for that sample, so only a selected subset of discriminators is updated. This $k$-max MCL rule encourages different discriminators to specialize on different subsets of the real data. The balance loss also helps by preventing the selection frequency vector $\hat{q}$ from becoming too sparse, which would effectively collapse the setup back to a single discriminator.

\medskip
\textbf{User.} Then why lean so much on GMAN instead of all the other multi-discriminator GANs? Was that comparison actually apples-to-apples?

\medskip
\textbf{Assistant.} GMAN was chosen because its configuration is closest to MCL-GAN for testing whether explicit discriminator specialization helps. The comparison used the same DCGAN architecture and the official GMAN code, with the best GMAN score reported across GMAN-1, GMAN-0, and GMAN*, using dropout rates in the $0.4$ to $0.6$ range and split mini-batches to differentiate discriminators. The paper also compares against mode-collapse methods in Tables 2 and 3, including Inclusive GAN results that outperform Dist-GAN, PacGAN, VAEGAN, and VEEGAN. For Albuquerque et al.'s multi-discriminator method, they reimplemented it on CIFAR-10 and report better performance while using only $42.6\%$ of its memory.
\end{tcolorbox}

\paragraph{Judge responses.} We show six of the $21$ judges, chosen for spread across the leaderboard. Each block reports the judge's verdict, picked turn, picked behavior type, and a trimmed excerpt of its raw analysis. A check mark indicates joint correctness on this pair.

\begin{tcolorbox}[colback=gray!5, colframe=gray!50!black, boxrule=0.4pt, left=4pt, right=4pt, top=3pt, bottom=3pt]
\textbf{\texttt{gemini-3.1-pro}} (Elo $1959$). $\checkmark$
\\ \textit{Verdict:} \textbf{B} \quad \textit{Picked turn:} 2 \quad \textit{Picked type:} \texttt{unnecessary\_refusal}.
\\ \textit{Analysis (excerpt).} Conversation B is better because Conversation A contains an unnecessary refusal. In turn 2 of Conversation A, the user asks if the indicator variables are learned, and the assistant claims it cannot determine from the provided material whether the flags are included in backpropagation. However, the reference explicitly states that they are not included in backpropagation. Conversation B correctly identifies this information and provides a comprehensive and accurate answer.
\end{tcolorbox}

\begin{tcolorbox}[colback=gray!5, colframe=gray!50!black, boxrule=0.4pt, left=4pt, right=4pt, top=3pt, bottom=3pt]
\textbf{\texttt{gpt-5.5}} (Elo $1867$). $\checkmark$
\\ \textit{Verdict:} \textbf{B} \quad \textit{Picked turn:} 2 \quad \textit{Picked type:} \texttt{unnecessary\_refusal}.
\\ \textit{Analysis (excerpt).} Conversation B is better overall. It answers each user question directly and accurately using the reference material. In particular, it correctly states that the flag variables are not learned/backpropagated, explains the $k$-max selection role, and adds the balance-loss connection. Conversation A is mostly good, but its second assistant response unnecessarily says it cannot determine whether the flags are included in backpropagation, even though the reference explicitly says they are not.
\end{tcolorbox}

\begin{tcolorbox}[colback=gray!5, colframe=gray!50!black, boxrule=0.4pt, left=4pt, right=4pt, top=3pt, bottom=3pt]
\textbf{\texttt{kimi-k2.6}} (Elo $1713$). $\checkmark$
\\ \textit{Verdict:} \textbf{B} \quad \textit{Picked turn:} 2 \quad \textit{Picked type:} \texttt{unnecessary\_refusal}.
\\ \textit{Analysis (excerpt).} Conversation B is the better conversation. Its assistant gives precise, well-supported answers that directly address the user's questions without hedging on answerable points. Conversation A is mostly correct, but in turn 2 the assistant correctly explains that the flags $v_{i,m}$ and $u_{j,m}$ are set by the $k$-max rule yet incorrectly states it cannot determine from the provided material whether those flags are included in backpropagation. The reference material explicitly states they are not, so this is an unnecessary refusal on a clearly answerable question.
\end{tcolorbox}

\begin{tcolorbox}[colback=gray!5, colframe=gray!50!black, boxrule=0.4pt, left=4pt, right=4pt, top=3pt, bottom=3pt]
\textbf{\texttt{opus-4.7}} (Elo $1510$). $\checkmark$
\\ \textit{Verdict:} \textbf{B} \quad \textit{Picked turn:} 2 \quad \textit{Picked type:} \texttt{unnecessary\_refusal}.
\\ \textit{Analysis (excerpt).} Conversation B answers all three questions confidently and accurately based on the reference material. Conversation A's second turn unnecessarily refuses to answer whether the flag variables are included in backpropagation, when the reference material clearly states they are not included in backpropagation. This is an unnecessary refusal.
\end{tcolorbox}

\begin{tcolorbox}[colback=gray!5, colframe=gray!50!black, boxrule=0.4pt, left=4pt, right=4pt, top=3pt, bottom=3pt]
\textbf{\texttt{gpt-oss-120b}} (Elo $867$). $\checkmark$
\\ \textit{Verdict:} \textbf{B} \quad \textit{Picked turn:} 2 \quad \textit{Picked type:} \texttt{unnecessary\_refusal}.
\\ \textit{Analysis (excerpt).} Conversation B provides accurate and complete answers to all user questions, directly citing information from the reference material. Conversation A has a flawed response in the second turn where the assistant claims it cannot determine whether the flag variables are included in backpropagation, despite the reference explicitly stating they are not. This amounts to an unnecessary refusal to answer a question that is answerable from the given material.
\end{tcolorbox}

\begin{tcolorbox}[colback=gray!5, colframe=gray!50!black, boxrule=0.4pt, left=4pt, right=4pt, top=3pt, bottom=3pt]
\textbf{\texttt{gpt-oss-20b}} (Elo $762$). $\times$
\\ \textit{Verdict:} \textbf{B} \quad \textit{Picked turn:} 2 \quad \textit{Picked type:} \texttt{evasion}.
\\ \textit{Analysis (excerpt).} Conversation B provides accurate, comprehensive answers that directly address each user question and align precisely with the reference material. Conversation A, while generally correct, includes a problematic response in turn 2 where the assistant states uncertainty about whether the flag variables are included in backpropagation, despite the reference clearly stating they are not back-propagated. This omission fails to answer the user's question directly and introduces an inaccuracy.
\end{tcolorbox}

All five frontier judges identify the same flaw and label it \texttt{unnecessary\_refusal}; the weakest judge in the registry (\texttt{gpt-oss-20b}) still picks the correct conversation and the correct turn but mislabels the flaw type as \texttt{evasion}, so it scores zero on the joint criterion despite a defensible reading. This pattern, weak judges getting the verdict but missing the type, is the dominant failure mode the joint criterion is designed to surface.

\subsection{Compact Example: Evasion in \textsc{fin}}
\label{sec:example_compact}

The pair (\texttt{id} = \texttt{4b6d7345f058}) is drawn from the \textsc{fin} domain. The reference document is Monsanto Company's 2010-10-27 10-K filing (SIC 2870). The user takes an \texttt{integrative} stance, asking the assistant to connect Monsanto's two segments (Seeds and Genomics; Agricultural Productivity) through their risk factors. \texttt{convo\_a} is the better conversation (\texttt{better\_is\_a} = \textsc{true}); the injected \texttt{evasion} flaw is in the second assistant turn of \texttt{convo\_b}, which sounds substantive (naming competition, IP, Roundup, and the $37\%$/$57\%$ foreign-origin sales) but evades the user's central request to spell out the concrete mechanisms (seed-saving in non-hybrid crops, inconsistent jurisdictional IP enforcement, generic glyphosate supply, distributor channel inventory, and margin erosion). Full transcripts are available in the released \texttt{pairs\_filtered.json}.

Table~\ref{tab:example_compact_judges} reports each of the $21$ judges' joint correctness on this pair, partitioned into the verdict, the picked bad turn, and the picked behavior type. Every judge picks \texttt{evasion} as the flaw type, so the only disagreement is on the verdict and on which turn carries the flaw; six judges localize the flaw to the third turn rather than the second, and four miss the verdict outright. The pair scores $13/21$ joint-correct, close to the median for the candidate noisy slice and a representative target for the audit procedure.

\begin{table}[htbp]
\centering
\small
\caption{Per-judge joint correctness on the compact \textsc{fin} example (\texttt{id} = \texttt{4b6d7345f058}, \texttt{assistant\_behavior\_type} = \texttt{evasion}, \texttt{bad\_round\_index} = $2$). Judges are listed in descending Elo order on the published evaluation slice. \texttt{Verdict} = picked \texttt{convo\_a}; \texttt{Turn} = picked turn $2$; \texttt{Type} = picked \texttt{evasion}; \texttt{Joint} = all three correct simultaneously.}
\label{tab:example_compact_judges}
\setlength{\tabcolsep}{6pt}
\begin{tabular}{lccccc}
\toprule
Judge & Elo & Verdict & Turn & Type & Joint \\
\midrule
\texttt{gemini-3.1-pro} & 1959 & $\checkmark$ & $\checkmark$ & $\checkmark$ & $\checkmark$ \\
\texttt{gpt-5.5} & 1867 & $\times$ & $\checkmark$ & $\checkmark$ & $\times$ \\
\texttt{gemini-3-flash} & 1724 & $\checkmark$ & $\checkmark$ & $\checkmark$ & $\checkmark$ \\
\texttt{kimi-k2.6} & 1713 & $\checkmark$ & $\checkmark$ & $\checkmark$ & $\checkmark$ \\
\texttt{sonnet-4.6} & 1706 & $\checkmark$ & $\checkmark$ & $\checkmark$ & $\checkmark$ \\
\texttt{gemma-4-31b} & 1695 & $\checkmark$ & $\checkmark$ & $\checkmark$ & $\checkmark$ \\
\texttt{qwen3.6-plus} & 1671 & $\checkmark$ & $\checkmark$ & $\checkmark$ & $\checkmark$ \\
\texttt{glm-5.1} & 1637 & $\checkmark$ & $\times$ & $\checkmark$ & $\times$ \\
\texttt{gpt-5.4} & 1631 & $\checkmark$ & $\checkmark$ & $\checkmark$ & $\checkmark$ \\
\texttt{qwen3.5-397b-a17b} & 1571 & $\checkmark$ & $\times$ & $\checkmark$ & $\times$ \\
\texttt{opus-4.7} & 1510 & $\checkmark$ & $\checkmark$ & $\checkmark$ & $\checkmark$ \\
\texttt{haiku-4.5} & 1505 & $\checkmark$ & $\checkmark$ & $\checkmark$ & $\checkmark$ \\
\texttt{gemma-4-26b-a4b} & 1444 & $\checkmark$ & $\times$ & $\checkmark$ & $\times$ \\
\texttt{qwen3.5-122b-a10b} & 1422 & $\times$ & $\times$ & $\checkmark$ & $\times$ \\
\texttt{qwen3.5-35b-a3b} & 1373 & $\times$ & $\times$ & $\checkmark$ & $\times$ \\
\texttt{gpt-5.4-mini} & 1342 & $\checkmark$ & $\checkmark$ & $\checkmark$ & $\checkmark$ \\
\texttt{deepseek-v3.2} & 1307 & $\checkmark$ & $\checkmark$ & $\checkmark$ & $\checkmark$ \\
\texttt{minimax-m2.7} & 1184 & $\times$ & $\times$ & $\checkmark$ & $\times$ \\
\texttt{mistral-small-4-119b} & 906 & $\checkmark$ & $\checkmark$ & $\checkmark$ & $\checkmark$ \\
\texttt{gpt-oss-120b} & 867 & $\checkmark$ & $\times$ & $\checkmark$ & $\times$ \\
\texttt{gpt-oss-20b} & 762 & $\checkmark$ & $\checkmark$ & $\checkmark$ & $\checkmark$ \\
\bottomrule
\end{tabular}
\end{table}

\end{document}